\documentclass{article}
\usepackage{arxiv}

\usepackage[utf8]{inputenc} % allow utf-8 input
\usepackage[T1]{fontenc}    % use 8-bit T1 fonts
\usepackage[hidelinks]{hyperref}       % hyperlinks
\usepackage{url}            % simple URL typesetting
\usepackage{booktabs}       % professional-quality tables
\usepackage{amsfonts}       % blackboard math symbols
\usepackage{nicefrac}       % compact symbols for 1/2, etc.
\usepackage{microtype}      % microtypography
\usepackage{lipsum}		% Can be removed after putting your text content
\usepackage{graphicx}
\usepackage{natbib}
\usepackage{doi}

\usepackage[dvipsnames]{xcolor}
\usepackage{tikz}
\usepackage{dsfont}
\usepackage{amsmath}
\usepackage{amssymb}
\usepackage{graphicx}
\usepackage{mathtools}
\usepackage{booktabs}
\usepackage{siunitx}
\usepackage{amsthm}
\usepackage{amssymb}
\usepackage{subcaption}
\usepackage{environ}
\usepackage{comment}

\makeatletter
\newtheorem*{rep@theorem}{\rep@title}
\newcommand{\newreptheorem}[2]{%
\newenvironment{rep#1}[1]{%
 \def\rep@title{#2 \ref{##1}}%
 \begin{rep@theorem}}%
 {\end{rep@theorem}}}
\makeatother

\newtheorem{definition}{Definition}
\newtheorem{proposition}{Proposition}

\newtheorem{example}{Example}
\newreptheorem{theorem}{Theorem}
\newreptheorem{proposition}{Proposition}
\newreptheorem{lemma}{Lemma}

\newcommand{\myeq}[1]{\stackrel{\mathclap{\normalfont\mbox{#1}}}{=}}

\usepackage[T1]{fontenc}    % use 8-bit T1 fonts
\usepackage{url}            % simple URL typesetting
\usepackage{booktabs}       % professional-quality tables
\usepackage{amsfonts}       % blackboard math symbols
\usepackage{nicefrac}       % compact symbols for 1/2, etc.
\usepackage{microtype}      % microtypography

\renewcommand{\xi}[1][i]{\mathbf{x}^{(#1)}}                                          % x^i, i-th observed value of x
\ifdefined\N                                                                % N, naturals
\renewcommand{\N}{\mathds{N}}                                                % N defined by "siunitx" (which we use), for "NEWTON"
\else
  \newcommand{\N}{\mathds{N}}
\fi
\ifdefined\C
  \renewcommand{\C}{\mathds{C}}                                             % C, complex
\else
  \newcommand{\C}{\mathds{C}}
\fi
\newcommand{\E}{\mathds{E}}                                                 % E, expectation
\newcommand{\var}{\text{Var}}                                             % Var, variance
\newcommand{\idp}{\perp}

\usepackage[ruled]{algorithm2e}
\usepackage[noend]{algpseudocode}

\algrenewcommand\alglinenumber[1]{
    {\sf\footnotesize\addfontfeatures{Colour=888888,Numbers=Monospaced}#1}}
\algrenewcommand\algorithmicrequire{\textbf{Precondition:}}
\algrenewcommand\algorithmicensure{\textbf{Postcondition:}}

\title{Improvement-Focused Causal Recourse (ICR)}

\usepackage[affil-it]{authblk}

\author[1,2,3]{Gunnar König}
\author[4,5]{Timo Freiesleben}
\author[2]{Moritz Grosse-Wentrup}
\affil[1]{Institute for Statistics, LMU Munich}
\affil[2]{Research Group Neuroinformatics, University of Vienna}
\affil[3]{Munich Center for Machine Leanring (MCML)}
\affil[4]{Cluster of Excellence: Machine Learning for Science, University of Tübingen}
\affil[5]{Munich Center for Mathematical Philosophy (MCMP), LMU Munich}

\date{}

\renewcommand{\headeright}{}

\renewcommand{\shorttitle}{Improvement-Focused Causal Recourse (ICR)}

\hypersetup{
pdftitle={A template for the arxiv style},
pdfsubject={q-bio.NC, q-bio.QM},
pdfauthor={David S.~Hippocampus, Elias D.~Striatum},
pdfkeywords={First keyword, Second keyword, More},
}

\begin{document}
\maketitle

\begin{abstract}
Algorithmic recourse recommendations, such
as Karimi et al.'s (2021) causal recourse (CR), inform stakeholders of how to act to revert unfavorable decisions.
However, there are actions that lead to acceptance (i.e., revert the model's decision) but do not lead to improvement (i.e., may not revert the underlying real-world state).
To recommend such actions is to recommend fooling the predictor.
We introduce a novel method, Improvement-Focused Causal Recourse (ICR), which involves a conceptual shift:
Firstly, we require ICR recommendations to guide towards improvement.
Secondly, we do not tailor the recommendations to be accepted by a specific predictor. Instead, we leverage causal knowledge to design decision systems that predict accurately pre- and post-recourse. As a result, improvement guarantees translate into acceptance guarantees.
We demonstrate that given correct causal knowledge ICR%, in contrast to existing approaches, 
guides towards both acceptance and improvement.
\end{abstract}

% keywords can be removed

\section{Introduction}
\label{sec:introduction}

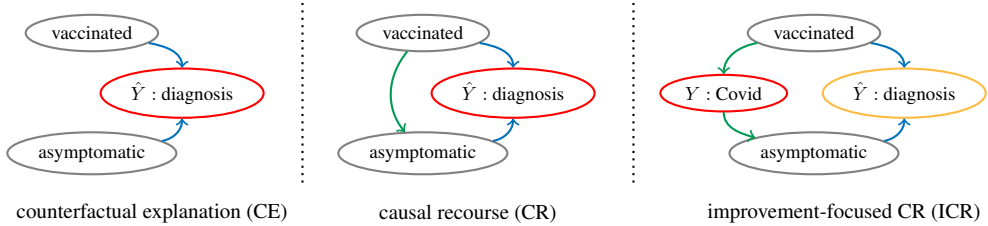
\begin{figure}[t]
    \centering
    \begin{tikzpicture}[thick, scale=0.8, 
    every node/.style={scale=0.8, 
    %line width=0.2mm, 
    %black, fill=white
    }]
    \usetikzlibrary{shapes}
    
        \node[draw=gray, ellipse, scale=0.9] (ux1) at (-4, 1) {vaccinated};
		\node[draw=red, ellipse, scale=0.9] (uy) at (-2.5,0) {$\hat{Y}:$ diagnosis};
		\node[draw=gray, ellipse, scale=0.9] (ux2) at (-4,-1) {asymptomatic};
		\draw[->, RoyalBlue] (ux1) to [out=350,in=90] (uy);
		\draw[->, RoyalBlue] (ux2) to [out=10,in=270] (uy);
		%\draw[->,gray] (ux1) -- (ux2);
		\node[align=center](ce) at (-3, -2){counterfactual explanation (CE)};
		
		\draw[-, dotted] (-.5, 1.5) -- (-.5, -1.5);
		
		\node[draw=gray, ellipse, scale=0.9] (ux1) at (1.5, 1) {vaccinated};
		\node[draw=red, ellipse, scale=0.9] (uy) at (3,0) {$\hat{Y}:$ diagnosis};
		\node[draw=gray, ellipse, scale=0.9] (ux2) at (1.5,-1) {asymptomatic};
		\draw[->, RoyalBlue] (ux1) to [out=350,in=90] (uy);
		\draw[->, RoyalBlue] (ux2) to [out=10,in=270] (uy);
		\draw[->, ForestGreen] (ux1) to [out=230,in=130] (ux2);
		\node[align=center](ar) at (2.25, -2){causal recourse (CR)};
		
		\draw[-, dotted] (5, 1.5) -- (5, -1.5);
		
		\node[draw=gray, ellipse, scale=0.9] (x1) at (8, 1) {vaccinated};
		\node[draw=red, ellipse, scale=0.9] (y) at (6.5,0) {$Y:$ Covid};
		\node[draw=Dandelion, ellipse, scale=0.9] (uy) at (9.5,0) {$\hat{Y}:$ diagnosis};
		\node[draw=gray, ellipse, scale=0.9] (x2) at (8,-1) {asymptomatic};
		\draw[->, RoyalBlue] (x1) to [out=350,in=90] (uy);
		\draw[->, RoyalBlue] (x2) to [out=10,in=270] (uy);
		\draw[->, ForestGreen] (x1) to [out=190,in=90] (y);
		\draw[->, ForestGreen] (y) to [out=270,in=165] (x2);
		\node[align=center](perspective) at (8.5, -2) {improvement-focused CR (ICR)};
    \end{tikzpicture}
    \caption{Directed Acyclic Graph (DAG) illustrating the perspective on model and data taken by counterfactual explanations (CE, left) and causal recourse (CR, center) in contrast to improvement-focused recourse (ICR, right). %Counterfactual explanations ignore causal relationships between features. In comparison,  causal recourse takes causal relationships between features into account. Both CE and CR do not take $Y$ into account.
    Blue edges represent the causal links induced by the prediction model, green edges the real-world causal links, gray nodes the covariates, and the red (yellow) node the primary (secondary) recourse target.
    CR respects the causal relationships but only between input features.
    ICR is the only approach that takes the target $Y$ into account.
    While CE and CR aim to revert the prediction $\hat{Y}$, ICR aims to revert the target $Y$.
    %A more detailed description of the model and data generating process is given in \ref{appendix:details:negative-result}.
    }
    \label{fig:example-intro}
\end{figure}

Predictive systems are increasingly deployed for high-stakes decisions, for instance in hiring \citep{Manish2020}, judicial systems \citep{zeng2015interpretable}, %loan approval \citep{van2017machine}, 
or when distributing medical resources \citep{obermeyer2019dissecting}. A range of work \citep{Wachter2018,ustun_actionable_2019,karimi_algorithmic_2021} develops tools that offer individuals possibilities for so-called algorithmic recourse (i.e. actions that revert unfavorable decisions).
Joining previous work in the field, we distinguish between reverting the model's prediction $\hat{Y}$ (acceptance) and reverting the underlying real-world state $Y$ (improvement) and argue that recourse should lead to acceptance \textit{and improvement} \citep{ustun_actionable_2019,barocas_hidden_2020}. Existing methods, such as counterfactual explanations (CE; \citet{Wachter2018}) or causal recourse (CR; \citet{karimi_algorithmic_2021}), ignore the underlying real-world state and only optimize for acceptance. Since ML models are not designed to predict accurately in interventional environments (i.e. environments where actions have changed the data distribution), acceptance does not necessarily imply improvement.\\
Let us consider a simple motivational example. The goal is to predict whether hospital visitors without recent test certificate are infected with Covid in order to restrict access to tested and low-risk individuals. In the example, the model's \textit{prediction} $\hat{Y}$ represents whether someone is classified to be infected, whereas the \textit{prediction target} $Y$ represents whether someone is actually infected. %As Figure \ref{fig:example-intro} illustrates, %even if the model's pred iction is accurate in a given test distribution,
Target and prediction differ %in their causal roles and, respectively, 
in how they are affected by actions. E.g., intervening on the \textit{symptoms} may change the diagnosis $\hat{Y}$, but will not affect whether someone is infected ($Y$).\\
%Whether an individual is \textit{vaccinated} causally influences whether someone is infected with Covid $Y$. Covid $Y$ causes typical \textit{symptoms}, as for example dry cough. In contrast, both \textit{vaccination} status and \textit{symptom} state are causal for the \textit{prediction} $\hat{Y}$, since the ML model learns to exploit not only the direct cause but also the associated variable.\\
%
Both counterfactual explanations (CE) and causal recourse (CR) only target $\hat{Y}$ (Figure \ref{fig:example-intro}). Therefore, CE and CR may suggest to alter the %non-causal variable 
\textit{symptoms} (e.g., by taking cough drops) and thereby may recommend to \textit{game} the predictor: Although the intervention leads to acceptance the actual Covid risk $Y$ is not improved.%This way, they change the Covid prediction ($\hat{Y}$) without actually reducing the probability of having Covid ($Y$), 
\footnote{In \ref{appendix:details:negative-result}, the case is formally demonstrated.}
\\
One may argue that this is an issue of the prediction model %that should not incentivize gaming. Thus, one 
and may adapt the predictor strategically to make gaming less lucrative than improvement \citep{miller_strategic_2020}. In our example, the model's reliance on the symptom state would need to be reduced. However, such strategic adaptions may come at the cost of predictive performance since gameable variables, like the symptom state, can be highly predictive \citep{shavit_causal_2020}. Thus, we tackle the problem by adjusting the explanation.
%
%We propose Improvement-Focused Causal Recourse (ICR) (Section \ref{sec:mcr}), a method that guides individuals towards improvement. %of the underlying target $Y$.
%Since estimating the effects of actions is a causal problem, we require knowledge of the structural causal model (SCM) or the causal graph.
%
%To ensure that improvement leads to acceptance, we show how to exploit said causal knowledge for accurate post-recourse prediction (Section \ref{sec:accurate-post-recourse} and \ref{sec:acceptance-guarantees}).

\paragraph{Contributions}{%In contrast to previous work on recourse, we separate reverting the prediction $\hat{Y}$ (acceptance) from reverting the underlying real-world state $Y$ (improvement). We demonstrate that even though causal recourse (CR) recommendations lead to acceptance by the predictor $(\hat{Y}=1)$, they may fail to improve the underlying target $Y$. %(Section \ref{sec:introduction}, \ref{appendix:details:negative-result}).
%Instead of strategically adapting the model at the cost of predictive performance, we suggest to strategically adapt the recourse explanations.
%
%More specifically, we argue that improvement should be the primary goal of recourse (Section \ref{sec:two-tales}). 
We present improvement-focused causal recourse (ICR), the first recourse method that targets improvement instead of acceptance. 
%improvement-focused causal recourse (ICR), a novel improvement-focused recourse technique that exploits either the knowledge of a structural causal model (SCM) or the causal graph (Section \ref{sec:mcr}).
%
Since estimating the effects of actions is a causal problem, causal knowledge is required. 
More specifically, we show how to exploit either knowledge of the structural causal model (SCMs) or the causal graph to guide towards improvement (Section \ref{sec:mcr}).
On a conceptual level we argue that the individual's improvement options should not be limited by an acceptance constraint (Section \ref{sec:two-tales}).
In order to nevertheless yield acceptance, we show how to exploit said causal knowledge to design post-recourse decision systems that in expectation recognize improvement (Section \ref{sec:accurate-post-recourse}), such that improvement guarantees translate into acceptance guarantees (Section \ref{sec:acceptance-guarantees}). On synthetic and semi-synthetic data, we demonstrate that ICR, in contrast to existing approaches, 
leads to improvement and acceptance (Section \ref{sec:simulation}).
%
%Our technical contributions include theoretical results on the estimation of the structural counterfactual given unobserved variable $Y$, the estimation of an individualized-post recourse predictor, the derivation of conditions under which optimal observational models predict accurately in post-recourse environments as well as conditions for CR and ICR to coincide.
%While the acceptance probability for CR is encoded in a non-interpretable hyperparameter, we derive interpretable post-recourse acceptance bounds, thereby enabling the explainee to make an informed choice.
}

\section{Related Work}
\label{sec:related-work}
\paragraph{Constrastive Explanations}{
Contrastive explanations explain decisions by contrasting them with alternative decision scenarios \citep{karimi2020survey,stepin2021survey}; a well known example are counterfactual explanations (CE) that highlight the minimal feature changes required to revert the decision of a predictor $\hat{f}(x)$ \citep{Wachter2018,dandl_multi-objective_2020}.
%Counterfactual explanations (CE) aim to find the smallest feature changes that would revert the decision of a predictor $\hat{f}(x)$ \citep{Wachter2018,dandl_multi-objective_2020,karimi_algorithmic_2021,ustun_actionable_2019}.
However, CEs are ignorant of causal dependencies in the data and therefore in general fail to guide action \citep{karimi_algorithmic_2021}. 
In contrast, the causal recourse (CR) framework by \citet{karimi2022towards} takes the causal dependencies between covariates into account: More specifically, \citet{karimi2022towards} use structural causal models or causal graphs to guide individuals towards acceptance.\footnote{For the interested reader, we formally introduce CR in our notation in \ref{sec:background-recourse}.}} The importance of improvement was discussed before \citep{ustun_actionable_2019,barocas_hidden_2020}, but as of now no improvement-focused recourse method was proposed.
%The idea to guide individuals towards improvement rather than acceptance has already been mentioned in \citep{ustun_actionable_2019,barocas_hidden_2020}, however, our approach is the first to develop it.
%
\paragraph{Strategic Classification}{
\label{sec:related:strategic}
The related field of strategic modeling investigates how the prediction mechanism incentivizes rational agents \citep{hardt2016strategic,tsirtsis2020decisions}. 
A range of work \citep{bechavod_causal_2020,chen2020linear,miller_strategic_2020} thereby distinguishes models that incentivize \textit{gaming} (i.e., interventions that affect the prediction $\hat{Y}$ but not the underlying target $Y$ in the desired way) and \textit{improvement} (i.e., actions that also yield the desired change in $Y$).
%\citet{chen2020linear} suggest adapting predictors such that incentivizing improvement and predictive accuracy are traded-off. They forgo a causal formalism and assume linear relationships and causally independent covariates.
%\citet{bechavod_causal_2020} demonstrate that strategic agents behaviour in combination with model refits can help to distinguish causal from non-causal features and to learn a causal model that does not incentivize gaming.
%\citet{tsirtsis2020decisions} link the field with contrastive explanations and investigate how to jointly design decision policies and counterfactual explanations to maximize utility.
%Strategic classification is not only studied in the context of ML but also in ecnomomics \citep{kleinberg2020classifiers,haghtalab2020maximizing}. %discuss the design of mechanisms such that improvement is supported but gaming is not.
Strategic modeling is concerned with adapting the model, where except for special cases the following three goals are in conflict: incentivizing improvement, predictive accuracy, and retrieving the true underlying mechanism \citep{shavit_causal_2020}.}
\paragraph{Robust algorithmic recourse}{
%
%According to \citet{barocas_hidden_2020} and \citet{venkatasubramanian_philosophical_2020} counterfactual explanations (CE) only provide reliable information about relevant alternative predictions if the model is stable over time. Recourse, on the other side, should be a \textit{robust good}: For individuals who implement recourse recommendations acceptance should be guaranteed even if model and data have shifted.
%In a similar vein, \citet{Wachter2018} suggest guaranteeing counterfactual-based recourse within a pre-specified period of time.
%
The robustness of CEs and CR has been investigated before \citep{rawal_can_2020,pawelczyk_counterfactual_2020,upadhyay2021towards,dominguez2021adversarial,pawelczyk2022algorithmic}, yet only with respect to generic shifts of model and data. 
Only \cite{pawelczyk_counterfactual_2020} investigate the robustness regarding refits on the same data. They find that on-the-manifold CEs are more robust than standard CEs. In contrast, we empirically compare the robustness of CE, CR and ICR with respect to refits on the same data.
%
%\citet{rawal_can_2020} and \citet{upadhyay_towards_2021} assess the impact of data correction, temporal and geospatial shifts. \citet{upadhyay_towards_2021} introduce ROAR, an algorithm inspired by adversarial training that optimizes a novel objective that incentivizes robustness of recourse.
%\citet{pawelczyk_counterfactual_2020} demonstrate that counterfactuals that lie within the support of the observational distribution are more robust to model multiplicity.
%\citet{dominguez2021adversarial,pawelczyk2022algorithmic} introduce causal recourse that is robust to uncertainty in the input features.

%Inspired by Goodhard's Law \cite{goodhart1984problems} and work in strategic prediction (Section \ref{sec:related:strategic}), we are the first to assess the robustness of recourse w.r.t. distribution shifts that were induced by the recourse actions themselves. We argue that MCR recommendations are more robust regarding refits of the model on post-recourse data than CR (Section \ref{subsec:mcr:robustness}) and support the claim on a simulated example (Section \ref{sec:simulation}).
}
\section{Background and Notation}
\label{sec:background-notation}
%
% Notation for P and p (probabilities): https://ocw.mit.edu/courses/mathematics/18-05-introduction-to-probability-and-statistics-spring-2014/readings/MIT18_05S14_Reading13b.pdf 
%
\paragraph{Prediction model}{
We assume binary probabilistic predictors and cross-entropy loss, such that the optimal score function $h^*(x)$ models the conditional probability $P(Y=1|X=x)$, which we abbreviate as $p(y|x)$. We denote the estimated score function as $\hat{h}(x)$, which can be transformed into the binary decision function $\hat{f}(x) := [\hat{h}(x) \geq t]$ via the decision threshold $t$. %. For the decision threshold $t$ the decision is positive $\hat{f}(x) = 1$ if $\hat{h}(x) \geq t$ and respectively $\hat{f}(x) = 0$ if $\hat{h}(x) < t$. 
%Given $t=0.5$ the most probable class is chosen.
}
\paragraph{Causal data model}{
We model the data generating process using a structural causal model (SCM) $\mathcal{M} \in \Pi$ \citep{Pearl2009,Peters2017book}. The model $\mathcal{M} = \langle X, U, \mathbb{F} \rangle$ consists of the endogenous variables $X \in \mathcal{X}$, the mutually independent exogenous variables $U \in \mathcal{U}$, and structural equations $\mathbb{F}: \mathcal{U} \to \mathcal{X}$. Each structural equation $f_j$ specifies how $X_j$ is determined by its endogenous causes and the corresponding exogenous variable $U_j$. The SCM entails a directed graph $\mathcal{G}$, where variables are connected to their direct effects via a directed edge.\\
The index set of endogenous variables is denoted as $D$. The parent indexes of node $j$ are referred to as $pa(j)$ and the children indexes as $ch(j)$. We refer to the respective variables as $X_{pa(j)}$. We write $X_{pa(j)}$ to denote all parents excluding $Y$ and $(X,Y)_{pa(j)}$ to denote all parents including $Y$. All ascendant indexes of a set $S$ are denoted as $asc(S)$, its complement as $nasc(S)$, all descendant indexes as $d(S)$, and its complement as $nd(S)$.\\
%The set of observed variables is denoted as $O$ where $O \subseteq D$. For the most part, we either assume $O=D$ or equivalently causal sufficiency of $O$, meaning that there is no variable $j \not \in O$ that is a common cause of two variables $k, l \in O$ (see e.g., \citep{Peters2017book}).
% The Markov blanket $MB_O(Y)$ is the minimal subset of $O$ that allows for an optimal prediction of $Y$ (i.e. for which $X_O \idp Y | X_{MB_O(Y)}$).\footnote{Sometimes the Markov blanket is defined as the minimal $d$-separating set. If faithfulness and the Markov property are fulfilled, both definitions coincide.}\\
%
SCMs allow to answer causal questions. This means that they cannot only be used to describe (conditional) distributions (observation, rung 1 on Pearl's ladder of causation \citep{Pearl2009}), but can also be used to predict the (average) effect of actions $do(x)$ (intervention, rung 2) and imagine the results of alternative actions in light of factual observation $(x, y)^F$ (counterfactuals, rung 3).\\
As such, we model actions as structural interventions $a : \Pi \to \Pi$, which can be constructed as $do(a) = do(\{X_i := \theta_i\}_{i \in I})$, where $I$ is the index set of features to be intervened upon.
A model of the interventional distribution can be obtained by fixing the intervened upon values to $\theta_I$ (e.g. by replacing the structural equation $f_I := \theta_I$).
Counterfactuals can be computed in three steps \citep{Pearl2009}: First, the factual distribution of exogenous variables $U$ given the factual observation of the endogenous variables $x^{F}$ is inferred (\textit{abduction}) (i.e., $P(U_j|X^{F})$). Second, the structural interventions corresponding to $do(a)$ are performed (\textit{action}). Finally, we can sample from the  counterfactual distribution $P(X^{SCF}|X=x^{F}, do(a))$ using the abducted noise and the intervened-upon structural equations (\textit{prediction}).}
%
%\paragraph{Recourse terminology} {I could potentially include a paragraph where we explain important terms such as explainee, data subject, recourse reocmmendation, improvement, acceptance.}

\section{The Two Tales of Contrastive Explanations}
\label{sec:two-tales}

In the introduction we have demonstrated that CE and CR may suggest to game the predictor (i.e. guide towards acceptance without improvement).
To tackle the issue, we will introduce a new explanation technique called improvement-focused causal recourse (ICR) in Section \ref{sec:mcr}.\\
In this section we lay the conceptual justification for our method.
More specifically, we argue that for recourse the acceptance constraint of CR should be \textit{replaced} by an improvement constraint.
Therefore, we first recall that a multitude of goals may be pursued with contrastive explanations \citep{Wachter2018} and separate two purposes of contrastive explanations: \textit{contestability of algorithmic decisions} and \textit{actionable recourse}.
We then argue that improvement is an essential requirement for recourse and that the individual's options for improvement should not be limited by acceptance constraints.
% (Section \ref{subsec:two-tales-improvement}).
%Moreover, recourse recommendations should not be constrained to reach acceptance by a potentially instable predictor.
% (Section \ref{subsec:meaningfulness-first}).
%Clearly, in order to contest algorithmic decisions, more diverse explanations should be offered as well (Section \ref{subsec:separate-for-contestability}).

\paragraph{Contestability and recourse are distinct goals.}{\textit{Contestability} is concerned with the question of whether the algorithmic decision is correct according to common sense, moral or legal standards. Explanations may help model authorities to detect violations of such standards or enable explainees to contest unfavorable decisions \citep{Wachter2018,freiesleben2020counterfactual}. 
%What is contested is the decision itself or the decision-making process.
Explanations that aim to enable contestability must 
%faithfully
reflect the model's rationale for an algorithmic decision. 
%\\
%Since not all violations of standards can be characterized exhaustively, it is important to offer diverse explanations \citep{russell2019efficient}.
%such that the explainee can contest decisions based on their personal beliefs and opinions.\\
%
% short description of recourse requirements that have been formulated so far
\textit{Recourse recommendations} on the other hand need to satisfy various constraints unrelated to the model, such as causal links between variables \citep{karimi_algorithmic_2021} or their actionability \citep{ustun_actionable_2019}. %Also, recourse recommendations must be plausible, i.e., make realistic suggestions that are jointly satisfiable and prefer sparse over widespread action recommendations \citep{karimi2020model,dandl_multi-objective_2020}.\\
%
% short description of status-quo differences
Consequently, explanations geared to contest are more complete and true to the model while recourse recommendations are more selective and true to the underlying process.\footnote{We do not claim that recourse and contestability always diverge, we only describe a difference in focus. If contesting is successful it may even provide an alternative route towards recourse.}
%and account for the limitations of the explainee. %\footnote{Yet, it could be argued that contesting decisions additionally offers an alternative route towards recourse in cases where model authorities admit mistakes and adapt their algorithmic decision-making system accordingly.}
We believe that the selectivity and reliance of recourse recommendations on factors besides the model itself is not a limitation but an indispensable condition for making explanations more relevant to the explainee.}

\paragraph{In the context of recourse, improvement is desirable for model authority and explainee.}{
% we go step further: require meaningfulness
We consider improvement to be an important normative requirement for recourse, both with respect to explainee and model authority. Valuable recourse recommendations enable explainees to plan and act; thus, such recommendations must either provide indefinite validity or a clear expiration date \citep{Wachter2018,barocas_hidden_2020,venkatasubramanian_philosophical_2020}. Problematically, when model authorities give guarantees for non-improving recourse, this constitutes a binding commitment to misclassification. However, if model authorities do not provide recourse guarantees over time, this diminishes the value of recourse recommendations to explainees. They might invest effort into non-improving actions that ultimately do not even lead to acceptance because the classifier changed.\footnote{For instance, in the introductory example, an intervention on the symptom state would only be honored by a refit of the model on pre- and post-recourse data for the small percentage of individuals who were already vaccinated, as documented in more detail in \ref{appendix:details:negative-result}. Also, gaming actions may not be robust concerning model multiplicity, as seen in the experiments (Section \ref{sec:simulation}).}  In contrast, improvement-focused recourse is honored by any accurate classifier. % and may therefore be guaranteed by model authorities. 
We conclude that, given these advantages for both model authority and explainee, recourse recommendations should help to improve the underlying target $Y$.\footnote{We do not claim that gaming is necessarily bad; it may be justified when predictors perform morally questionable tasks.}}

\paragraph{Improvement should come first, acceptance second.}{
% should we go another step, optimize both, improvement and acceptance?
Taken that we constrain the optimization on improvement, how to guarantee acceptance remains an open question. One approach would be to constrain the optimization on both improvement and acceptance. However, a restriction on acceptance is either redundant or, from our moral standpoint, questionable: If improvement already implies acceptance, the constraint is redundant. In the remaining cases, we can predict improvement with the available causal knowledge but would withhold these (potentially less costly) improvement options because of the limitations of the observational predictor. 
% improvement options (that are even potentially easier to realize) would be withheld from the explainee because of the observational model's deficiency. %\footnote{Respectively, we would argue that if improvement does not lead to acceptance the explainee is given grounds contest the decision.}
%or questionable, because the explainee's agency would be restricted by the model's inability to predict accurately in interventional environments.
To ensure that acceptance ensues improvement, we instead suggest to exploit the assumed causal knowledge for accurate post-recourse prediction (Section \ref{sec:accurate-post-recourse}), such that acceptance guarantees can be made (Section \ref{sec:acceptance-guarantees}).  %In layman's terms, given that recourse leads to improvement, to ensure that recourse leads to acceptance
%Therefore the predictor must predict accurately post-recourse.
%Since recourse leads to distribution shifts (as demonstrated in the introductory example), accurate post-recourse prediction requires that the model is stable w.r.t recourse interventions.
%In order to design such intervention stable predictors, we can exploit the causal knowledge that is anyway required to guide towards improvement .\\
%For the explainees to maintain agency it is important that they can make an informed choice. As such, an intelligible metric like the acceptance probability for a suggested action must be communicated to the individuals.
}
%
% what if other setting -> use other techniques
%\paragraph{Separate explanations to contest algorithmic decisions}{
%\label{subsec:separate-for-contestability}
%Meaningful recourse guides individuals towards actions that help them to improve, e.g., it recommends a vaccination to lower the risk to get infected with Covid. If, however, a explainee is more interested in contesting the algorithmic decision, (meaningful) recourse recommendations are not suitable. Think of an individual who is denied entrance to an event because of their high Covid risk prediction, which is based on a non-causal, spurious association with their country of origin\footnote{E.g., due to a spurious association with the causal variable \textit{type of vaccine}.}. In such situations, we suggest to additionally show explainees diverse explanations, which reflect the model's prediction mechanism and therefore enable to contest the decision. For example, such an explanation could be: if your country of origin would be different, your predicted Covid risk would have been lower.}

\section{Improvement-Focused Causal Recourse (ICR)}
\label{sec:mcr}

We continue with the formal introduction of ICR, an explanation technique that targets improvement ($Y=1$) instead of acceptance $(\hat{Y}=1)$. Therefore we first define the improvement confidence $\gamma$, which can be optimized to yield ICR.
Like previous work in the field \citep{karimi_algorithmic_2020}, we distinguish two settings: In the first setting, knowledge of the SCM can be assumed, such that we can leverage structural counterfactuals (rung 3 on Pearl's ladder of causation) to introduce the individualized improvement confidence $\gamma^{ind}$. In the second setting only the causal graph is known, which we exploit to propose the subpopulation-based improvement confidence $\gamma^{sub}$ (rung 2).

%For each setting we introduce a notion of meaningfulness. 
%Further, we demonstrate how to design accurate decision systems such that in expectation the post-recourse model predicts the respective improvement probability. As a consequence, we are able to show that implementing ICR recommendations (which target improvement) also lead to acceptance (Section \ref{subsec:mcr:acceptance-guarantees}). 
%
%\subsection{Individualized Meaningful Causal Recourse (inICR)}
%\label{sec:individualized-mar}
\paragraph{Individualized improvement confidence}{
\label{subsec:marin-optimization}
%
% why is SCM/counterfacutal mighty
For the individualized improvement confidence $\gamma^{ind}$ we exploit knowledge of a SCM. SCMs can be used to answer counterfactual questions (rung 3). In contrast to rung-2-predictions, counterfactuals are tailored to the individual and their situation \citep{Pearl2009}: They ask what would have been if one had acted differently and thereby exploit the individual's factual observation. Given unchanged circumstances, counterfactuals can be seen as individualized causal effect predictions.\\
%
% how does the setting differ from karimi
In contrast to existing SCM-based recourse techniques \citep{karimi2022towards} we include both the prediction $\hat{Y}$ and the target variable $Y$ as separate variables in the SCM. As a result, the SCM can be used not only to model the individualized probability of acceptance, but also the individualized probability of improvement.
%, and therefore to introduce an individualized notion of meaningfulness (Section \ref{subsec:marin-optimization}).
%Moreover we suggest to exploit the SCM not only to generate explanations, but also for accurate post-recourse decision making (Section \ref{subsec:individualized-prediction}), such that acceptance guarantees can be derived (Section \ref{subsec:mcr:acceptance-guarantees}).
%
%
%
% objective on one sentence
%More specifically, we consider an action $a$ to be meaningful for an individual with observation $x^{pre}$ if the counterfactual outcome $y^{post}$ is favorable.
%This counterfactual is probabilistic.\footnote{Since $Y$ cannot be observed the counterfactual is only deterministic if (1) $Y$ can be perfectly predicted from the covariates and (2) the respective structural equation is invertible.} 
%Therefore, we require the counterfactual outcome to be favorable with user-specified probability $\gamma$.
%\footnote{Building on the deterministic criterion in \citep{karimi_algorithmic_2021}, \citep{karimi_algorithmic_2020} present probabilistic causal recourse optimization problems (Equations \ref{eq:ar-individualized} and \ref{eq:ar-subpopulation}). They require the expectation of the counterfactual prediction to exceed some context-dependent threshold. The formulation is similar to our formulation since the threshold can be translated into a confidence $\gamma$. For more details refer to \ref{appendix:details:expectation-objective}.}
%
\begin{definition}[Individualized improvement confidence]  For pre-recourse observation $x^{pre}$ and action $a$ we define the individualized improvement confidence as
  \[\gamma^{ind}(a) = \gamma(a, x^{pre}) := P(Y^{post} = 1|do(a), x^{pre}).\]
%where $P(Y^{post, a} = 1)$ is the shorthand for the counterfactual probability $P(Y = 1|do(X_{I^a} = \theta_a), x^{pre})$.
\label{def:meaningful-individualized}
\end{definition}
%
%inICR differs from inCR (Equation \ref{eq:ar-individualized}) in its target: we optimize the counterfactual outcome of $Y$ instead of $\hat{Y}$. %
Since the pre-recourse (factual) target $Y$ cannot be observed, standard counterfactual prediction cannot be applied directly. However, we can regard the distribution as a mixture with two components, one for each possible state of $Y$. We can estimate the mixing weights using $h^*$ and each component using standard counterfactual prediction. Details including pseudocode are provided in \ref{appendix:estimation:individualized}.}
\paragraph{Subpopulation-based improvement confidence}{
\label{subsubsec:mcr:subpopulation:meaningfulness}
%\subsection{Subpopulation-based Meaningful Causal Recourse (sICR)}
%\label{sec:subpopulation-mar}
%
For the estimation of the individualized improvement confidence $\gamma^{ind}$ knowledge of the SCM is required.
%For the individualized estimation of $\gamma^{ind}$ the pre-recourse observation and knowledge of the SCM are exploited to accurately estimate the individualized effect of an intervention.
%However, as it relies on structural counterfactuals, individualized ICR requires knowledge of the SCM.
If the SCM is not specified, but the causal graph is known instead and there are no unobserved confounders (causal sufficiency), we can still estimate the effect of interventions (rung 2).\\
In contrast to counterfactual distributions (rung 3), interventional distributions describe the whole population and therefore provide limited insight into the effects of actions on specific individuals.
Building on \citet{karimi_algorithmic_2020}, we thus narrow the population down to a subpopulation of similar individuals, for which we then estimate the subpopulation-based causal effect. More specifically, we consider individuals to belong to the same subgroup if the variables that are not affected by the intervention take the same values. For action $a$, we define the subgroup characteristics as $ G_a := nd(I_a)$ (i.e., the non-descendants of the intervened-upon variables in the causal graph).\footnote{The estimand resembles the conditional treatment effect with $G_a$ being effect modifiers \citep{hernan_ma_causal_2020}.} More formally, we define the subpopulation-based improvement confidence $\gamma^{sub}$ as the probability of $Y$ taking the favorable outcome in the subgroup of similar individuals (Definition \ref{def:subpop-meaningful}).
\begin{definition}[Subpopulation-based improvement confidence] 
Let $a$ be an action that potentially affects $Y$, i.e. $I_a \cap asc(Y) \neq \emptyset$.\footnote{If $a$ cannot affect $Y$, we can predict $P(Y|x^{pre}, do(a))=P(Y|x^{pre})$ using the optimal observational predictor $h^*$.} Then we define the subpopulation-based improvement confidence as
\begin{equation*}
 \gamma^{sub}(a) = \gamma(a, x^{pre}_{G_a}) := P(Y^{post} = 1|do(a), x^{pre}_{G_{a}}).
\label{eq:subpopulation-meaningful}
\end{equation*}
\label{def:subpop-meaningful}
\end{definition}
The set $G_a$ is chosen for practical reasons. In order to make the estimation more accurate, we would like to 
%make the subgroup as specific as possible, and therefore 
condition on as many characteristics as possible. However, without access to the SCM, one can only identify interventional distributions for subgroups of the population by conditioning on their (unobserved) post-intervention characteristics (but not by conditioning on their pre-intervention characteristics) \citep{Pearl2009,glymour2016causal}. If we were to select a subgroup from a post-recourse distribution by conditioning on pre-recourse characteristics that are affected by $a$ (e.g. strong pre-recourse symptoms), we yield a group that the individual may not be part of (e.g. people with strong post-recourse symptoms). In contrast, 
for $X_{G_a}$ pre- and post-intervention values coincide, such that we can estimate $\gamma^{sub}$: %by estimating $P(Y^{post}=1|do(a),X_{G_a}^{post}=x_{G_a}^{pre})$
Assuming causal sufficiency, the standard procedure to sample interventional distributions can be applied, only that additionally $X_{G_a}^{post} := x_{G_a}^{pre}$. Based on the sample $\gamma^{sub}$ can be estimated (as detailed in \ref{appendix:estimation:sampling-sub}).\\
%
%We can estimate $\gamma^{sub}$ using the standard sequential procedure for sampling from interventional distributions, only that we additionally condition 
%Given that there are no unobserved confounders (causal sufficiency), $\gamma^{sub}$ can be estimated using the causal graph $\mathcal{G}$ and the conditional distributions of nodes given their parents (\ref{appendix:details:observational-identifiability}).
%
The estimation of $\gamma^{sub}$ does not require knowledge of the SCM, but is less accurate than $\gamma^{ind}$. In the introductory example, for the action \textit{get vaccinated} the set of subgroup-characteristics $G_a$ is empty. As such, $\gamma^{sub}$ is concerned with the effect of a vaccination over the whole population. If we were to observe \textit{zip code}, a variable that is not affected by \textit{vaccination}, $\gamma^{sub}$ would indicate the effect of vaccination for subjects that share the explainee's \textit{zip code}. In contrast, $\gamma^{ind}$ also takes the explainee's \textit{symptom state} into account. 
}
\paragraph{Optimization problem}{
%
% How can the optimization be performed?
To generate ICR recommendations, we can optimize Equation \ref{eq:optimization-problem}. We aim to find actions that meet a user-specified improvement target confidence $\overline{\gamma}$ with minimal cost for the recourse seeking individual. The cost function cost$(a, x^{pre})$ captures the effort the individual requires to perform action $a$ \citep{karimi_algorithmic_2020}.\\
As for CE or CR,
the optimization problem for ICR is computationally challenging (\ref{appendix:details:optimization}). It can be seen as a two-level problem, where on the first level the intervention targets $I_a$, and on the second level the corresponding intervention values $\theta_a$ are optimized \citep{karimi_algorithmic_2020}. Since we target improvement, we can restrict $I_a$ to causes of $Y$. Following \citet{dandl_multi-objective_2020}, we use the genetic algorithm NSGA-II \citep{deb2002fast} for optimization.
%It can be seen as a two-stage problem, where in the first stage the intervention targets $I_a$, and in the second stage the corresponding intervention values $\theta_a$ are optimized \citep{karimi_algorithmic_2020}.
%For the selection of intervention targets $I_a$ alone $2^{d'}$ combinations exist, with $d'<d$ being the number of causes of $Y$. %We use the \textit{Nondominated Sorting Genetic Algorithm II} (NSGA-II) \citep{deb2002fast}.
%The computational complexity of optimizing the intervention value $\theta$ depends on the type of data. 
%
%For mixed categorical and continuous data, previous work in the field \citep{dandl_multi-objective_2020} suggests to use NSGA-II \citep{deb2002fast} in combination with \textit{mixed integer evaluation strategies} \citep{li2013mixed}.We discuss the complexity of the problem and solution strategies in \ref{appendix:details:optimization}.
%
\begin{equation}
    \text{argmin}_{a=do(X_I = \theta)} \quad \text{cost}(a, x^{pre}) \quad \text{s.t.} \quad \text{$\gamma(a) \geq \overline{\gamma}$}.
    \label{eq:optimization-problem}
\end{equation}
}
%\citet{karimi_algorithmic_2020} exploit assumptions about the SCM and distribution to derive a gradient-based solution.
%For our experiments binary variables are used, such that the second stage optimization is not necessary.

\section{Accurate Post-Recourse Prediction}
\label{sec:accurate-post-recourse}

Recourse recommendations should not only lead to improvement $Y$ but also revert the decision $\hat{Y}$.
Whether acceptance guarantees naturally ensue from $\gamma$ depends on the ability of the predictor to recognize improvements.
As follows, we demonstrate how the assumed causal knowledge can be exploited to design accurate post-recourse predictors. We find that an individualized post-recourse predictor is required to translate $\gamma^{ind}$ into an individualized acceptance guarantee, but curiously that the observational predictor is sufficient in supopulation-based settings. %Again, we distinguish between an individualized and a subpopulation-based setting. For the post-recourse predictors, we will translate improvement guarantees into acceptance guarantees in Section \ref{sec:acceptance-guarantees}.
\paragraph{Individualized post-recourse prediction}{
\label{subsec:individualized-prediction}
%
% decision model and how to construct it from knowledge of the SCM
If we were to use the optimal pre-recourse observational predictor $h^*$ for post-recourse prediction, there would be an imbalance in predictive capability between ML model and individualized ICR:
ICR individualizes its predictions using $x^{pre}$ and the SCM. This knowledge is not accessible by the predictor $h^*$, which only makes use of $x^{post}$. As such, improvement that was accurately predicted by ICR is not necessarily recognized by $h^*$ and $\gamma^{ind}$ cannot be directly translated into an acceptance bound.
We demonstrate the issue at an Example in \ref{appendix:details:imbalance}.\footnote{One may also argue that standard predictive models are not suitable since optimality of the predictor in the pre-recourse distribution does not necessarily imply optimality in interventional environments (as Example \ref{example:covid-admission}, \ref{appendix:details:negative-result} demonstrates). We can refute this criticism using Proposition \ref{proposition:all-observed}, where we learn that $\hat{h}^*$ is stable with respect to ICR actions.}\\
In order to settle the imbalance between ICR and the predictor, we suggest to leverage the SCM not only when generating individualized ICR recommendations but also when predicting post-recourse, 
such that the predictor is at least as accurate as $\gamma^{ind}$.
More formally, we suggest to estimate the post-recourse distribution of $Y$ conditional on $x^{pre}$, $do(a)$, and the post-recourse observation $x^{post,a}$ (Definition \ref{def:individualized-post-recourse-prediction}).
This post-recourse prediction resembles the counterfactual distribution, except that we additionally take the factual post-recourse observation of the covariates into account.
\begin{definition}[Individualized post-recourse predictor]
We define the individualized post-recourse predictor as
\[h^{*, ind}(x^{post}) = P(Y^{post}=1|x^{post}, x^{pre}, do(a))\]
\label{def:individualized-post-recourse-prediction}
\end{definition}
For SCMs with invertible equations, $h^{*,ind}$ can be estimated using a closed form solution. Otherwise we can sample from the counterfactual post-recourse distribution $p(y^{post}, x^{post}| x^{pre}, do(a))$ (as we did for the estimation of $\gamma^{ind}$), select the samples that conform with $x^{post}$ and compute the proportion of favorable outcomes (details in \ref{appendix:estimation:individualized-post-recourse-predictor}).\\
For the individualized post-recourse predictor, improvement probability and prediction are closely linked (Proposition \ref{prop:ind-post-recourse:link-gamma}). More specifically, the expected post-recourse prediction $h^{*,ind}$ is equal to the individualized improvement probability $\gamma(x^{pre},a)$. We will exploit Proposition \ref{prop:ind-post-recourse:link-gamma} in Section \ref{subsec:mcr:acceptance-guarantees}, where we derive acceptance guarantees for ICR.
\begin{proposition}
The expected individualized post-recourse score is equal to the individualized improvement probability $\gamma^{ind}(x^{pre},a) := P(Y^{post}=1|x^{pre}, do(a))$, i.e.
\begin{equation*}
E[\hat{h}^{*, ind}(x^{post})|x^{pre}, do(a)] = \gamma^{ind}(a).
\end{equation*}
\label{prop:ind-post-recourse:link-gamma}
\end{proposition}
}
\paragraph{Subpopulation-based post-recourse prediction}{
\label{subsec:mcr:subpopulation:acceptance-guarantees}
%
% what about the prediction model: does it accept the model's decision? how does the previous goal align with modeling goal?
%In order to enable recourse guarantees, the decision model must honor the made improvements and therefore predict accurately despite the distribution shift induced by recourse.
%
% it may be more difficult in interventional environments
Curiously we find that for ICR actions $a$ the optimal observational pre-recourse predictor $h^*$ remains accurate: in the subpopulation of similar individuals the expected post-recourse prediction corresponds to the improvement probability $\gamma^{sub}(a)$ (Proposition \ref{proposition:all-observed}). This allows us to derive acceptance guarantees for $h^*$ in Section \ref{sec:acceptance-guarantees}.\\
%
% however it is robust with respect to intervention on causes
This result is in contrast to the negative results for CR, where actions may not affect prediction and the underlying target coherently, such that the predictive performance deteriorates (as demonstrated in the introduction, and more formally in \ref{appendix:details:negative-result}). %The reason is that optimal observational predictors are stable w.r.t ICR interventions, but may not be stable w.r.t. CR actions.
The key difference to CR is that ICR actions exclusively intervene on causes of $Y$:
Interventions on non-causal variables may lead to a shift in the conditional distribution $P(Y|X_S)$ (where $S \subseteq D$ is any set of variables that allows for optimal prediction). In contrast, given causal sufficiency, the conditional $P(Y|X_S)$ is stable to interventions on causes of $Y$. %As Proposition \ref{proposition:all-observed} demonstrates, the expected prediction and the subgroup-based improvement probability coincide. 
\begin{proposition}
Given nonzero cost for all interventions, ICR exclusively suggests actions on causes of $Y$. Assuming causal sufficiency, for optimal models the conditional distribution of $Y$ given the variables $X_S$ that the model uses (i.e. $P(Y|X_S)$) is stable w.r.t interventions on causes. Therefore, optimal predictors are intervention stable w.r.t. ICR actions.%\\
%\normalfont{Proof (sketch):} Any optimal predictor relies on (a superset of) the so-called markov blanket. Any superset of the markov blanket is stable w.r.t. interventions on causes, such that every optimal predictor is stable w.r.t. ICR actions. The full proof is given in \ref{proof:proposition:all-observed}.
\label{prop:causes-only}
\end{proposition}
\begin{proposition}
Given causal sufficiency and positivity\footnote{Positivity ensures that the post-recourse observation lies within the observational support \citep{neal2020introduction}, where the model was trained (i.e., $p^{pre}(x^{post}) > 0)$).}, %any set $S$ that allows for an optimal prediction (i.e., $MB(Y) \subseteq S$) is stable with respect to ICR actions.
%As a consequence, 
for interventions on causes
the expected subgroup-wide optimal score $h^*$ is equal to the subgroup-wide improvement probability $\gamma^{sub}(a) := P(Y^{post} = 1|do(a), x^{pre}_{G_{a}})$, i.e.
\[E[\hat{h}^{*}(x^{post})|x^{pre}_{G_{a}}, do(a)] = \gamma^{sub}(a).\]
\label{proposition:all-observed}
\end{proposition}
\textit{Link between CR and ICR}: Proposition \ref{prop:causes-only} has further interesting consequences. For CR actions $a$ that only intervene on causes of $Y$ and that are guaranteed to yield a predicted score $\zeta$ in the subpopulation, we can infer that $\gamma^{sub}(a) \geq \zeta$. For instance, if acceptance with respect to a $0.5$ decision threshold can be guaranteed, that implies improvement with at least $50\%$ probability. As such, in subpopulation-based settings (1) improvement guarantees can be made for CR if only interventions on causes are lucrative, and (2) CR can be adapted to also guide towards improvement by a restricting actions to intervene on causes.
%
%We will use the property to link sICR to sCR (Section \ref{subsubsec:mcr:subpopulation:links-cr}) and to derive acceptance guarantees (Section \ref{subsec:mcr:acceptance-guarantees}).
}
\section{Acceptance Guarantees}
\label{sec:acceptance-guarantees}

%\subsection{Acceptance guarantees}
\label{subsec:mcr:acceptance-guarantees}
%\subsubsection{Acceptance guarantees:}
%\label{subsec:mcr:individualized:acceptance-rates}
%
For the presented accurate post-recourse predictors, improvement guarantees translate into acceptance guarantees (Proposition \ref{prop:acceptance-bound}). The reason is that the post-recourse prediction is linked to $\gamma$ (Propositions \ref{prop:ind-post-recourse:link-gamma} and \ref{proposition:all-observed}).

\begin{proposition}
%Let $x_S^{pre}$ be the subset of the pre-recourse observed variables that were taken into account to compute ICR, i.e. $S=D$ for individualized recourse and $S=G_{a'}$ for subpopulation-based recourse.
%For subpopulation-based ICR, we furthermore assume that $p^{pre}(x^{post}) > 0$.
Let $g$ be a predictor with $E[g(x^{post}) | x_S^{pre}, do(a)] = \gamma(x_S^{pre}, a)$.
%For pre-recourse state $x^{pre}$, a $\gamma$-confident action $a$, the (individualized) optimal predictor $h^{*}$ and 
Then for a decision threshold $t$ the post-recourse acceptance probability $\eta(t; x^{pre}_S, a) := P(g(x^{post}) > t|x^{pre}_S, do(a))$ is lower bounded by the respective improvement probability:
%
%$$\eta(t; x^{pre}_S, a) = P(\hat{h}^{*}(x^{post,a}) > t) = \left. \frac{\gamma - FNR(t)}{TPR(t) - FNR(t)} \right|_{x^{pre}_S, a},$$
%
%Furthermore, since for any threshold $t$ $0 \leq FNR \leq t$ and $1 \geq TPR \geq t$, it holds that
%
\[ \eta(t; x^{pre}_S, a, g) \geq \frac{\gamma(x^{pre}_S,a) - t}{1 - t}.\]
%
%where $\gamma(x^{pre}_S, a) = p(Y^{post,a}=1|x^{pre}_S, a)$. 
%
\normalfont{Proof (sketch):} We decompose the expected prediction ($\gamma$) into true positive rate (TPR), false negative rate (FNR) and acceptance rate. By bounding TPR and FNR we yield the presented acceptance bound. The proof is provided in \ref{appendix:proof:acceptance-bound}.
\label{prop:acceptance-bound}
\end{proposition}

Using Proposition \ref{prop:acceptance-bound}, we can tune confidence $\gamma$ and the model's decision threshold to yield a desired acceptance rate. For instance, we can guarantee acceptance with (subgroup-wide) probability $\eta \geq 0.9$ given $\gamma = 0.95$ and a global decision threshold $t = 0.5$ .\\
Furthermore we can leverage the sampling procedures that we use to compute $\gamma$ to estimate the individualized or subpopulation-based acceptance rate $\eta(t; x^{pre}_S, a, g)$ (as detailed in \ref{appendix:estimation:individualized} and \ref{appendix:estimation:sampling-sub}). To guarantee acceptance with certainty, the decision threshold can be set to $t=0$.\\
For the explainee, it is vital that the acceptance guarantee is presented in a human-intelligible fashion. In contrast to previous work in the field, we suggest to communicate the acceptance guarantee in terms of a probability.\footnote{For CR, the acceptance confidence is encoded in a hyperparameter, as explained in \ref{appendix:details:expectation-objective}.}
Furthermore, for subpopulation-based recourse, the set of subgroup characteristics should be transparent. In the hospital admission example, the subpopulation-based acceptance guarantee could be communicated as follows: \textit{Within a group of individuals that share your zip code, a vaccination leads to acceptance with at least probability $\eta$.}

\section{Experiments}
\label{sec:simulation}
%
%\caption{%Observed improvement rate $\gamma^{obs}$, acceptance rate $\eta^{obs}$, mean acceptance rate on five refits $\eta^{obs, refit}$, and cost of individuals implementing CE, CR and ICR recommendations on 3var-causal (blue), 3var-noncausal (orange), 7var-covid (green). 
%Experimental results for CE, CR and ICR on four datasets over $10$ runs on $200$ individuals each.
%For the probabilistic methods the confidences $0.75, 0.85, 0.9, 0.95$ were targeted (for CR: $\overline{\eta}$, for ICR: $\overline{\gamma}$). For CE no slack is allowed, such that the results correspond to a confidence level of $1.0$. Values are reported on a quadratic scale.}
%\label{fig:experiment-results}
%\end{figure*}
%
In the experiments we evaluate the following questions, assuming correct causal knowledge and accurate models of the conditional distributions in the data:\\
\\
%\footnote{Further aspects of recourse like the robustness of recourse to refits on pre- and post-recourse data are assessed in \ref{appendix:additional-experiments}.}
%
% more specifically, we investigate the following questions
% Q1: Is ICR more meaningful that CR? I.e. does it lead to improvement, where CR does not lead to improvement?
\textit{Q1:} Do CE, CR and ICR lead to improvement?\\
%
% Q2: Given the suggested post-recourse predictors, does acceptance ensue from improvement? What about normal predictor?
\textit{Q2:} Do CE, CR and ICR lead to acceptance (by pre- and post- post-recourse predictor)?\\ %More specifically, given the suggested post-recourse predictors, do the proposed acceptance guarantees hold?\\
\textit{Q3:} Do CE, CR and ICR lead to acceptance by other predictors with comparable test error?\footnote{The problem that refits on the same data with similar performance have different mechanism is known as the Rashomon problem or model multiplicity \citep{Breiman2001cultures,pawelczyk_counterfactual_2020,marx_predictive_2020}.}\\
% Q3: What is the cost
\textit{Q4:} How costly are CE, CR and ICR recommendations?
\paragraph{Setup} We evaluate CE, individualized and subpopulation-based CR and ICR with various confidence levels, over multiple runs, and on multiple synthetic and semi-synthetic datasets with known ground-truth (listed below).\footnote{For ground-truth counterfactuals, simulations are necessary \citep{holland1986statistics}.} %More specifically, we evaluate the methods on the problems listed below. 
%The SCMs include continuous and categorical variables, linear an
%non-linear relationships, as well as various noise distributions.
Random forests were used for prediction, except in the \textit{3var} settings where logistic regression models were used. Following \citet{dandl_multi-objective_2020}, we use NSGA-II \citep{deb2002fast} for optimization.
For a full specification of the SCMs including the linear cost functions we refer to \ref{appendix:datasets}. Details on the implementation and access to the code are provided in \ref{appendix:implementation-details}.\\
%For full descriptions of the SCMs, implementation details and user-friendly code we refer to \ref{appendix:implementation-details} and \ref{appendix:datasets}. 
\\
%\begin{myident}
\textit{3var-causal:} A linear gaussian SCM with binary target $Y$, where all features are causes of $Y$.\\
\textit{3var-noncausal:} The same setup as \textit{3var-causal}, except that one of the features is an effect of $Y$. \\
%\textbf{5var-nonlinear:} A SCM with five continuous variables with nonlinear relationships and causes, effects and spouses of $Y$.\\
\textit{5var-skill:} A categorical semi-synthetic SCM where programming skill-level is predicted from causes (e.g. \textit{university degree}) and non-causal indicators extracted from GitHub (e.g. \textit{commit count}).\\
\textit{7var-covid:} A semi-synthetic dataset inspired by a real-world covid screening model \citep{jehi2020individualizing,wynants2020prediction}.\footnote{The real-world screening model is used to decide whether individuals need a test certificate to enter a hospital. It can be accessed via \url{https://riskcalc.org/COVID19/}.} The model includes typical causes like \textit{covid vaccination} or \textit{population density} and symptoms like \textit{fever} and \textit{fatigue}. The variables are mixed categorical and continuous with various noise distributions. Their relationships include nonlinear structural equations.
%\end{myident}
%
%\paragraph{Methods}{
%We run CR and ICR with various confidence levels and over multiple runs. 
%Over $3$ runs on $200$ recourse seeking individuals each, 
%We report observed improvement rate $\gamma^{obs}$, the acceptance rate $\eta^{obs}$, recourse cost, and mean acceptance rate for refits on the same training data $\eta^{obs, refit}$ in Figure \ref{}. %All the aforementioned metrics are computed for those individuals where a recourse recommendation could be made. The percentage of individuals, where a recourse recommendation could be made, is denoted as $\zeta$.
%
%}
%
\paragraph{Results}{The results are visualized in Figure \ref{fig:experiment-results} and provided in tabular form in \ref{appendix:additional-experiments}. %For each setting CE, CR and ICR explanations were computed over $10$ runs on $200$ individuals each. For CR and ICR the confidences $0.75, 0.85, 0.9, 0.95$ were targeted (for CR: $\overline{\eta}$, for ICR: $\overline{\gamma}$). For CE no slack is allowed, such that the results correspond to a confidence level of $1.0$. Values are plotted on quadratic scales.
\begin{figure}[t]
\centering
\subcaptionbox{Observed improvement rates $\gamma^{obs}$ (Q1).\label{subfig:improvement}}[0.4\textwidth]{
    \centering
  \includegraphics[width=.99\linewidth]{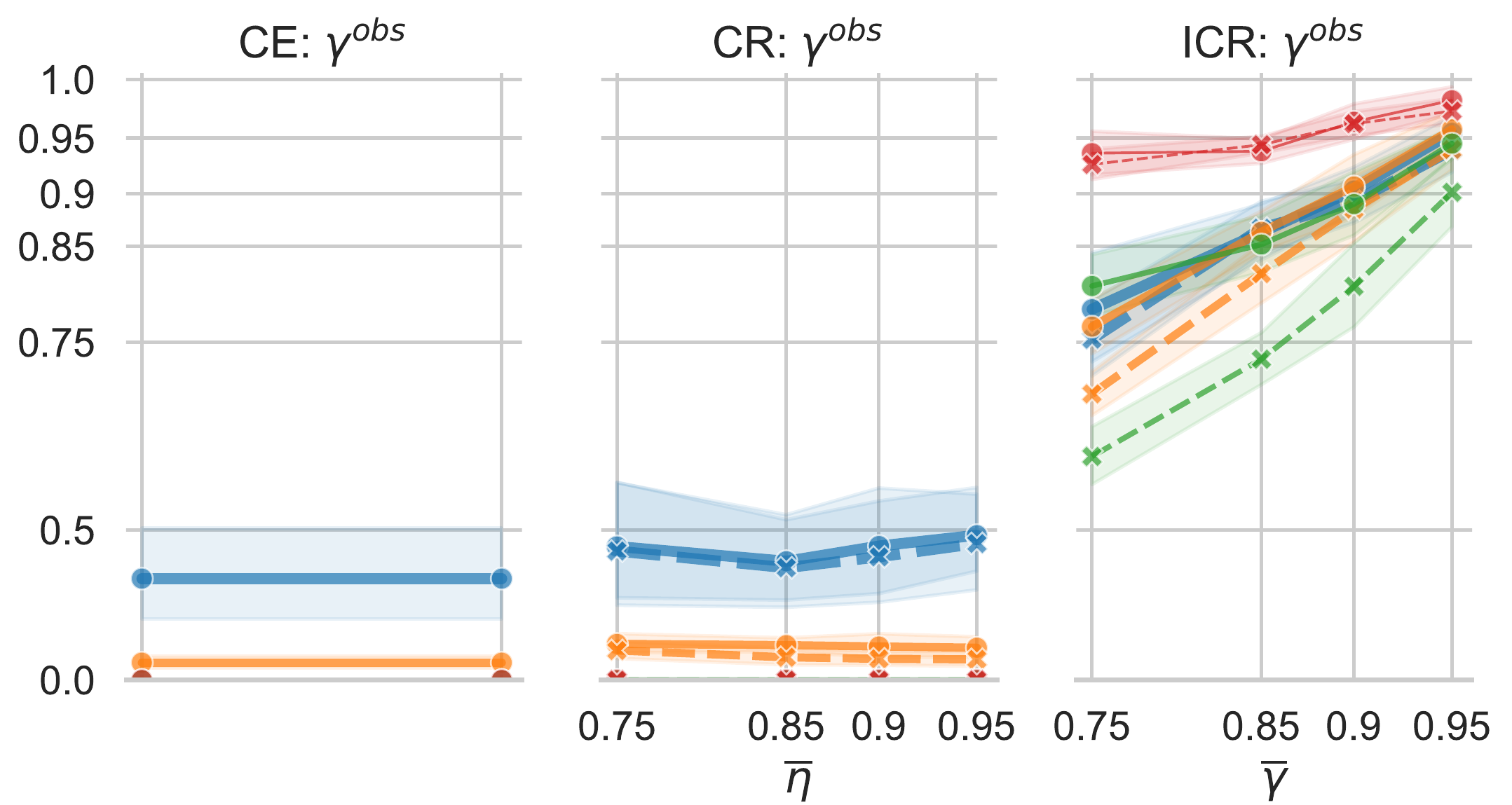}
}
\hfill
\subcaptionbox{causal graphs\label{subfig:graphs}}[0.20\textwidth]{
\centering
  \begin{tikzpicture}[thick, scale=.3, every node/.style={scale=.65, line width=0.25mm, black, fill=white}]
    \usetikzlibrary{shapes}
    
        \node[fill=RoyalBlue, ellipse, scale=0.9] (a1) at (-4, 2) {};
        \node[fill=RoyalBlue, ellipse, scale=0.9] (a2) at (-1, 2) {};
        \node[fill=RoyalBlue, ellipse, scale=0.9] (a3) at (-2.5, 1) {};
        \node[fill=black, ellipse, scale=0.9] (ay) at (-2.5, -0.5) {};
        
        \draw[-stealth, gray, scale=0.3] (a1) -- (a2);
        \draw[-stealth, gray, scale=0.3] (a1) -- (a3);
        \draw[-stealth, gray, scale=0.3] (a1) -- (ay);
        \draw[-stealth, gray, scale=0.3] (a2) -- (a3);
        \draw[-stealth, gray, scale=0.3] (a2) -- (ay);
        \draw[-stealth, gray, scale=0.3] (a3) -- (ay);

        \node[fill=orange, ellipse, scale=0.9] (b1) at (3, 2) {};
        \node[fill=orange, ellipse, scale=0.9] (b2) at (0, 2) {};
        \node[fill=black, ellipse, scale=0.9] (b3) at (1.5, 1) {};
        \node[fill=orange, ellipse, scale=0.9] (by) at (1.5, -0.5) {};
        
        \draw[-stealth, gray, scale=0.3] (b1) -- (b2);
        \draw[-stealth, gray, scale=0.3] (b1) -- (b3);
        \draw[-stealth, gray, scale=0.3] (b1) -- (by);
        \draw[-stealth, gray, scale=0.3] (b2) -- (b3);
        \draw[-stealth, gray, scale=0.3] (b2) -- (by);
        \draw[-stealth, gray, scale=0.3] (b3) -- (by);
        
        \node[fill=Green, ellipse, scale=0.9] (c1) at (3, -3) {};
        \node[fill=Green, ellipse, scale=0.9] (c2) at (2, -3) {};
        \node[fill=Green, ellipse, scale=0.9] (c3) at (1, -3) {};
        \node[fill=Green, ellipse, scale=0.9] (c4) at (2, -3) {};
        \node[fill=black, ellipse, scale=0.9] (cy) at (1.5, -4.5) {};
        \node[fill=Green, ellipse, scale=0.9] (c5) at (2.5, -6) {};
        \node[fill=Green, ellipse, scale=0.9] (c6) at (1.5, -6) {};
        \node[fill=Green, ellipse, scale=0.9] (c7) at (0.5, -6) {};
        
        \draw[-stealth, gray, scale=0.3] (c1) -- (cy);
        \draw[-stealth, gray, scale=0.3] (c2) -- (cy);
        \draw[-stealth, gray, scale=0.3] (c3) -- (cy);
        \draw[-stealth, gray, scale=0.3] (c4) -- (cy);
        \draw[-stealth, gray, scale=0.3] (c1) -- (c5);
        \draw[-stealth, gray, scale=0.3] (cy) -- (c5);
        \draw[-stealth, gray, scale=0.3] (cy) -- (c6);
        \draw[-stealth, gray, scale=0.3] (cy) -- (c7);

        \node[fill=red, ellipse, scale=0.9] (d1) at (-3.5, -3) {};
        \node[fill=red, ellipse, scale=0.9] (d2) at (-1.5, -3) {};
        \node[fill=black, ellipse, scale=0.9] (dy) at (-2.5, -4.5) {};
        \node[fill=red, ellipse, scale=0.9] (d3) at (-4, -6) {};
        \node[fill=red, ellipse, scale=0.9] (d4) at (-2.5, -6) {};
        \node[fill=red, ellipse, scale=0.9] (d5) at (-1, -6) {};
        
        \draw[-stealth, gray, scale=0.3] (d1) -- (dy);
        \draw[-stealth, gray, scale=0.3] (d2) -- (dy);
        \draw[-stealth, gray, scale=0.3] (dy) -- (d3);
        \draw[-stealth, gray, scale=0.3] (dy) -- (d4);
        \draw[-stealth, gray, scale=0.3] (dy) -- (d5);
        \draw[-stealth, gray, scale=0.3] (d1) -- (d3);
    \end{tikzpicture}
  \vspace{0.4cm}
}
\hfill
\subcaptionbox{Recourse cost (Q4).}[0.3\textwidth]{
\centering
%\begin{table}[]
%      \centering
      \begin{tabular}{c c}
           \toprule
           method & cost  \\
           \midrule
           CE & 1.82 $\pm$ 1.09  \\
           ind. CR & 1.34  $\pm$  1.14\\
           subp. CR & 1.65 $\pm$ 1.02\\
        ind. ICR & 4.26 $\pm$ 3.34\\
          subp. ICR & 4.20 $\pm$ 3.33\\
          \bottomrule
    \end{tabular}
%    \caption{Caption}
%  \label{tab:my_label}
%\end{table}
%\small
%      \begin{tabular}{r r r r r}
%        
%           \toprule
%         CE & ind. CR & sub. CR & ind. ICR & sub. ICR  \\
%           \midrule
%           %1.82 $\pm$ 1.09 & 1.34  $\pm$  1.14 &1.65 $\pm$ 1.0 & 4.26 $\pm$ 3.34 & 4.20 $\pm$ 3.33\\
%            1.8 $\pm$ 1.1 & 1.3  $\pm$  1.1 &1.7 $\pm$ 1.0 & 4.3 $\pm$ 3.3 & 4.2 $\pm$ 3.3\\
%           \bottomrule
%      \end{tabular}

\vspace{0.5cm}
}

\bigskip
%\hfill
\subcaptionbox{Observed acceptance rates $\eta^{obs}$ w.r.t. $h^*$; for ind. ICR additionally w.r.t. $h^{*,ind}$ (Q2).  \label{subfig:acceptance}}[0.4\textwidth]{
%\begin{subfigure}{0.43\textwidth}
  %\centering
  \includegraphics[width=.99\linewidth]{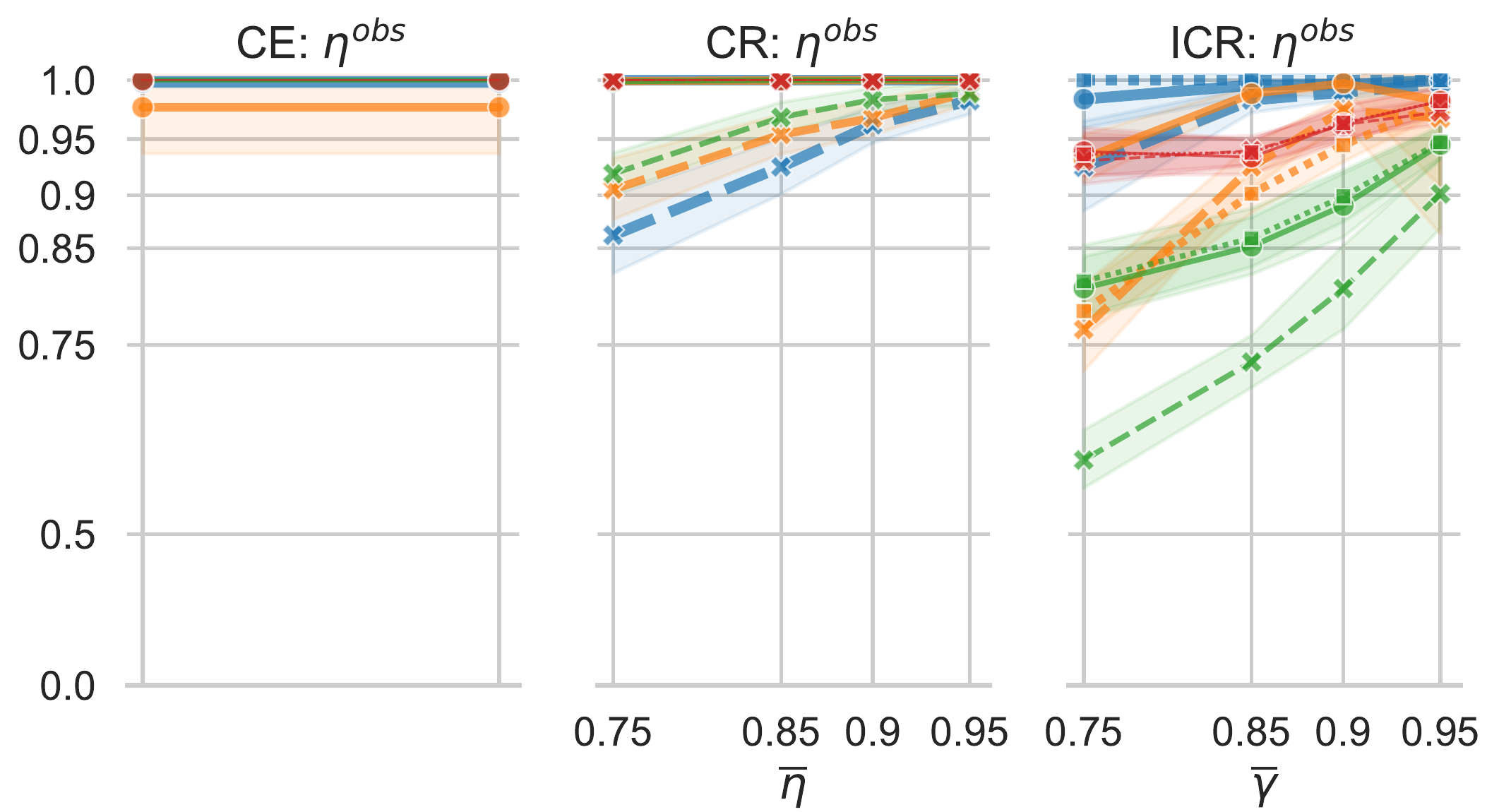}
 % \caption{observed acceptance rates $\eta^{obs}$ w.r.t. the pre-recourse predictor, for ind. ICR additionally w.r.t. the ind. post-recourse predictor (squared markers) (Q2)}
%\end{subfigure}
}
\hfill
\subcaptionbox{Observed acceptance rates for other fits with comparable test set performance $\eta^{obs, \text{refit}}$ (Q3).\label{subfig:acceptance-refits}}[0.4\textwidth]{
%\begin{subfigure}{0.43\textwidth}
  %\centering
  \includegraphics[width=.99\linewidth]{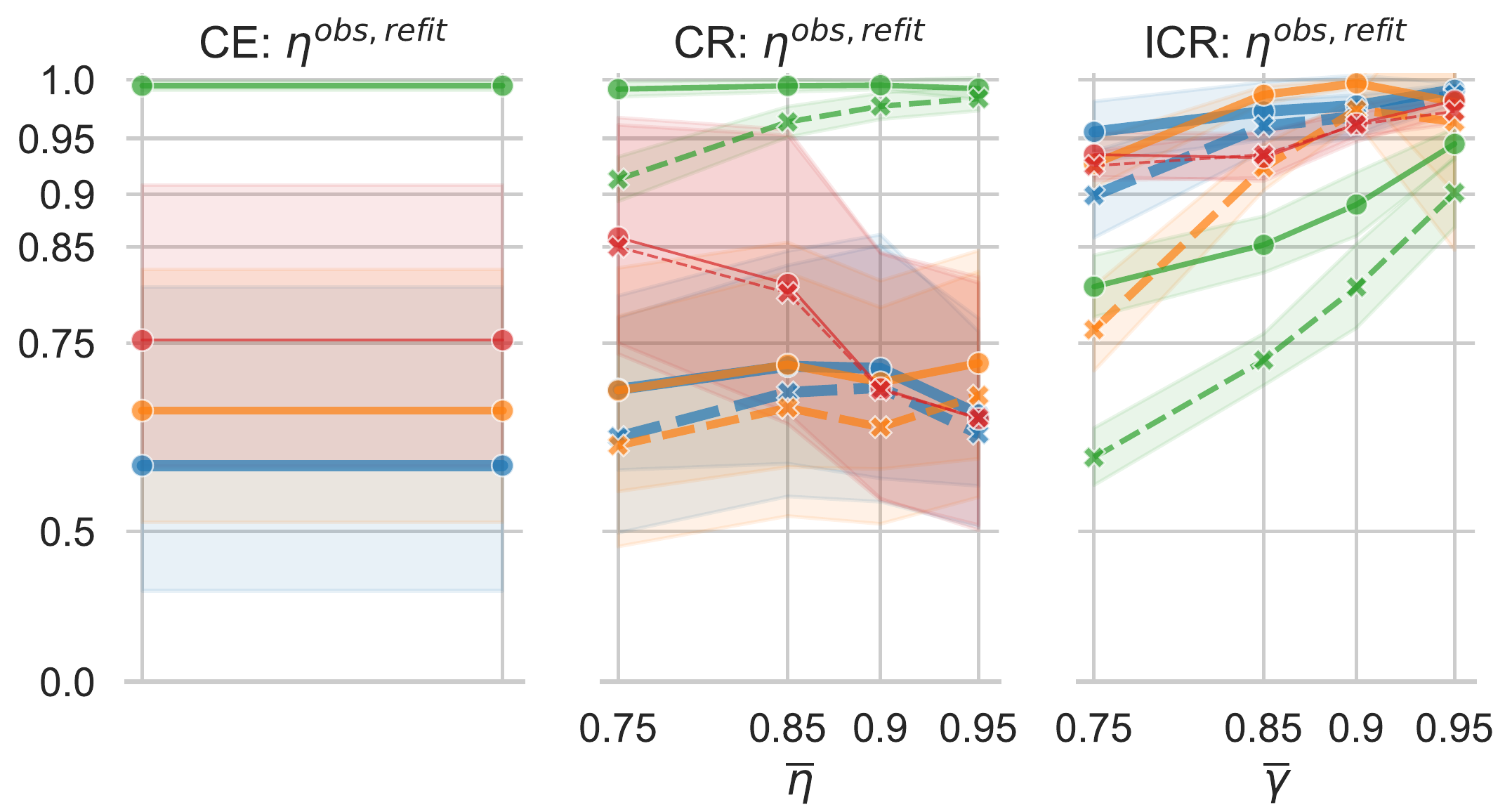}
  %\caption{observed acceptance rates for other fits on the same data with comparable test set performance $\eta^{obs, \text{refit}}$ (Q3)}
%\end{subfigure}
}
\hfill
\subcaptionbox{legend\label{subfig:legend}}[0.15\textwidth]{
%\begin{subfigure}{0.12\textwidth}
  %\subcaptionbox{legend}[0.2\textwidth]{
    \centering
  \includegraphics[width=.99\linewidth]{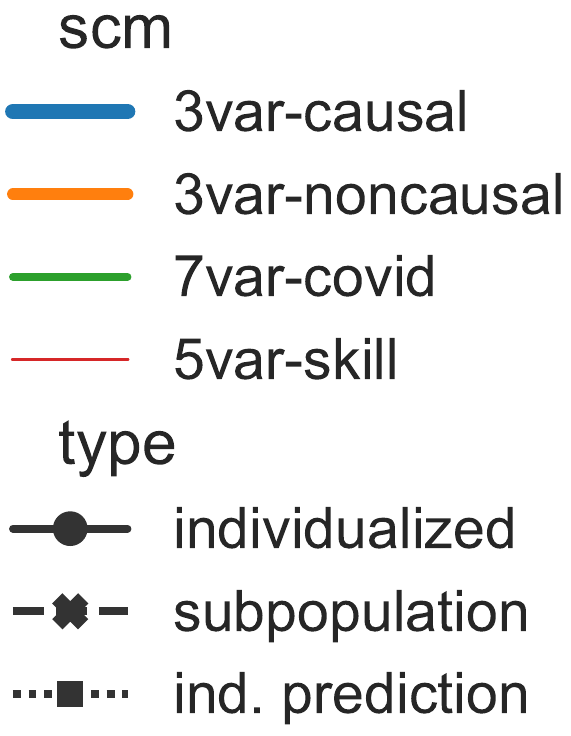}
 % \caption{legend}
  \vspace{0.025cm}
%}
%\end{subfigure}
}
%\hfill
\caption{%Observed improvement rate $\gamma^{obs}$, acceptance rate $\eta^{obs}$, mean acceptance rate on five refits $\eta^{obs, refit}$, and cost of individuals implementing CE, CR and ICR recommendations on 3var-causal (blue), 3var-noncausal (orange), 7var-covid (green). 
Experimental results for CE, CR and ICR on four datasets over $10$ runs on $200$ individuals each.
For the probabilistic methods the confidences $0.75, 0.85, 0.9, 0.95$ were targeted (for CR: $\overline{\eta}$, for ICR: $\overline{\gamma}$). For CE no slack is allowed, such that the results correspond to a confidence level of $1.0$. Values are reported on a quadratic scale.}
\label{fig:experiment-results}
\end{figure}
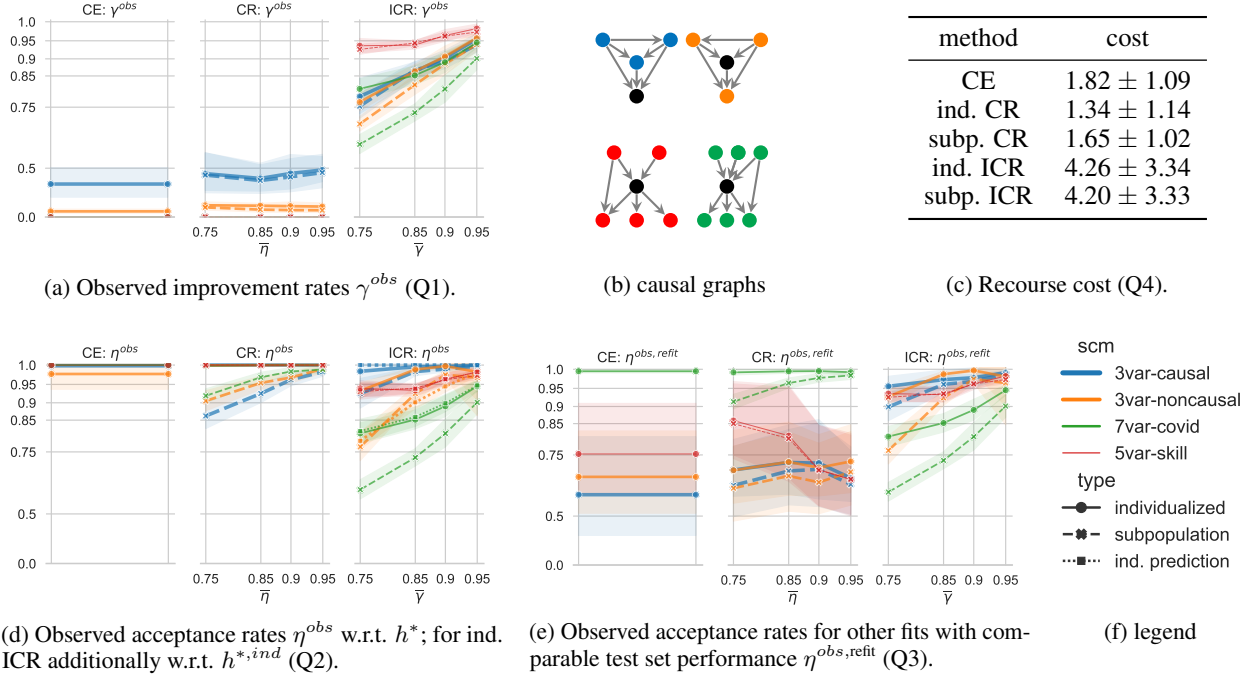

\textit{Q1 (Figure \ref{fig:experiment-results}a):} In scenarios where gaming is possible and lucrative (\textit{3var-noncausal}, \textit{5var-skill} and \textit{7var-covid}) ICR reliably guides towards improvement, but CE and CR game the predictor and yield improvement rates close to zero. For instance, on \textit{5var-skill} CE and CR exclusively suggest to tune the GitHub profile (e.g. by adding more commits). Since the employer offered recourse it should be honored although the applicants remain unqualified. In contrast, ICR suggests to get a degree or to gain experience, such that recourse implementing individuals are suited for the job.\\
On \textit{3var-causal}, where gaming is not possible, CR also achieves improvement. However, since acceptance w.r.t to a decision treshold $t=0.5$ is targeted, only improvement rates close to $50\%$ are achieved (the expected predicted score translates into $\gamma^{sub}$ (Proposition \ref{proposition:all-observed})).\\
For subp. ICR, $\gamma^{obs}$ is below $\overline{\gamma}$, because the subpopulation may include individuals that were already accepted pre-recourse, such that $\gamma^{sub}$ and $\gamma^{obs}$ may not coincide.

\textit{Q2 (Figure \ref{fig:experiment-results}d):} All methods yield the desired acceptance rates w.r.t.~to the pre-recourse predictor.\footnote{ICR holds the acceptance rates from Proposition \ref{prop:acceptance-bound}, as analyzed in more detail in \ref{appendix:additional-experiments}.} For CE and CR $\eta^{obs}$ is higher than for ICR, and for ind.~recourse higher than for subp.~recourse. Curiously, although no acceptance guarantees could be derived for the pre-recourse predictor and ind. ICR, we find that both pre- and ind.~post-recourse predictor reliably lead to acceptance.\footnote{Given that the ind. post-recourse predictor is much more difficult to estimate, the pre-recourse predictor in combination with individualized acceptance guarantees (\ref{appendix:estimation:individualized}) may cautiously be used as fallback.}

\textit{Q3 (Figure \ref{fig:experiment-results}e):} 
We observe that CE and CR actions are unlikely to be honored by other model fits with similar performance on the same data. This result is highly relevant to practitioners, since models deployed in real-world scenarios are regularly refitted. As such, individuals that implemented acceptance-focused recourse may not be accepted after all, since the decision model was refitted in the meantime. 
In contrast, ICR acceptance rates are nearly unaffected by refits.
The result confirms our argument that improvement-focused recourse may be more desirable for explainees (Section \ref{sec:two-tales}).

\textit{Q4 (Table \ref{fig:experiment-results}c):} CR actions are cheaper than ICR actions, since improvement may require more effort than gaming. As such, CR has benefits for the explainee: For instance, on \textit{5var-skill}, CR suggests to tune the GitHub profile (e.g. by adding more commits), which requires less effort than earning a degree or gaining job experience. Detailed results on cost are reported in \ref{appendix:additional-experiments}.
% For instance, for (subpopulation-based) CR with confidence $0.95$ the average cost over all datasets and runs was $1.53 \pm 1.18$ ($1.73 \pm 1.16$). For ICR the corresponding (subpopulation-based) cost was $1.92 \pm 0.95$ ($2.05 \pm 0.84$). CEs were on par with ICR with cost $2.05 \pm 1.08$. CEs are more expensive than CR since they ignore causal relationships. 
}

In conclusion, ICR actions require more effort than CR, but lead to improvement and acceptance while being more robust to refits of the model.
\section{Limitations and Discussion}
\label{sec:limitations-discussion}
\paragraph{Causal knowledge and assumptions}{
Individualized ICR requires a fully specified SCM; Subpopulation-based ICR is less demanding but still requires the causal graph and causal sufficiency. SCMs and causal graphs are rarely readily available in practice \citep{Peters2017book} and causal sufficiency is difficult to test \citep{janzing2011detecting}. 
Research on causal inference gives reason for cautious optimism that the difficulties in constructing SCMs and causal graphs can eventually be overcome \citep{spirtes2016causal,Peters2017book,heinze2018causal,malinsky2018causal,glymour2019review}.\\
There are further foundational problems linked to causality that affect our approach: causal cycles, an ontologically vague target $Y$ (e.g. in hiring), disparities in our data, or causal model misspecification \citep{barocas2016big,barocas2017fairness,bongers2021foundations}. All of these factors are considered difficult open problems and may have detrimental impact on our, as well as on any other, recourse framework.\\
Guiding action without causal knowledge is impossible; when causal knowledge is available, our work provides a normative framework for improvement-focused recourse recommendations. Thus, we join a range of work in explainability \citep{frye2020asymmetric,heskes2020causal,wang2021shapley,zhao2021causal} and fairness \citep{kilbertus2017avoiding,kusner2017counterfactual,zhang_fairness_2018,makhlouf2020survey} that highlights the importance of causal knowledge.
}
%Individualized ICR requires a fully specified SCM.  Subpopulation-based ICR relaxes the assumptions of inICR to requiring knowledge of the causal graph and causal sufficiency. However, both SCM and causal graph are hard to identify \citep{Peters2017book} and causal sufficiency is difficult to test \citep{janzing2011detecting}. As such, sufficient prior knowledge is not always available.\\
%
%Since guiding action without causal assumptions is impossible, causal inference is an active area of research \citep{Peters2017book,heinze2018causal,malinsky2018causal,glymour2019review,spirtes2016causal}.
%We join a broad range of work in explainability \citep{heskes2020causal,frye2020asymmetric,zhao2021causal,wang2021shapley} and fairness \citep{kusner2017counterfactual,zhang_fairness_2018,kilbertus2017axvoiding,makhlouf2020survey} that highlights the importance of causal knowledge.
%
%\paragraph{Compliance}{
%In our work, we assume that recourse seeking individuals comply with the suggested actions and/or %that the individual's compliance can be assessed. TODO discuss this assumption.
%}
\paragraph{Contestability}{Improvement-focused recourse guides individuals towards actions that help them to improve, e.g., it recommends a vaccination to lower the risk to get infected with Covid. If, however, a explainee is more interested in contesting the algorithmic decision, (improvement-focused) recourse recommendations are not sufficient.
Think of an individual who is denied entrance to an event because of their high Covid risk prediction, which is based on a non-causal, spurious association with their country of origin\footnote{E.g., due to a spurious association with the causal variable \textit{type of vaccine}.}.
In such situations, we suggest to additionally show explainees diverse explanations, which %reflect the model's prediction mechanism and therefore 
enable to contest the decision. For example, such an explanation could be: if your country of origin would be different, your predicted Covid risk would have been lower.}
\section{Conclusion}

In the present paper, we took a causal perspective and investigated the effect of recourse recommendations on the underlying target variable.
We demonstrated that acceptance-focused recourse recommendations like counterfactual explanations or causal recourse may not improve the underlying prediction but game the predictor instead. The problem stems from predictive, but non-causal relationships, which are abundant in machine learning applications.\footnote{For instance, in hiring, certain keywords in the CV may be associated with qualification, but adding them to the CV does not improve aptitude \citep{strong_mit_nodate}.}\\
We tackled the problem in the explanation domain and introduced Improvement-Focused Causal Recourse (ICR), an explanation technique that guides towards improvement of the prediction target and demonstrated how to design post-recourse predictors such that improvement leads to acceptance. We confirm the theoretical results in experiments.\\
With ICR we hope to inspire a shift from acceptance- to improvement-focused recourse. 

\section*{Acknowledgements}

This work was supported by the Graduate School of Systemic Neuroscience (GSN) of the LMU Munich and by the German Federal Ministry of Education and Research (BMBF).

\newpage
\bibliographystyle{unsrtnat}
%\bibliography{references} 

\newpage
\appendix

\begin{table*}[b]
    \centering
    \caption{Overview of important terms and their meanings.}
    \begin{tabular}{lcl}
    \toprule
        term & & meaning \\
        \midrule
        explainee && individual for whom the explanation is generated, e.g. loan applicant \\
        model authority && decision-making entity, e.g. credit institute\\
        recourse && action of the explainee that reverts unfavorable decision\\
        acceptance && desirable model prediction ($\hat{Y} = 1$)\\
        improvement && (yield) desirable state of the underlying target ($Y = 1$)\\
        gaming && yield acceptance without improvement, e.g. treating the symptoms\\
        pre-/post-recourse && before/after implementing recourse recommendation\\
        contestability && the explainee's ability to contest an algorithmic decision\\
        robustness of recourse && probability that recourse is accepted despite model/data shifts\\
        \bottomrule
    \end{tabular}
    \label{tab:term-overview}
\end{table*}

\section{Extended Background}

As follows, we recapitulate well-known definitions in our notation, provide more detailed background on related work and recapitulate results that we use in the proofs. Readers who are already familiar with recourse terminology and $d$-separation (\ref{appendix:extended-background:terms} and \ref{appendix:d-separation}), and who are not interested in more detailed introductions of intervention stability (\ref{sec:background-stability}, only required for the proof of Proposition \ref{prop:causes-only}) or causal recourse (\ref{sec:background-recourse}), may skip this section.

\subsection{Overview of important terms}
\label{appendix:extended-background:terms}

An overview of important terms is provided in Table \ref{tab:term-overview}.

\subsection{d-separation}
\label{appendix:d-separation}

Two variable sets $X, Y$ are called $d$-separated \citep{geiger1990identifying,Spirtes2001} by the variable set $Z$ in a graph $\mathcal{G}$ (denoted as $X \idp_{\mathcal{G}} Y | Z$), if, and only if, for every path $p$ it either holds that (i) $p$ contains a chain $i \rightarrow m \rightarrow j$ or a fork $i \leftarrow m \rightarrow j$ where $m \in Z$ or (ii) $p$ contains a collider $i \rightarrow m \leftarrow j$ such that $m$ and for all of its descendants $n$ it holds that $m, n \not \in Z$. Given the causal Markov property, $d$-separation in a causal graph implies (conditional) independence in the data \citep{Peters2017book}.

\subsection{Generalizability and intervention stability}
\label{sec:background-stability}
For Proposition \ref{prop:causes-only}, we leverage necessary conditions for invariant conditional distributions as derived in \citep{pfister_stabilizing_2019}. The authors introduce a $d$-separation based intervention stability criterion that is applied to a modified version of $\mathcal{G}$. For every intervened upon variable $X_l$ an auxiliary intervention variable, denoted as $I_l$, is added as direct cause of $X_l$, yielding $\mathcal{G}^*$. The intervention variable can be seen as a switch between different mechanisms. A set $S \subseteq \{1, \dots, d\}$ is called \textit{intervention stable} regarding a set of actions if for all intervened upon variables $X_l$ (where $l \in I^\text{total}$) the $d$-separation $I^l \idp_{\mathcal{G}^*} Y | X_S$ holds in $\mathcal{G}^*$. The authors show that intervention stability implies an invariant conditional distribution, i.e., for all actions $a, b \in \mathbb{A}$ with $I^a, I^b \subseteq I^{\text{total}}$ it holds that $p(y^a|x_S) = p(y^b|x_S)$ (\citet{pfister_stabilizing_2019}, Appendix A).

\subsection{Causal recourse}
\label{sec:background-recourse}

ICR is closely related to the CR framework \citep{karimi_algorithmic_2020, karimi_algorithmic_2021}, but differs substantially in its motivation and target. In order to allow for a direct comparison we briefly sketch the main ideas and the central CR definitions in our notation. %For a detailed introduction and discussion of the approach we refer the reader to \citet{karimi_algorithmic_2020}. 
Like ICR, CR aims to guide individuals to revert unfavorable algorithmic decisions (recourse). Therefore, they suggest to search for cost-efficient actions that lead to acceptance by the prediction model. Actions are modeled as structural interventions $a : \Pi \to \Pi$, which can be constructed as $a = do(\{X_i := \theta_i\}_{i \in I})$, where $I$ is the index set of features to be intervened upon \citep{karimi_algorithmic_2021}.
The conservativeness of the suggested actions can be adjusted using the hyperparameter $\gamma_{LCB}$, that determines the adaptive threshold $\texttt{thresh}(a)$ and thereby how many standard deviations the expected prediction shall be away from the model's decision threshold $t$.
In order to accommodate different levels of causal knowledge, two probabilistic versions of CR were introduced \citep{karimi_algorithmic_2020}: While individualized recourse assumes knowledge of the SCM, subpopulation-based CR only assumes knowledge of the causal graph. 

\paragraph{Individualized recourse}{ Individualized recourse predicts the effect of actions using structural counterfactuals \citep{karimi_algorithmic_2021}, which require a full specification of the SCM.

Given a function that evaluates the cost of actions ($\text{cost}(a, x^{pre})$), the optimization goal for individualized causal recourse is given below. The adaptive threshold $\texttt{thresh}$ bounds the prediction away from the decision threshold.\footnote{Further constraints have been suggested, e.g., $x^{post,a} \in \mathcal{P}\text{lausible}$ or $a \in \mathcal{F}\text{easible}$ \citep{laugel2019dangers,ustun_actionable_2019,mahajan2019preserving,dandl_multi-objective_2020,karimi_algorithmic_2021}.}

\begin{align*}
a^* \in \underset{a \in \mathbb{A}}{\text{argmin}} \quad  \text{cost}(a, x^{pre}) 
\quad &\text{s.t. } \E%_{X^{post,a}}
[\hat{h}(x^{post})|do(a), x^{pre}] \geq \texttt{thresh}(a)\\
\quad &\text{with } \texttt{thresh}(a) := 0.5 + \gamma_{LCB} \sqrt{\var%_{X^{post,a}}
[\hat{h}(x^{post,a})]}
%\label{eq:ar-individualized}
\end{align*}
}

\paragraph{Subpopulation-based recourse:}{ If no knowledge of the SCM is given, counterfactual distributions cannot be estimated and consequently individualized recourse recommendations cannot be computed. %Given knowledge of the causal graph $\mathcal{G}$, subpopulation-based recourse recommendations can be made. 
Subpopulation-based CR is based on the average treatment effect within a subgroup of similar individuals \citep{karimi_algorithmic_2020}. More specifically individuals belong to the same group if the non-descendants $nd(I)$ of intervention variables (which ceteris paribus remain constant despite the intervention) take the same value. %This grouping is designed to make use of as many characteristics as possible while still allowing observational identifiability of the effect.
The subpopulation-based objective is given below.
\begin{equation*}
%\begin{multline*}
a^* \in \underset{a \in \mathbb{A}}{\text{argmin}} \text{    cost}(a, x^{pre}) \text{  s.t. } 
\E_{X_{d(I)}|do(X_{I}=\theta), x^{pre}_{nd(I)}}[\hat{h}(x^{pre}_{nd(I)}, \theta, X_{d(I)})]
\geq \texttt{thresh}(a).
\label{eq:ar-subpopulation}
%\end{multline*}
\end{equation*}
}

\newpage
\section{Estimation and Optimization}
\label{appendix:estimation}

As follows we provide detailed explanations of the proposed estimation procedures.
First, we explain how to sample from the individualized post-recourse distribution, which allows us to estimate the individualized improvement and acceptance rates ($\gamma^{ind}$ and $\eta^{ind}$, \ref{appendix:estimation:individualized}). Based on the same sampling mechanism we can also estimate the individualized post-recourse prediction $h^{*,ind}$ (\ref{appendix:estimation:individualized-post-recourse-predictor}). Then we explain how to sample from the subpopulation-based post-recourse distribution, which allows us to estimate the subpopulation-based improvement and acceptance rates ($\gamma^{sub}$ and $\eta^{sub}$, \ref{appendix:estimation:sampling-sub}). Furthermore, we provide details on optimization (\ref{appendix:details:optimization}) and demonstrate that the optimal observational predictor $h^*$ can also be estimated using the SCM (\ref{appendix:estimation-optimal-pre-recourse}).

\subsection{Estimation of the individualized improvement confidence \texorpdfstring{$\gamma^{ind}$}{} and individualized acceptance rate \texorpdfstring{$\eta^{ind}$}{}}
\label{appendix:estimation:individualized}
We recall that $\gamma^{ind}$ is the counterfactual probability of the underlying target $Y$ taking the favorable outcome, and $\eta^{ind}$ the counterfactual probability of the prediction $\hat{Y}$ taking the favorable outcome. In order to estimate $\gamma^{ind}$ and $\eta^{ind}$ we first sample covariates and target from the counterfactual post-recourse distribution and then compute the proportion of favorable outcomes for $Y$ and $\hat{Y}$ in the sample.\\
In general, sampling from counterfactual distributions based on a SCM is performed in three steps (Section \ref{sec:background-notation}, \citep{Pearl2009}).
\begin{enumerate}
    \item \textit{Abduction}: The exogenous noise variables are reconstructed from the observations, i.e., $p(u_{Y, D}|x^{pre})$ is estimated.
    \item \textit{Intervention}: The intervention $do(a)$ on the SCM $\mathcal{M}$ is performed by replacing the respective structural equations $f_{I_a} := \theta_{I_a}$, yielding $\mathcal{M}_{do(a)}$.
    \item \textit{Prediction}: The abducted noise variables are sampled from $p(u_{Y, D}|x^{pre})$ and passed through the model $\mathcal{M}_{do(a)}$ to sample from the counterfactual distribution $P(Y^{post}, X^{post}|x^{pre}, do(a))$.
\end{enumerate}
%
%For details on the intervention and prediction steps, we refer to standard literature \citep{Pearl2009,Peters2017book}. 
Given knowledge of the SCM, the challenge is to sample the exogeneous variables from $p(u_{Y, D}|x^{pre})$ (abduction). As follows we explain the abduction in two steps. First, we explain how we can abduct $u_j$ for variables for which both the node $x_j$ and all parents $(x,y)_{pa(j)}$ are observed, which we refer to as the standard abduction case. Then we factorize the abduction of the joint $p(u_{Y, D}|x^{pre})$ into several components which can be reduced to said standard abduction case. The sampling procedure is summarized in Algorithm \ref{alg:sample-individualized-covariate}.

\subsubsection{Recap: Standard abduction}
If for a node $u_j$ both the node $(x,y)_j$ and the parents $(x,y)_pa(j)$ are observed, we can apply standard abduction. The standard abduction procedure depends on the type of structural equation and exogenous noise distribution.\\
Given invertible structural equations, observation of $x_j,x_{pa(j)}$ determines $u_j$. More specifically, $u_j$ can be reconstructed using
\begin{equation*}
    u_j = f^{-1}(x_j; x_{pa(j)}).
    %\label{eq:individualized-abduction-all-observed-invertible}
\end{equation*}
For instance, for additive structural equations $f_j(u_j; x_{pa(j)}) = g(x_{pa(j)}) + u_j$, the inversion is given by $f_j^{-1}(x_j; x_{pa(j)}) = x_j - g(x_{pa(j)})$.\\
In our experiments we also included binomial variables with a sigmoidal (non-invertible) structural equation. More specifically, the structural equations are defined as $x_j = [\sigma(l(x_{pa(j))}) \leq u_j]$ with $U_j \sim Unif(0,1)$. Here $\sigma$ refers to the sigmoid function and $l$ to some linear combination. $[cond]$ evaluates to $1$ when the condition is true and otherwise to $0$. Intuitively, $\sigma(l(x_{pa(j))})$ can be seen as a nonlinear activation function which determines the probability of the node being activated ($x_j=1$). $u_j$ acts as a dice, where values $\leq \sigma(l(x_{pa(j))})$ imply $x_j = 1$ and vice versa. \\
For those variables, if $x_j=1$, we know that $u_j \leq \sigma(l(x_{pa(j))})$ and vice versa, such that we can abduct $U_j$ as follows (and can therefore sample $u_j$):
\begin{equation*}
    P(U_j|x_j; x_{pa(j)}) = \left\{\begin{array}{lr}
        Unif(0, \sigma(l(x_{pa(j)}))), & \text{for } x_j = 1\\
        Unif(\sigma(l(x_{pa(j)})), 1), & \text{for } x_j = 0
        \end{array}\right.
    %\label{eq:individualized-abduction-all-observed-sigmoidal}
\end{equation*}
As we will see in the next section, our estimation procedure can be flexibly extended to SCMs with different types of structural equations, as long as a procedure to sample from the abducted exogneous noise variable for the standard case (where parents and the node itself are observed) is available.\\
\subsubsection{Factorization of \texorpdfstring{$p(u|x)$}{the abduction}}
We have demonstrated how to abduct individual nodes in the standard setting where the corresponding endogenous variable and its parents are observed.\\
As follows we demonstrate how to sample from the joint distribution of the exogenous variables given an observation of $X$ (and without observing $Y$). %Therefore we factorize the joint of the abducted exogneous variables into components that can be estimated with standard abduction. 
%Therefore, we divide $p(u|x)$ into two subproblems: estimating and sampling $y'$ from $p(y|x)$, and then sampling each exogeneous variable given $(x, y')$ using standard abduction.\\
Therefore, we show that $p(u|x)$ can be seen as a mixture of two distributions, one for each possible state $y'$ of $Y$. In order to sample from it, we (1) need to sample $y'$ from the mixing distribution $p(y|x)$ and (2) given $y'$, sample from the respective abducted noise variable $p(u|y',x)$. 
\begin{align}
    &p(u|x) \qquad 
    \myeq{law tot. prob.} \qquad \sum_{y' \in \{0, 1\}} p(u, y' | x) 
    \qquad\myeq{cond. prob.} \qquad \sum_{y' \in \{0, 1\}} p(u| y', x) p(y'|x)
    %&\myeq{d-sep} \qquad \sum_{y' \in \{0, 1\}} p(u| y', x_{pa(y)}) p(y'|x)
\end{align}
The binomial mixing distribution $p(y|x)$ can be obtained and sampled from by leveraging the cross-entropy optimal predictor $h^*$ (which can for instance be derived from the SCM, see \ref{appendix:estimation-optimal-pre-recourse}). In order to sample from $p(u| y', x)$ we leverage the Markov factorization, which allows us to sample each component independently using the standard abduction procedure described above.
\begin{equation}
\begin{split}
p(u|x, y') \quad \myeq{d-sep.} \quad &P(u_Y| x_{pa(Y)}, y') 
                                \prod_{k \in ch(Y)} P(u_k|x_k, x_{pa(k)}, y')
                                 \prod_{k \not \in ch(Y)} P(u_k|x_k, x_{pa(k)}).
\end{split}
\end{equation}
The overall procedure is summarized in Algorithm \ref{alg:sample-individualized-covariate}.
\subsubsection{Estimation of \texorpdfstring{$\gamma^{ind}$}{individualized improvement confidence} and \texorpdfstring{$\eta^{ind}$}{acceptance rate}}

Given the procedure to sample from the individualized post-recourse distribution we can estimate $\gamma^{ind}$ by taking the mean over the samples taken for $Y^{post}$. Similarly, for each sample for $X^{post}$ we can compute the prediction $\hat{y}^{post}$ using either $h \geq t$ or $h^{ind} \geq t$. By taking the mean over all sampled predictions $\hat{y}^{post}$ we can estimate the respective acceptance probability $\eta(t; x^{pre}, a, h)$ or $\eta(t; x^{pre}, a, h^{ind})$.

\begin{algorithm}[t]
\caption{Sampling from the individualized post-recourse distribution}%\label{alg:sampling-individualized-covariate}
\KwData{pre-recourse observation $x^{pre}$, action $a$ (where $do(a) := do(X_{I_a} := \theta)$), sample size $M$, structural causal model $\mathcal{M}$ with structural equations $f_j$, observational predictor $h$}
\KwResult{sample from $p(y^{post}, x^{post}|x^{pre}, do(a))$}
get $\mathcal{M}_{do(a)}$ by updating $f_i(x_{pa(i)}; u_i) := \theta_i$ for $i \in I_a$ \;
\For{$m \textbf{ in } (0, ..., M-1)$}{
 sample $y'$ from $Binomial(h(x^{pre}))$ \;
 \For{$j \textbf{ in } D$}{
    sample $u_j^{(m)}$ from $p(u_j|(x,y')_j, (x,y')_{pa(j)})$
    \Comment{comment: leveraging standard abduction}\;
 }
 sample $u_Y^{(m)}$ from $p(u_Y|y', x_{pa(Y)})$ \;
 compute $(x^{post}, y^{post})^{(m)} = f_{\mathcal{M}_{do(a)}}(u^{(m)})$ \;
}
\label{alg:sample-individualized-covariate}
\end{algorithm}

\subsection{Estimation of the individualized post-recourse prediction}
\label{appendix:estimation:individualized-post-recourse-predictor}

\begin{algorithm}[t]
\caption{Estimating $h^{*,ind}$}%\label{alg:sampling-individualized-covariate}
\KwData{pre-recourse observation $x^{pre}$, action $a$, sample size $M$, structural causal model $\mathcal{M}$, observational predictor $h$, $m = 0$}
\KwResult{$\hat{h}^{ind}(x^{post};x^{pre}, do(a))$}
%get $\mathcal{M}_{do(a)}$ by updating $f_i(x_{pa(i)}; u_i) := \theta_i$ for $i \in I_a$ \;
%\For{$m \textbf{ in } (0, ..., M-1)$}{
% sample $y'$ from $Binomial(h^*(x^{pre}))$ \;
% \For{$j \textbf{ in } D$}{
%    sample $u_j^{(m)}$ from $p(u_j|(x,y')_j, (x,y')_{pa(j)})$\;
% }
% sample $u_Y^{(m)}$ from $p(u_Y|y', x_{pa(Y)})$ \;
% compute $(x^{post}, y^{post})^{(m)} = f_{\mathcal{M}_{do(a)}}(u^{(m)})$ \;
%}
%
\While{$m < M$}{
  sample $(x', y')$ using Alg. \ref{alg:sample-individualized-covariate} and $x^{pre}, a, \mathcal{M}, h$\; 
  \If{$x' = x^{post}$} {
    $m = m + 1$; store $y'$ as $y'^{(m)}$ \;
  }
}
$\hat{h}^{ind}(x^{post}) = \frac{1}{M}\sum_{m=1}^M y'^{(m)}$
\label{alg:individualized-post-recourse-prediction}
\end{algorithm}

%In order to allow the estimation of the post-recourse prediction, we decompose Definition \ref{def:individualized-post-recourse-prediction} into tractable components. More specifically, the post-recourse prediction can be composed of the conditional distributions of the endogenous variables given a state of the exogenous variables, and the abducted probabilities of the exogenous variables given the pre-recourse observation. 
We continue to show how the individualized post-recourse prediction can be estimated. We recall that $h^{*,ind}$ is
\begin{equation*}
    h^{*,ind}(x^{post};x^{pre}, a) = P(Y^{post}=1|x^{post}, x^{pre}, do(a)).
\end{equation*}
We can estimate $h^{*,ind}$ by leveraging the procedure to sample from the post-recourse covariate distribution %(which we also used to estimate $\gamma^{ind}$ and $\eta^{ind}$, 
(Algorithm \ref{alg:sample-individualized-covariate}).
More specifically, we draw samples $(y',x')$ from $P(Y^{post}, X^{post}|do(a), x^{pre})$ and keep those that conform with $x^{post}$ (i.e., $x' = x^{post}$). Within the subsample, we compute the proportion of samples for which $y' = 1$ to estimate $p(y^{post}|x^{pre}, x^{post}, do(a))$. In more formal terms, we approximate Eq. \ref{eq:individualized-post-recourse-prediction} using rejection sampling and Monte Carlo integration \citep{koller2009probabilistic}.\\
If the structural equations are invertible\footnote{Meaning that the abducted joint distribution has point mass probability for two configurations, one for each possible state of $Y$.} or the nodes are categorical the procedure is tractable, since many or all samples conform with $x^{post}$. Otherwise the estimation may become intractable. %, because only few samples will exactly conform with the post-recourse state $x^{post}$ (and therefore many samples must be drawn for the estimation to converge). 
%For more complicated settings, 
We see the application of likelihood weighting or MCMC as promising directions
%but leave a detailed investigation for future work. For an introduction to conditional sampling in Bayesian networks we 
and refer interested readers to \citet{koller2009probabilistic}.\\
In addition to the sampling-based procedure we also derive a closed-form solution for settings with invertible structural equations, which is provided in Proposition \ref{prop:individualized-post-recourse-prediction}, Eq. \ref{eq:individualized-post-recourse-prediction-invertible}.\\
%More specifically, we provide two possible solutions. An analytical solution, given 
%Therefore we leverage the procedure to sample from the post-recourse distribution that was described in Algorithm \ref{alg:sample-individualized-covariate}.\\
%If the sampled values confirm with $(y', x^{post})$ then $p(y', x^{post}|u, do(a))$ evaluates to $1$ (and otherwise to $0$).\\
%A general formula as well as a simplified version for invertible structural equations is given in Proposition \ref{prop:individualized-post-recourse-prediction}. A proof can be found in \ref{appendix:proofs:prop:individualized-post-recourse-prediction}.
%
\begin{proposition}
In general, the individualized post-recourse predictor can be estimated as
\begin{align}
\begin{split}
&p(y^{post}|x^{pre}, x^{post}, do(a))\\
&= \frac{\int_{\mathcal{U}} p(y^{post}, x^{post}|u, do(a)) p(u|x^{pre})du}{\sum_{y' \in \{0, 1\}} \left( \int_{\mathcal{U}} p(y', x^{post}|u, do(a)) p(u|x^{pre}) du\right)}
\end{split}
\label{eq:individualized-post-recourse-prediction}
\end{align}
Given invertible structural equations, the individualized post-recourse prediction function reduces to
\begin{align}
  \begin{split}
    &p(y^{post}|x^{post}, x^{pre}, do(a)) \\
    &= \frac{p(U_{-I} = f_{do(a)}^{-1}(y^{post}, x^{post})|x^{pre}, do(a))}{\sum_{y' \in \{0, 1\}} p(U_{-I} = f_{do(a)}^{-1}(y', x^{post})|x^{pre}, do(a))}.  
  \end{split}
\label{eq:individualized-post-recourse-prediction-invertible}
\end{align}
\label{prop:individualized-post-recourse-prediction}
\end{proposition}
%
%The integrals in Equation \ref{eq:individualized-post-recourse-prediction} can be approximated using Monte Carlo integration: we can draw samples from $p(u|x^{pre})$ and compute the respective post-recourse realizations of covariates and target (as detailed in Algorithm \ref{alg:sample-individualized-covariate}). If the sampled values confirm with $(y', x^{post})$ then $p(y', x^{post}|u, do(a))$ evaluates to $1$ (and otherwise to $0$).\\
%If several nodes have non-invertible structural equations and continuous noise variables, estimating the individualized post-recourse predictor may quickly become intractable, because only few samples $u$ will conform with the post-recourse state $x^{post}$ (and therefore many samples must be drawn for the estimation to converge). We see the application of probabilistic inference \citep{koller2009probabilistic} as promising research direction, but leave a detailed investigation for future work.\\
%Apart from that, knowledge about the functional form of the SCM allows for an analytical solution as e.g. in Equation \ref{eq:individualized-post-recourse-prediction-invertible}.
%Since only very few samples from $u$ will  For binary $Y$, the integral over $Y$ reduces to a sum, as in Equation \ref{eq:individualized-post-recourse-prediction-invertible}.
%
% how can it be computed

\subsection{Estimation of the subpopulation-based improvement confidence \texorpdfstring{$\gamma^{sub}$}{} and the subpopulation-based acceptance rate \texorpdfstring{$\eta^{sub}$}{}}
\label{appendix:estimation:sampling-sub}

%\subsubsection{Estimation of the subpopulation-based improvement confidence}
%\label{appendix:details:observational-identifiability}
As follows we detail how to estimate $\gamma^{sub}$ and $\eta^{sub}$. We focus on actions $a$ that potentially affect $Y$, meaning that they intervene on causes of $Y$.\footnote{Actions that do not affect $Y$ trivially do not lead to improvement. The respective probability of $Y=1$ can be estimated using the optimal observational predictor.}\\
In order to estimate $\gamma^{sub}$ and $\eta^{sub}$ we sample $(x',y')$ from the subpopulation-based post-recourse distribution. Given a sample from the subpopulation-based post-recourse distribution we can estimate $\gamma^{sub}$ and $\eta^{sub}$ by taking the respective sample means.\\
We explain the sampling procedure in two steps: We first recall how causal graphs can be leveraged to sample interventional distributions, and then explain why we can apply the procedure to sample from the subpopulation-based post-recourse distribution.

\paragraph{Recap: Sampling interventional distributions leveraging a causally sufficient causal graph $\mathcal{G}$}
{Given a causal graph $\mathcal{G}$ (that fulfills the global Markov property), the joint distribution $P(X,Y)$ can be reformulated using the Markov factorization, which makes use of the $d$-separations in the graph. 
\begin{equation*}
    p(x,y) = p(y|x_{pa(y)}) \prod_{j \in D} p(x_j|(x,y)_{pa(j)})
\end{equation*}
As a consequence, we can sample from the joint distribution by sampling each component given its respective parents. In order to ensure that the parents for each node have been sampled already, the graph is traversed in topological order, starting with the root node and ending with the sink nodes \citep{koller2009probabilistic}.\\
Given that causal sufficiency (no unobserved confounders) and the principle of independent mechanisms hold, the same procedure can also be applied when sampling from interventional distributions of the form $p(x,y|do(a))$ by leveraging the so-called truncated factorization. The intervened upon nodes are not sampled from their parents, but fixed to the values $\theta_a$. The remaining nodes $\Gamma$ are sampled as before:
\begin{align*}
    \begin{split}
    &p((x,y)_{\Gamma}|do(a)) = \prod_{j \in \Gamma} p((x,y)_j | (x,y)_{pa(j) \cap \Gamma}, \theta_{pa(j) \cap I_a}) \\
    &\text{with} \quad \Gamma := D \backslash I_a
    \end{split}
\label{eq:interventional-distribution-factorization}
\end{align*}
%
%Furthermore, the procedure can be directly applied to sample from conditional distributions of the form $p(x,y|do(a), x_S)$, if $S$ is some set of variables for which all ascendants are in $S$ or $I_a$ (i.e., $asc(S) \subseteq (S \cup I_a)$). More specifically, the unfixed nodes $\{D, Y\} \backslash (S \cup I_a)$ can be sampled from their parents in topological order (while intervened upon or conditioned upon nodes are held constant). The condition that $asc(S) \subseteq S \cup I_a$ matters since it ensures that all samples that are drawn with the procedure conform with $x_S$. If there would be a node in $S$ that has an unfixed parent $p$, then we draw samples for $x_p$ may not conform with $x_S$.\\
%We recall that in general without access to the SCM sampling post-intervention values conditional on pre-intervention characteristics is impossible. The conditioning values $x_S$ always have to refer to the post-intervention state \citep{glymour2016causal}.
}

\begin{figure}[t]
    \centering
    \includegraphics[width=0.6\linewidth]{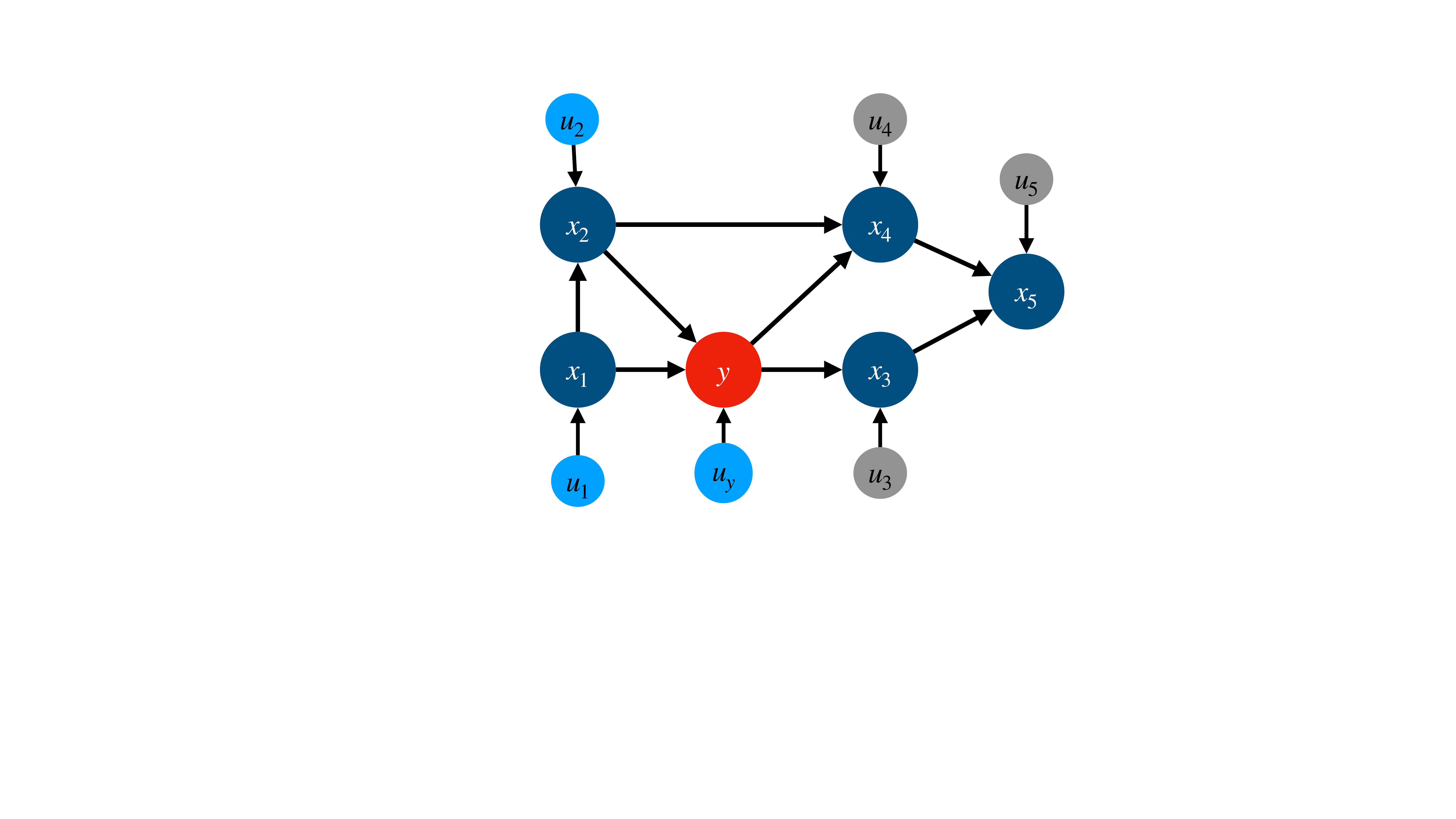}
    \caption{Causal graph $\mathcal{G}_{\overline{I_a}}$ visualizing the subpopulation-based post-recourse setting, including the prediction target $Y$ (light blue), intervened-upon variables $I_a$ (red), the subgroup characteristics $G_a$ (cyan) and the descendants $\Gamma$ that shall be resampled (dark blue). $\overline{I_a}$ indicates that incoming edges to $I_a$ were removed. Right: Causal graph $\mathcal{G}_{\overline{I_a}\underline{G_a}}$ where incoming edges to $I_a$ and outgoing edges from $G_a$ were removed. We observe that in this manipulated graph $G_a$ is $d$-separated from $\Gamma$. Thus, according to the second rule of $do$-calculus, for $G_a$ intervention and conditioning coincide.}
    \label{fig:subpopulation-sampling}
\end{figure}

\paragraph{Sampling from the subpopulation-based post-recourse distribution using $\mathcal{G}$} We recall that for actions $a$ that potentially affect $Y$ the subpopulation-based post-recourse distribution is defined as
\begin{equation}
    P(Y^{post}, X^{post} | do(a), X^{post}_{G_a}=x^{pre}_{G_a}).
\label{eq:sub-post-recourse-distribution}
\end{equation}
As we will see, the previously described sampling procedure can be applied.
Therefore we apply the second rule of $do$-calculus to show that in Equation \ref{eq:sub-post-recourse-distribution} conditioning on $x_{G_a}$ is equal to intervening $do(X_{G_a}=x_{G_a})$. More specifically, if we remove all outgoing edges from $X_{G_a}$ and all incoming edges to $I_a$, then $X_{G_a}$ and $X_\Gamma$ with $\Gamma := D \backslash I_a \cap G_a = d(I_a)$ are $d$-separated, meaning that conditioning and intervention are equivalent (Figure \ref{fig:subpopulation-sampling}). 
\begin{align*}
    &P((Y, X)^{post}_\Gamma | do(a), X^{post}_{G_a} = x^{pre}_{G_a})\\
    &= P((Y, X)^{post}_\Gamma | do(a), do(X^{post}_{G_a}=x^{pre}_{G_a}))
\end{align*}
As follows we can leverage the procedure to sample interventional distributions to sample from the subpopulation-based post-recourse distribution. The procedure is illustrated in Algorithm \ref{alg:evaluating-subpopulation-based-confidence}.
%
\begin{comment}
%More generally, given causal sufficiency, $\gamma^{sub}$ is observationally identifiable (Proposition \ref{prop:subpopulation-meaningful-identifiable}). Here $\Gamma := \{r : r \in d(I)\}$ is the set of descendants of $X_I$.
%

%\begin{proposition}
%Given causal sufficiency of $\mathcal{G}$ and $G_a := nd(I_a)$, the conditional interventional distribution is observationally identifiable. 
%
%\begin{align*}
%&p(y, x_{\Gamma}| do(a), x^{pre}_{G_{a}}) \\
%&= %\int_{\mathcal{X}_\Gamma}
%\left. p(y|x_{pa(Y)}) \prod_{r \in \Gamma} p(x_r|(x,y)_{pa(r)}) 
%d x_\Gamma 
%\right|_{do(a), x_{G_a}^{pre}}    
%\end{align*}
%For actions $a$ that exclusively intervene on non-causal variables ($I^a_{asc} = \emptyset$) it holds that $p(Y = 1|do(X_{I^a}=\theta), x^{pre}_{G_{a}}) = p(Y=1|x^{pre})$.
%\label{prop:subpopulation-meaningful-identifiable}
%\end{proposition}
%
\end{comment}

\begin{algorithm}[t]
\caption{Sampling from the subpopulation-based post-recourse distribution}
\KwData{pre-recourse observation $x^{pre}$, action $a$ with $I_a \cap asc(Y) \neq \emptyset$ ($do(a) := do(X_{I_a} := \theta)$), sample size $M$, causal graph $\mathcal{G}$, conditional distributions $P(X_j|X_{pa(j)})$ for $j \in \Gamma$ with $\Gamma := \{r : r \in asc(Y) \wedge r \in d(I)\}$}
\KwResult{sample from $p(y, x_\Gamma | do(a), x_{G_a})$}
\For{$m \gets 0$ \KwTo $M$}{
  $\Gamma^{sorted} \gets \text{topologicalsort}(\;\Gamma; \mathcal{G}_{do(a)})$ \Comment{sort such that causes precede effects} \;
  \For{$j \textbf{ in }\Gamma^{sorted}$}{
  %\Comment{Note: $x_{pa(j)}^{post}$ is known since all $i \in pa(j)$ where conditioned upon ($i \in G_a$ or $i \in I_a$) or sampled already}\;
    sample $(x,y)_j^{post, (m)}$ $\sim P((X,Y)_j|(X,Y)_{pa(j)} = (x,y)_{pa(j)}^{post})$ \;
  }
  %sample $y^{post, (m)}$ $\sim P(Y|X_{pa(Y)}=x_{pa(Y)}^{post})$ \;
  %\Comment{Note: We only need to sample non-ascendants of $Y$ if we are interested in the acceptance rate}\;
  %\For{$j \textbf{ in }\Gamma^{sorted}_{-asc(Y)}$}{
  %  sample $x_j^{post, (m)}$ $\sim P(X_j|X_{pa(j)} = x_{pa(j)}^{post})$ \;
  %}
  %compute $h(x_{\Gamma}^{post}, \theta_{I_a}, x_{G_a}) \geq t$ to yield $\hat{y}^{post,(m)}$
}
%take mean over samples $y^{post, (m)}$ to yield $\gamma^{sub}(a)$\;
%take mean over samples $\hat{y}^{post, (m)}$ to yield $\eta^{sub}(a)$\;
\label{alg:evaluating-subpopulation-based-confidence}
\end{algorithm}

\subsubsection{Learning the conditional distributions \texorpdfstring{$P(X_j|x_{pa(j)})$}{of the nodes given their parents}}
In this work we assume that we have prior knowledge that allows us to sample from the components of the factorization ($P(X_j|x_{pa(j)})$, e.g. available if we know the SCM).\\
If the conditional distributions are not known, they can be learned from observational data; depending on which assumptions about distribution and functional can be made, different techniques may be employed. For categorical variables the problem reduces to standard supervised learning with cross-entropy loss. For linear Gaussian data, the conditional distribution can be estimated analytically from the covariance matrix \citep{page1984multivariate}. A variety of estimation techniques exist for continuous settings with nonlinearities \citep{bishop1994mixture,bashtannyk2001bandwidth,sohn2015learning,trippe2018conditional,winkler2019learning,hothorn2021predictive}.\\
%
%In \ref{appendix:estimation:sampling-sub} we show how to estimate the subgroup-based post-recourse distribution of $Y$ from observational data. Therefore we draw samples from the post-recourse distribution as described in Algorithm \ref{alg:evaluating-subpopulation-based-confidence}. By applying the predictor we yield the post-recourse prediction distribution, which can be used to compute the subpopulation-based acceptance rate.
%The result can be directly translated to estimate the respective post-recourse distribution for all covariates. %More specifically, Proposition \ref{prop:subpopulation-meaningful-identifiable} (and the respective proof) also holds when replacing $Y$ with $(Y, X_{D \backslash (I_a \cup nd(I_a))})$.

\subsection{Optimization}
\label{appendix:details:optimization}

Like the optimization problems for CE \citep{Wachter2018,tsirtsis2020decisions} or CR \citep{karimi_algorithmic_2020},
the optimization problem for ICR is computationally challenging.
It can be seen as a two-stage problem, where in the first stage the intervention targets $I_a$, and in the second stage the corresponding intervention values $\theta_a$ are optimized \citep{karimi_algorithmic_2020}.
For the selection of intervention targets $I_a$ alone $2^{d'}$ combinations exist, with $d' \leq d$ being the number of causes of $Y$. %We use the \textit{Nondominated Sorting Genetic Algorithm II} (NSGA-II) \citep{deb2002fast}.
%The computational complexity of optimizing the intervention value $\theta$ depends on the type of data. 
We jointly optimize the intervention targets and the intervention values using a genetic algorithm called NSGA-II \citep{deb2002fast}.
For mixed categorical and continuous data, previous work in the field \citep{dandl_multi-objective_2020} suggests to use NSGA-II in combination with \textit{mixed integer evaluation strategies} \citep{li2013mixed}. The exact hyperparameter configurations are reported in \ref{appendix:additional-experiments}.

\subsection{Estimation of the optimal observational predictor \texorpdfstring{$h^*$}{} using the SCM}
\label{appendix:estimation-optimal-pre-recourse}

Instead of leveraging supervised learning with cross-entropy loss, we can factorize the optimal observational predictor as shown in Proposition \ref{prop:optimal-observational-from-SCM} and then leverage the SCM for the estimation.

\begin{proposition}

The optimal observational predictor can be factorized into conditional distributions of nodes given their parents (using the Markov factorization). More specifically, we yield 
\begin{align}
&p(y|x) = \frac{p(x, y)}{p(x)} = \frac{p(x,y)}{\sum_{y' \in \{0, 1\}} p(x,y)}\\
&\quad \myeq{M.f.} \frac{p(y|x_{pa(j)})\prod_{j \in D} p(x_j|(x,y)_{pa(j)})}
{\sum_{y' \in \{0, 1\}} p(y'|x_{pa(j)})\prod_{j \in D} p(x_j|(x,y')_{pa(j)})}\\
& \quad = \frac{p(y|x_{pa(j)})\prod_{j \in ch(y)} p(x_j|x_{pa(j)}, y)}
{\sum_{y' \in \{0, 1\}} p(y'|x_{pa(j)})\prod_{j \in ch(y)} p(x_j|x_{pa(j)}, y')}.
\end{align}
\label{prop:optimal-observational-from-SCM}
\end{proposition}
It remains to show how the conditional distribution $p(x_j|x_{pa(j)})$ of a node given its parents can be estimated. Generally it holds that
\begin{align}
    &p(x_j|x_{pa(j)})\\[5pt] 
    &\qquad \myeq{law tot. prob.} \qquad \int_{\mathcal{U}_j} p(x_j|x_{pa(j)}, u_j) p(u_j|x_{pa(j)}) du \\
    & \qquad \myeq{SCM, $u_j \perp x_{pa(j)}$} \quad \qquad \int_{\mathcal{U}_j} [f(x_{pa(j)}, u_j) = x_j] p(u_j) du.
\end{align}
The integral can be approximated using Monte Carlo integration: we can sample from $p(u_j)$, compute the respective $\tilde{x}_j = f_j(x_{pa(j)}, \tilde{u}_j)$ and compute the proportion of cases where $x_j = \tilde{x}_j$. If $X_j$ and $U_j$ are continuous, this may require huge sample sizes to converge.\\
Furthermore, we may be able to leverage assumptions about $f_j$ to derive a closed form solution. If $f_j$ is invertible, the integral reduces to $p(x_j|x_{pa(j)}) = p(U_j=f_j^{-1}(x_j, x_{pa(j)}))$. For binary nodes with $x_j := [\sigma(l(x_{pa(j)})) \leq u_j]$ and $U_j \sim Unif(0,1)$, we directly see that $p(x_j|x_{pa(j)})=\sigma(l(x_{pa(j)}))$.

%
%\subsection{Link to causal recourse}{
%\label{subsubsec:mcr:subpopulation:links-cr}
%
%Since under causal sufficiency optimal predictors reliably estimate the improvement under interventions on causes we can link subpopulation-based CR and ICR.

%\begin{proposition}
%For actions $a$ that only intervene on causes of $Y$, $p^{pre}(x^{post})>0$, a cross-entropy optimal binary predictor and causal sufficiency, we can equivalently define $\gamma$-subpopulation meaningfulness as
%
%$$E[h^*(\theta_{I_a}, x^{pre}_{G_{a'}}, x_{-(G_{a'} \cup I_a)}^{post})|do(a), x^{pre}_{G_{a'}}] \geq \gamma.$$
%
%\label{prop:effective-meaningfulness}
%\end{proposition}

%Proposition \ref{prop:effective-meaningfulness} can be interpreted in two directions: Firstly, we see that in scenarios where all observed variables are causes of $Y$, CR recommendations are $\texttt{thresh}(a)$-subpopulation meaningful.\footnote{$\texttt{thresh}(a)$ is the CR equivalent to $\gamma$.} Secondly, we can transform subpopulation-based CR into a $\texttt{thresh}(a)$ subpopulation-based meaningfulness technique by adding a constraint that only allows interventions on causes of $Y$.}

\newpage
\section{Details on Experiments}

In this section we provide additional details on the experiments. More specifically, we explain which open-source libraries we use, how to access our code and how to reproduce the results in \ref{appendix:implementation-details}. We formally introduce the synthetic and semi-synthetic datasets that we used in our experiments in \ref{appendix:datasets} and the corresponding figures. Details on hyperparameters, models as well as detailed results are reported in \ref{appendix:additional-experiments} and the corresponding tables. 

\subsection{Implementation}
\label{appendix:implementation-details}

The code relies of efficient tensor calculations with \texttt{numpy} \citep{harris2020array}, \texttt{pytorch} \citep{NEURIPS2019_9015} and \texttt{jax} \citep{jax2018github}. For named dataframes we use \texttt{pandas} \citep{reback2020pandas}. For plotting we rely on \texttt{matplotlib} \citep{Hunter:2007} and \texttt{seaborn} \citep{Waskom2021}. We use the evolutionary optimization library deap \citep{DEAP_JMLR2012} and NSGA-II \citep{deb2002fast} to solve the combinatorial optimization problem.\footnote{We also implemented abduction based on probabilistic inference. Thereby we rely on on \texttt{pyro} \citep{bingham2018pyro} for discrete inference and \texttt{numpyro} \citep{phan2019composable} for MCMC inference of continuous variables. For our experiments we used the analytical formulas presented in \ref{appendix:estimation}} In order to speed up the computation, we cache queries and results for the improvement confidence using \texttt{functools.cache}. For continuous variables the intervention can be rounded to a specified number of digits to increase the probability of reusing a cached result (with neglectable loss of precision).\footnote{All packages are open source. For detailed license information we refer to the respective package websites.}

All code is publicly available via \url{https://github.com/gcskoenig/icr}. The repository contains the user-friendly python package \texttt{icr}, which we use in our experiments to generate and evaluate recourse. Furthermore, the scripts for the experiments, the scripts for the visualization of the results as well as a \texttt{README.md} with instructions for the installation of all dependencies are contained in the repository, such that the experiments are reproducible.

\subsection{Synthetic and Semi-Synthetic Datasets}
\label{appendix:datasets}

\begin{figure*}[p]
    \centering
    \hfill
    \begin{subfigure}{0.49\linewidth}
      \includegraphics[width=0.99\linewidth]{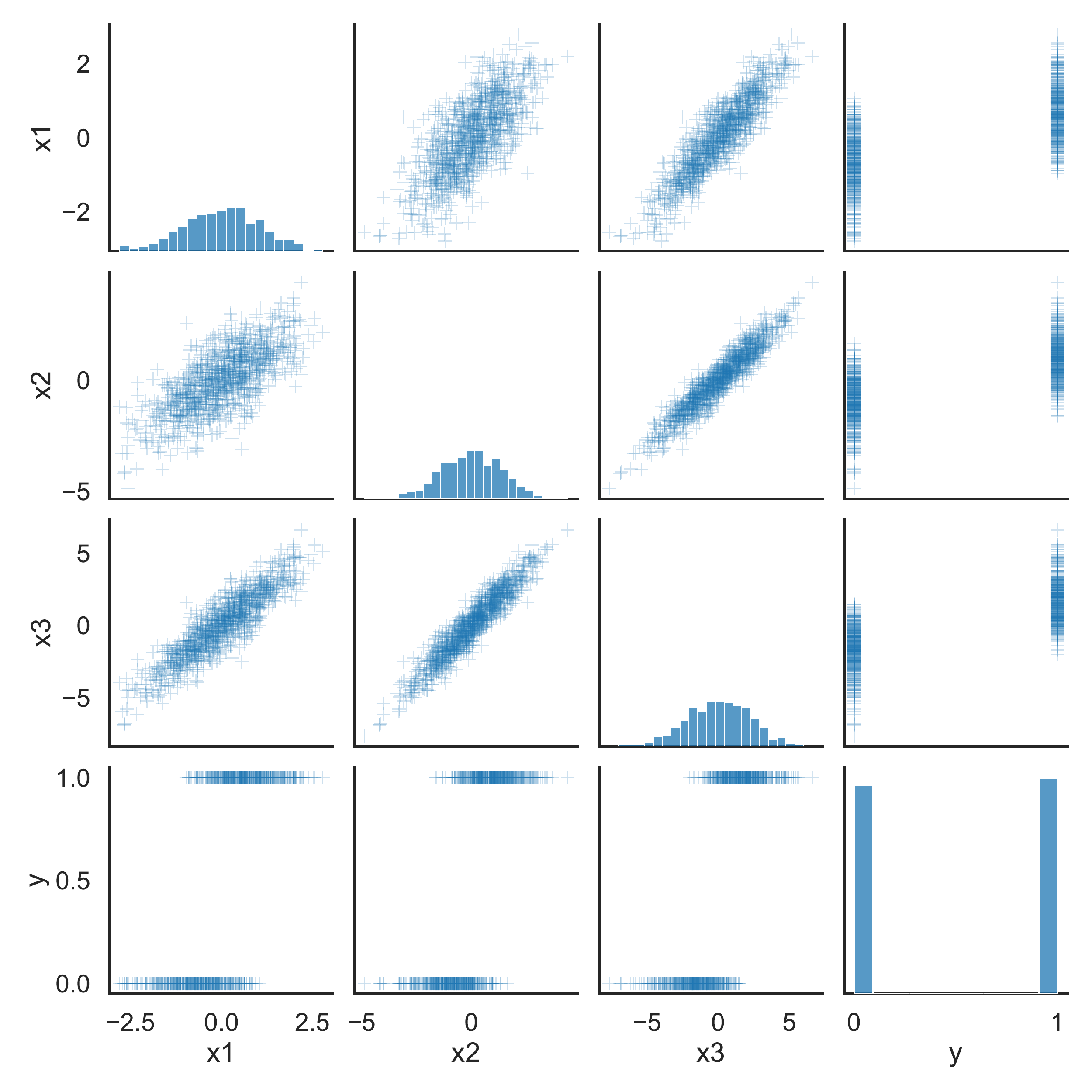}
      \caption{Pairplot for \textit{3var-causal}.}
    \end{subfigure}
    \hfill
    \begin{subfigure}{0.49\linewidth}
      \includegraphics[width=0.99\linewidth]{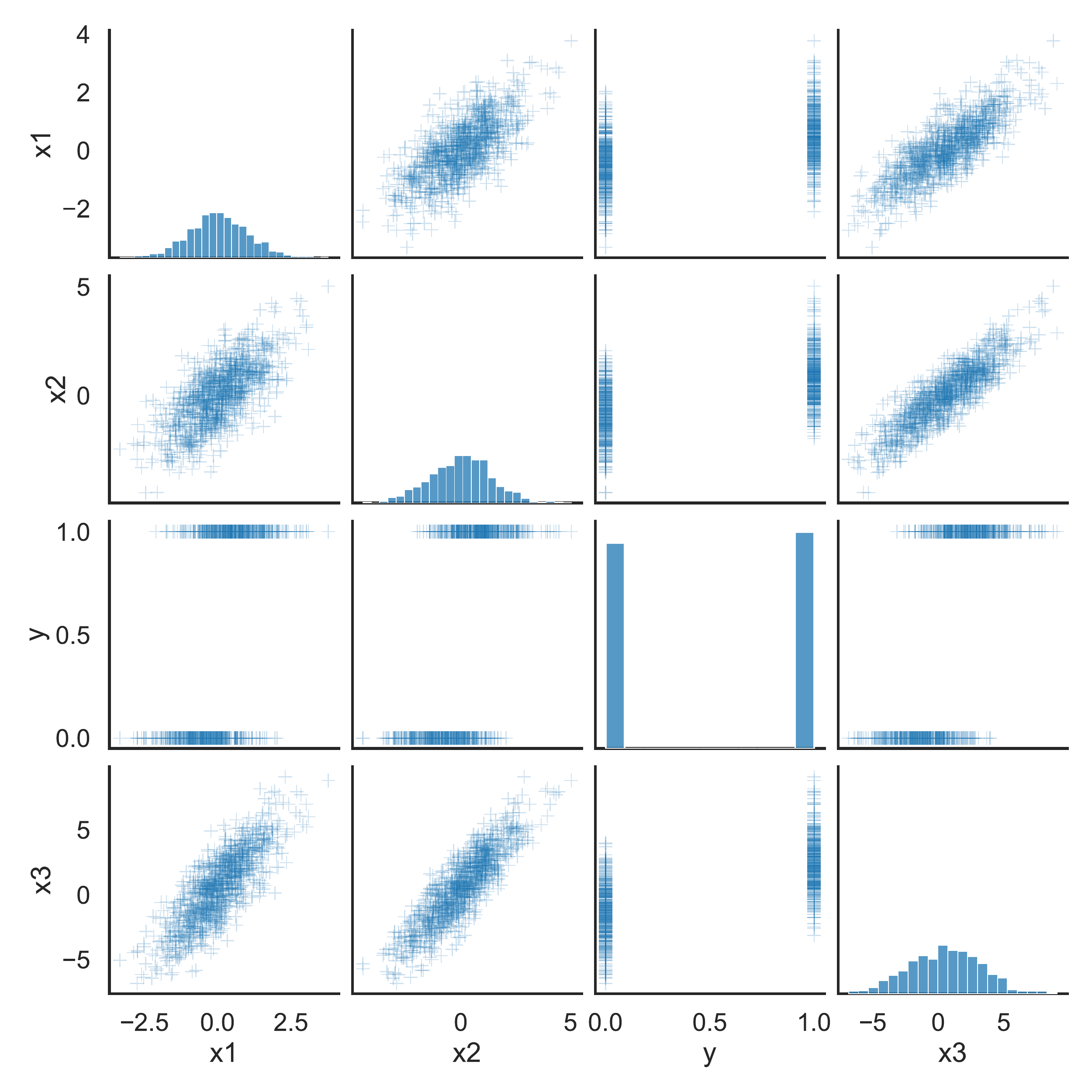}
      \caption{Pairplot for \textit{3var-noncausal}.}
    \end{subfigure}
    \hfill
    
    \bigskip
    \bigskip
    \hfill
    \begin{subfigure}{0.49\linewidth}
        \includegraphics[width=0.99\linewidth]{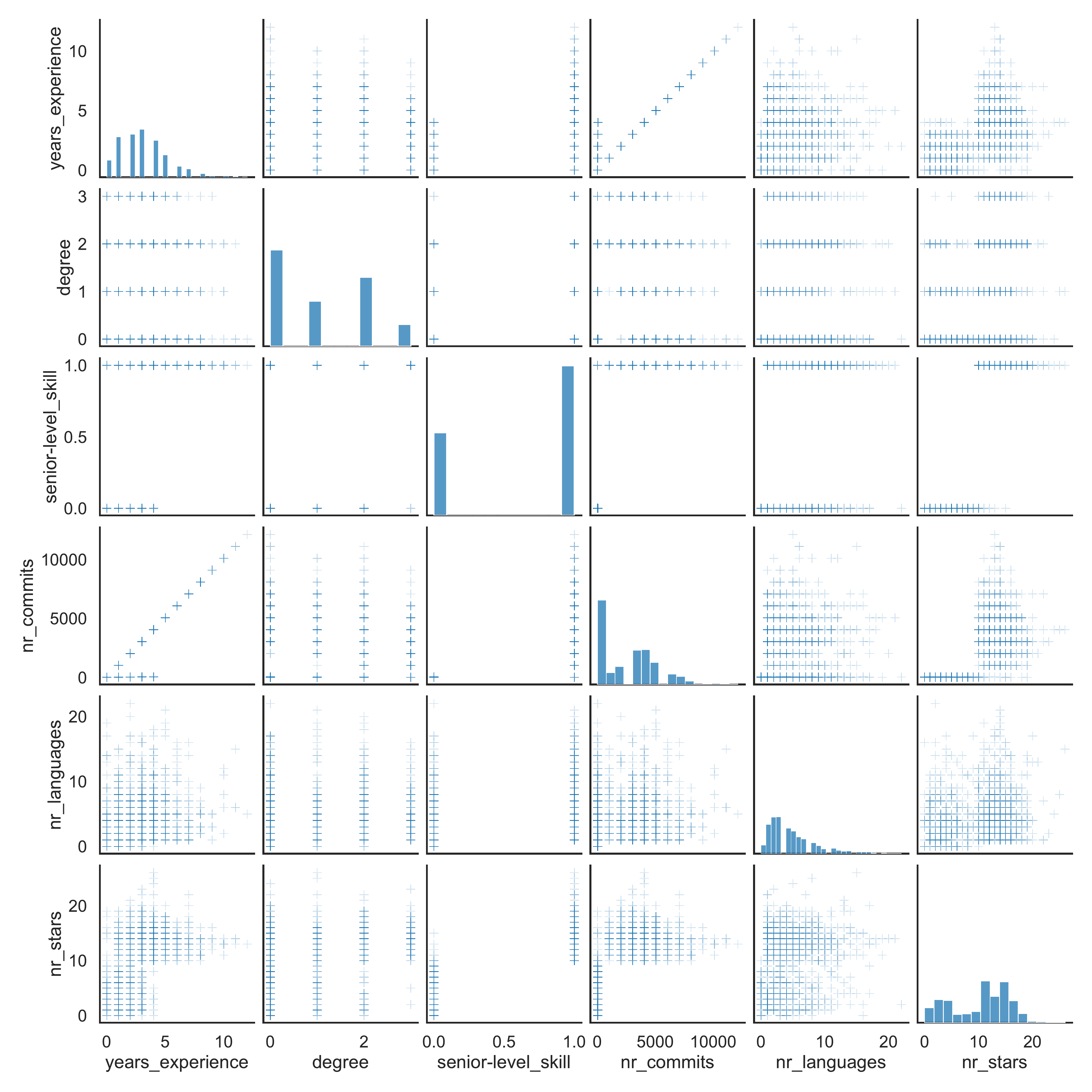}
        \caption{Pairplot for \textit{5var-skill}.}
    \end{subfigure}
    \hfill
    \begin{subfigure}{0.49\linewidth}
      \includegraphics[width=0.99\linewidth]{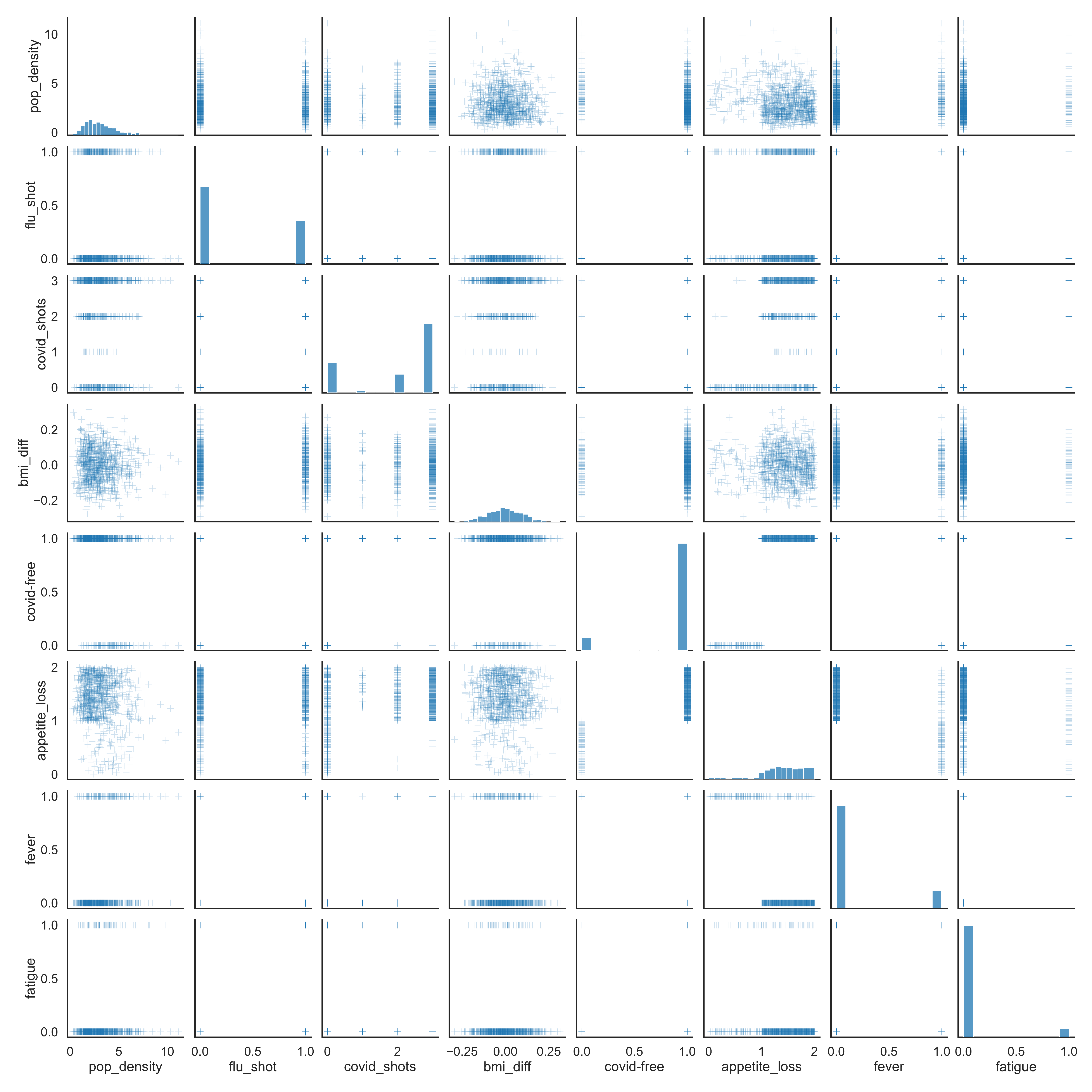}
      \caption{Pairplot for \textit{7var-covid}.}
    \end{subfigure}
    \hfill
    \caption{Pairplots for the SCMs.}
    \label{fig:pairplots}
\end{figure*}

\textit{3var-causal} and \textit{3var-noncausal} are abstract, synthetic settings. \textit{5var-skill} is inspired by \citet{montandon2021mining}, who use GitHub profiles to detect the role of a developer. In our SCM we model \textit{senior-level skill} as a binary variable which is caused by \textit{programming experience} and the education \textit{degree}. The skill is causal for GitHub metrics such as the number of \textit{commits}, the number of programming \textit{languages} and the number of \textit{stars}. The \textit{7var-covid} dataset is inspired by \citet{jehi2020individualizing}. The following variables are introduced: population density $D$, flu vaccination $V_I$, number of covid vaccination shots $V_C$, deviation from average BMI $B$, whether someone is free of covid disease $C$, whether the individual has influence $I$, appetite loss $S_A$, fever $S_{Fe}$ and fatigue $S_{Fa}$. The corresponding structural equations, noise distributions and causal graphs are provided in Figure \ref{fig:3var-causal} (\textit{3var-causal}), \ref{fig:3var-noncausal} (\textit{3var-noncausal}), \ref{fig:5var-skill} (\textit{5var-skill}) and \ref{fig:7var-covid} (\textit{7var-covid}). A pairplot for each dataset is presented in Figure \ref{fig:pairplots}. In our notation $\sigma$ is the sigmoid function, $N$ the Gaussian distribution, $Cat$ a categorical distribution, $Unif$ the uniform distribution, $Bern$ a Bernoulli distribution and $GaP$ a Gamma-Poisson mixture.
$\left [ cond \right ]$ is $1$ when the condition is met and $0$ if not.
As a consequence variables with $\left [ Z \leq U \right ]$ and $U \sim Unif(0,1)$ are bernoulli distributed with $Bern(Z)$.

\begin{figure*}[p]
\centering
\hfill
\subcaptionbox{Causal graph}[0.32\textwidth]{
\centering
  \vspace{\fill}
  \begin{tikzpicture}[thick, scale=.8, every node/.style={scale=.75, line width=0.25mm, black, fill=white}]
    \usetikzlibrary{shapes}
    
        \node[fill=RoyalBlue, ellipse, scale=0.9, text=white] (a1) at (-4, 2) {$X_1$};
        \node[fill=RoyalBlue, ellipse, scale=0.9, text=white] (a2) at (-1, 2) {$X_2$};
        \node[fill=RoyalBlue, ellipse, scale=0.9, text=white] (a3) at (-2.5, 1) {$X_3$};
        \node[fill=black, ellipse, scale=0.9, text=white] (ay) at (-2.5, -0.5) {$Y$};
        
        \draw[-stealth, gray, scale=0.3] (a1) -- (a2);
        \draw[-stealth, gray, scale=0.3] (a1) -- (a3);
        \draw[-stealth, gray, scale=0.3] (a1) -- (ay);
        \draw[-stealth, gray, scale=0.3] (a2) -- (a3);
        \draw[-stealth, gray, scale=0.3] (a2) -- (ay);
        \draw[-stealth, gray, scale=0.3] (a3) -- (ay);
    
    \end{tikzpicture}
    \vspace{\fill}
  \label{subfig:graph-1}
}
\subcaptionbox{Structural Equations}[0.67\textwidth]{
    \centering
  \begin{align*}
X_1 &:= U_1, &U_1 \sim N(0, 1)\\
X_2 &:= X_1 + U_2, &U_2 \sim N(0, 1)\\
X_3 &:= X_1 + X_2 + U_3, &U_3 \sim N(0, 1)\\
Y &\sim \left [ \sigma(X_1 + X_2 + X_3) \leq U_Y \right ], &U_Y \sim Unif(0,1)
\end{align*}
  \label{subfig:sem-1}
}
\caption{SCM for \textit{3var-causal}. The cost function is given as $cost(a) = \delta_1 + \delta_2 + \delta_3$, where $\delta$ is the vector of absolute changes to the intervened upon variables. E.g., for $do(a) = do(X_1=x_1')$, $\delta_1 = |x_1' - x_1|$ and $\delta_2 = \delta_3 = 0$}
\label{fig:3var-causal}
\end{figure*}

\begin{figure*}[p]
\centering
\hfill
\subcaptionbox{Causal graph}[0.32\textwidth]{
\centering
  \begin{tikzpicture}[thick, scale=.8, every node/.style={scale=.75, line width=0.25mm, black, fill=white}]
    \usetikzlibrary{shapes}
    
        \node[fill=orange, ellipse, scale=0.9, text=white] (b1) at (4, 2) {$X_1$};
        \node[fill=orange, ellipse, scale=0.9, text=white] (b2) at (1, 2) {$X_2$};
        \node[fill=black, ellipse, scale=0.9, text=white] (b3) at (2.5, 1) {$Y$};
        \node[fill=orange, ellipse, scale=0.9, text=white] (by) at (2.5, -0.5) {$X_3$};
        
        \draw[-stealth, gray, scale=0.3] (b1) -- (b2);
        \draw[-stealth, gray, scale=0.3] (b1) -- (b3);
        \draw[-stealth, gray, scale=0.3] (b1) -- (by);
        \draw[-stealth, gray, scale=0.3] (b2) -- (b3);
        \draw[-stealth, gray, scale=0.3] (b2) -- (by);
        \draw[-stealth, gray, scale=0.3] (b3) -- (by);
    \end{tikzpicture}
  \label{subfig:graph-2}
}
\subcaptionbox{Structural Equations}[0.67\textwidth]{
    \centering
\begin{align*}
X_1 &:= U_1, &U_1 \sim N(0, 1)\\
X_2 &:= X_1 + U_1, &U_1 \sim N(0, 1)\\
Y &:= \left [ \sigma(X_1 + X_2) \leq U_Y \right ], &U_Y \sim Unif(0,1)\\
X_3 &:= X_1 + X_2 + Y + U_3, &U_3 \sim N(0, 0.1)   
\end{align*}
  \label{subfig:sem-2}
}
\caption{SCM for \textit{3var-noncausal} with $cost(a) = \delta_1 + \delta_2 + \delta_3$.}
\label{fig:3var-noncausal}
\end{figure*}

\begin{figure*}[p]
\centering
\hfill
\subcaptionbox{Causal graph}[0.4\textwidth]{
\centering
  \begin{tikzpicture}[thick, scale=.65, every node/.style={scale=.6, line width=0.25mm, black, fill=white}]
    \usetikzlibrary{shapes}
    
        \node[fill=red, ellipse, scale=0.9, text=white] (d1) at (-7, -3) {experience $E$};
        \node[fill=red, ellipse, scale=0.9, text=white] (d2) at (-3, -3) {degree $D$};
        \node[fill=black, ellipse, scale=0.9, text=white] (dy) at (-5, -4.5) {senior-level skill $S$};
        \node[fill=red, ellipse, scale=0.9, text=white] (d3) at (-8, -6) {nr commits $G_C$};
        \node[fill=red, ellipse, scale=0.9, text=white] (d4) at (-5, -6) {nr languages $G_L$};
        \node[fill=red, ellipse, scale=0.9, text=white] (d5) at (-2, -6) {nr stars $G_S$};
        
        \draw[-stealth, gray, scale=0.3] (d1) -- (dy);
        \draw[-stealth, gray, scale=0.3] (d2) -- (dy);
        \draw[-stealth, gray, scale=0.3] (dy) -- (d3);
        \draw[-stealth, gray, scale=0.3] (dy) -- (d4);
        \draw[-stealth, gray, scale=0.3] (dy) -- (d5);
        \draw[-stealth, gray, scale=0.3] (d1) -- (d3);
    
    \end{tikzpicture}
    
    \bigskip
    \bigskip
  \label{subfig:graph-3}
}
\subcaptionbox{Structural Equations}[0.55\textwidth]{
    \centering
\begin{align*}
E &:= U_E; U_E \sim GaP(8, 8/3)\\
D &:= U_D; U_D \sim Cat(0.4, 0.2, 0.3, 0.1)\\
S &:= \left [\sigma(-10 + 3E + 4D)) \leq U_S \right ]; U_S \sim Unif(0,1)\\
G_C &:= 10E (11+100D) + U_{G_C}; U_{G_C} \sim GaP(40, 40/4)\\
G_L &:= \sigma(10S) + U_{G_L}; U_{G_L} \sim GaP(2, 2/4)\\
G_S &:= 10S + U_{G_S}; U_{G_S} \sim GaP(5, 5/4)   
\end{align*}
  \label{subfig:sem-3}
}
\caption{SCM for \textit{5var-skill} with $cost(a) = 5\delta_E + 5\delta_D + 0.0001\delta_{G_C} + 0.01 \delta_{G_L} + 0.1 \delta{G_S}$.}
\label{fig:5var-skill}
\end{figure*}

\begin{figure*}[p]
\centering
\subcaptionbox{Causal graph}[0.38\textwidth]{
\centering
\vspace{\fill}
  \begin{tikzpicture}[thick, scale=0.5, every node/.style={scale=.6, line width=0.25mm, black, fill=white}]
    \usetikzlibrary{shapes}
    
        \node[fill=Green, ellipse, scale=0.9, text=white] (c1) at (4.5, -3) {density  $D$};
        \node[fill=Green, ellipse, scale=0.9, text=white] (c2) at (1.5, -3) {flu vacc $V_I$};
        \node[fill=Green, ellipse, scale=0.9, text=white] (c3) at (-1.5, -3) {covid shots $V_C$};
        \node[fill=Green, ellipse, scale=0.9, text=white] (c4) at (-4.5, -3) {BMI $B$};
        \node[fill=black, ellipse, scale=0.9, text=white] (cy) at (0, -4.5) {covid-free $C$};
        \node[fill=Green, ellipse, scale=0.9, text=white] (c5) at (3, -6) {appetite $S_A$};
        \node[fill=Green, ellipse, scale=0.9, text=white] (c6) at (0, -6) {fever $S_{Fe}$};
        \node[fill=Green, ellipse, scale=0.9, text=white] (c7) at (-3, -6) {fatigue $S_{Fa}$};
        
        \draw[-stealth, gray, scale=0.3] (c1) -- (cy);
        \draw[-stealth, gray, scale=0.3] (c2) -- (cy);
        \draw[-stealth, gray, scale=0.3] (c3) -- (cy);
        \draw[-stealth, gray, scale=0.3] (c4) -- (cy);
        \draw[-stealth, gray, scale=0.3] (c1) -- (c5);
        \draw[-stealth, gray, scale=0.3] (cy) -- (c5);
        \draw[-stealth, gray, scale=0.3] (cy) -- (c6);
        \draw[-stealth, gray, scale=0.3] (cy) -- (c7);
    
    \end{tikzpicture}
    \bigskip
    \bigskip
    \bigskip
    \bigskip
    \vspace{\fill}
  \label{subfig:graph-4}
}
\hfill
\subcaptionbox{Structural Equations}[0.5\textwidth]{
    \centering
\begin{align*}
    D &:= U_D; U_D \sim \Gamma(4, 4/3)\\
    V_I &:= U_{V_I}; U_{V_I} \sim Bern(0.39)\\
    V_C &:= U_{V_C}; U_{V_C} \sim Cat(0.24, 0.02, 0.15, 0.59)\\
    B &:= U_B; U_B \sim N(0, 1)\\
    C &:= \left[\sigma(-(D-3 - V_I - 2.5 V_C + 0.2 B^2)) \leq U_C \right];\\
    U_C &\sim Unif(0, 1)\\
    S_A &:= \left [ \sigma( -2C) \leq U_{S_A} \right]; U_{S_A} \sim Unif(0,1)\\
    S_{Fe} &:= \left[ \sigma(5 -9 C) \leq U_{S_{Fe}} \right]; U_{S_{Fe}} \sim Unif(0,1)\\
    S_{Fa} &:= \left[ \sigma (-1 + B^2 -2C) \leq U_{S_{Fa}} \right];\\
    U_{S_{Fa}} &\sim Unif(0,1)
\end{align*}
  \label{subfig:sem-4}
}
\caption{SCM for \textit{7var-covid} with cost function $cost(a) = \delta_D + \delta_{V_I} + \delta_{V_C} + \delta_B + \delta_{S_A} + \delta_{S_{Fe}} + \delta_{S_{Fa}}$.}
\label{fig:7var-covid}
\end{figure*}

\subsection{Detailed Results}
\label{appendix:additional-experiments}

\begin{table*}[p]
\caption{Results for 3var-causal.}
\centering
%\ra{1.3}
\begin{tabular}{@{}llrrrrrrrrrrrr@{}}\toprule
%\multicolumn{18}{c}{Covid office admission}\\
\textbf{3var-causal} 
&$\overline{\gamma}$ / $\overline{\eta}$
&$\gamma_{\text{obs.}}$
& $\pm$ 
& $\eta_{\text{obs.}}$
& $\pm$ 
& $\eta^{individ.}_{\text{obs.}}$
& $\pm$
& $\eta^{\text{refit}}_{\text{obs.}}$
& $\pm$
& $\emptyset$ cost
& $\pm$\\
%\cmidrule{2-3} \cmidrule{5-6} \cmidrule{8-9} \cmidrule{11-12} \cmidrule{14-15} \cmidrule{17-18}
%
%$\gamma/\texttt{thresh}:$  &$0.9$  & $0.95$ & & $0.9$  & $0.95$ & &$0.9$  & $0.95$ & &$0.9$  & $0.95$ & &$0.9$  & $0.95$ & &causes  & other\\
%
\midrule
CE & - &0.41  &0.09  &1.00  &0.00  &-  &-  &0.60  &0.20  &3.08  &0.41\\ 
\midrule
ind. CR  &0.75  &0.47  &0.10  &1.00  &0.00  &-  &-  &0.70  &0.10  &2.46  &0.37\\ 
ind. CR  &0.85  &0.44  &0.08  &1.00  &0.00  &-  &-  &0.72  &0.12  &2.39  &0.25\\ 
ind. CR  &0.90  &0.47  &0.09  &1.00  &0.00  &-  &-  &0.72  &0.14  &2.36  &0.35\\ 
ind. CR  &0.95  &0.49  &0.07  &1.00  &0.00  &-  &-  &0.67  &0.10  &2.44  &0.31\\ 
\midrule
subp. CR  &0.75  &0.46  &0.11  &0.86  &0.04  &-  &-  &0.64  &0.14  &2.66  &0.41\\ 
subp. CR  &0.85  &0.43  &0.08  &0.93  &0.02  &-  &-  &0.69  &0.14  &2.64  &0.32\\ 
subp. CR  &0.90  &0.45  &0.09  &0.96  &0.02  &-  &-  &0.70  &0.15  &2.73  &0.42\\ 
subp. CR  &0.95  &0.48  &0.09  &0.98  &0.01  &-  &-  &0.64  &0.14  &2.86  &0.41\\ 
\midrule
ind. ICR  &0.75  &0.79  &0.06  &0.98  &0.02  &1.0  &0.0  &0.96  &0.03  &3.27  &0.50\\ 
ind. ICR  &0.85  &0.86  &0.03  &1.00  &0.01  &1.0  &0.0  &0.97  &0.02  &3.82  &0.30\\ 
ind. ICR  &0.90  &0.90  &0.02  &1.00  &0.01  &1.0  &0.0  &0.98  &0.03  &3.70  &0.31\\ 
ind. ICR  &0.95  &0.95  &0.01  &1.00  &0.00  &1.0  &0.0  &0.99  &0.01  &4.08  &0.24\\ 
\midrule
subp. ICR  &0.75  &0.75  &0.04  &0.93  &0.04  &-  &-  &0.90  &0.04  &3.34  &0.49\\ 
subp. ICR  &0.85  &0.87  &0.03  &0.98  &0.01  &-  &-  &0.96  &0.02  &4.05  &0.29\\ 
subp. ICR  &0.90  &0.89  &0.02  &0.99  &0.01  &-  &-  &0.97  &0.02  &3.87  &0.25\\ 
subp. ICR  &0.95  &0.94  &0.02  &1.00  &0.00  &-  &-  &0.99  &0.01  &4.22  &0.28\\

\bottomrule
\end{tabular}
\label{table:3var-causal}
\end{table*}

\begin{table*}[p]
\caption{Results for 3var-noncausal}
\centering
%\ra{1.3}
\begin{tabular}{@{}llrrrrrrrrrrrr@{}}\toprule
%\multicolumn{18}{c}{Covid office admission}\\
\textbf{3var-noncausal} 
&$\overline{\gamma}$ / $\overline{\eta}$
&$\gamma_{\text{obs.}}$
& $\pm$ 
& $\eta_{\text{obs.}}$
& $\pm$ 
& $\eta^{individ.}_{\text{obs.}}$
& $\pm$
& $\eta^{\text{refit}}_{\text{obs.}}$
& $\pm$
& $\emptyset$ cost
& $\pm$\\
\midrule
CE  &-  &0.17  &0.03  &0.98  &0.04  &-  &-  &0.67  &0.15  &2.28  &0.26\\ 
\midrule
ind. CR  &0.75  &0.25  &0.03  &1.00  &0.00  &-  &-  &0.70  &0.13  &2.28  &0.21\\ 
ind. CR  &0.85  &0.24  &0.02  &1.00  &0.00  &-  &-  &0.73  &0.13  &2.29  &0.17\\ 
ind. CR  &0.90  &0.24  &0.04  &1.00  &0.00  &-  &-  &0.71  &0.11  &2.24  &0.16\\ 
ind. CR  &0.95  &0.23  &0.04  &1.00  &0.00  &-  &-  &0.73  &0.12  &2.18  &0.32\\ 
\midrule
subp. CR  &0.75  &0.22  &0.03  &0.91  &0.03  &-  &-  &0.63  &0.15  &2.18  &0.12\\ 
subp. CR  &0.85  &0.19  &0.03  &0.95  &0.02  &-  &-  &0.67  &0.15  &2.33  &0.21\\ 
subp. CR  &0.90  &0.19  &0.03  &0.97  &0.01  &-  &-  &0.65  &0.14  &2.42  &0.19\\ 
subp. CR  &0.95  &0.19  &0.03  &0.99  &0.01  &-  &-  &0.69  &0.14  &2.26  &0.32\\ 
\midrule
ind. ICR  &0.75  &0.77  &0.03  &0.93  &0.02  &0.79  &0.03  &0.93  &0.02  &2.16  &0.11\\ 
ind. ICR  &0.85  &0.86  &0.02  &0.99  &0.01  &0.90  &0.02  &0.99  &0.01  &2.51  &0.08\\ 
ind. ICR  &0.90  &0.91  &0.03  &1.00  &0.00  &0.94  &0.01  &1.00  &0.00  &3.00  &0.08\\ 
ind. ICR  &0.95  &0.96  &0.02  &0.98  &0.07  &0.98  &0.01  &0.98  &0.08  &3.32  &0.16\\ 
\midrule
subp. ICR  &0.75  &0.69  &0.03  &0.77  &0.05  &-  &-  &0.76  &0.05  &2.11  &0.20\\ 
subp. ICR  &0.85  &0.82  &0.03  &0.93  &0.02  &-  &-  &0.92  &0.02  &2.42  &0.11\\ 
subp. ICR  &0.90  &0.89  &0.03  &0.98  &0.01  &-  &-  &0.97  &0.01  &2.86  &0.13\\ 
subp. ICR  &0.95  &0.94  &0.02  &0.97  &0.10  &-  &-  &0.96  &0.12  &3.19  &0.15\\
\bottomrule
\end{tabular}
\label{table:3var-noncausal}
\end{table*}

\begin{table*}[p]
\caption{Results for 5var-skill}
\centering
%\ra{1.3}
\begin{tabular}{@{}llrrrrrrrrrrrr@{}}\toprule
%\multicolumn{18}{c}{Covid office admission}\\
\textbf{5var-skill} 
&$\overline{\gamma}$ / $\overline{\eta}$
&$\gamma_{\text{obs.}}$
& $\pm$ 
& $\eta_{\text{obs.}}$
& $\pm$ 
& $\eta^{individ.}_{\text{obs.}}$
& $\pm$
& $\eta^{\text{refit}}_{\text{obs.}}$
& $\pm$
& $\emptyset$ cost
& $\pm$\\
\midrule
CE  &-  &0.00  &0.00  &1.00  &0.00  &-  &-  &0.76  &0.14  &1.34  &1.28\\ 
\midrule
ind. CR  &0.75  &0.00  &0.00  &1.00  &0.00  &-  &-  &0.86  &0.11  &0.27  &0.28\\ 
ind. CR  &0.85  &0.00  &0.00  &1.00  &0.00  &-  &-  &0.81  &0.14  &0.24  &0.20\\ 
ind. CR  &0.90  &0.00  &0.01  &1.00  &0.00  &-  &-  &0.70  &0.15  &0.10  &0.00\\ 
ind. CR  &0.95  &0.00  &0.00  &1.00  &0.00  &-  &-  &0.66  &0.16  &0.11  &0.03\\ 
\midrule
subp. CR  &0.75  &0.00  &0.00  &1.00  &0.00  &-  &-  &0.85  &0.11  &4.06  &4.97\\ 
subp. CR  &0.85  &0.00  &0.00  &1.00  &0.00  &-  &-  &0.80  &0.15  &0.24  &0.19\\ 
subp. CR  &0.90  &0.00  &0.01  &1.00  &0.00  &-  &-  &0.70  &0.15  &0.10  &0.01\\ 
subp. CR  &0.95  &0.00  &0.00  &1.00  &0.00  &-  &-  &0.66  &0.15  &0.12  &0.04\\ 
\midrule
ind. ICR  &0.75  &0.94  &0.02  &0.94  &0.02  &0.94  &0.02  &0.94  &0.02  &4.95  &5.32\\ 
ind. ICR  &0.85  &0.94  &0.01  &0.93  &0.02  &0.94  &0.01  &0.93  &0.02  &9.80  &0.27\\ 
ind. ICR  &0.90  &0.96  &0.02  &0.96  &0.02  &0.96  &0.02  &0.96  &0.02  &10.38  &0.23\\ 
ind. ICR  &0.95  &0.98  &0.01  &0.98  &0.01  &0.98  &0.01  &0.98  &0.01  &11.23  &0.21\\ 
\midrule
subp. ICR  &0.75  &0.93  &0.01  &0.93  &0.02  &-  &-  &0.93  &0.01  &4.72  &5.08\\ 
subp. ICR  &0.85  &0.94  &0.01  &0.94  &0.01  &-  &-  &0.94  &0.02  &9.74  &0.17\\ 
subp. ICR  &0.90  &0.96  &0.01  &0.96  &0.01  &-  &-  &0.96  &0.01  &10.46  &0.53\\ 
subp. ICR  &0.95  &0.97  &0.01  &0.97  &0.01  &-  &-  &0.97  &0.01  &10.88  &0.21\\
\bottomrule
\end{tabular}
\label{table:5var-skill}
\end{table*}

\begin{table*}[p]
\caption{Results for 7var-covid}
\centering
%\ra{1.3}
\begin{tabular}{@{}llrrrrrrrrrrrr@{}}\toprule
%\multicolumn{18}{c}{Covid office admission}\\
\textbf{7var-covid} 
&$\overline{\gamma}$ / $\overline{\eta}$
&$\gamma_{\text{obs.}}$
& $\pm$ 
& $\eta_{\text{obs.}}$
& $\pm$ 
& $\eta^{individ.}_{\text{obs.}}$
& $\pm$
& $\eta^{\text{refit}}_{\text{obs.}}$
& $\pm$
& $\emptyset$ cost
& $\pm$\\
\midrule
CE  &-  &0.00  &0.00  &1.00  &0.00  &-  &-  &1.00  &0.00  &0.60  &0.12\\
\midrule
ind. CR  &0.75  &0.01  &0.00  &1.00  &0.00  &-  &-  &0.99  &0.01  &0.56  &0.02\\ 
ind. CR  &0.85  &0.00  &0.00  &1.00  &0.00  &-  &-  &0.99  &0.00  &0.55  &0.02\\ 
ind. CR  &0.90  &0.00  &0.00  &1.00  &0.00  &-  &-  &1.00  &0.00  &0.55  &0.03\\ 
ind. CR  &0.95  &0.00  &0.00  &1.00  &0.00  &-  &-  &0.99  &0.01  &0.54  &0.07\\
\midrule
subp. CR  &0.75  &0.01  &0.01  &0.92  &0.02  &-  &-  &0.91  &0.02  &0.52  &0.03\\ 
subp. CR  &0.85  &0.00  &0.01  &0.97  &0.01  &-  &-  &0.96  &0.01  &0.75  &0.40\\ 
subp. CR  &0.90  &0.00  &0.00  &0.98  &0.01  &-  &-  &0.98  &0.01  &0.55  &0.03\\ 
subp. CR  &0.95  &0.00  &0.00  &0.99  &0.01  &-  &-  &0.98  &0.01  &0.51  &0.07\\  
\midrule
ind. ICR  &0.75  &0.81  &0.03  &0.81  &0.03  &0.82  &0.04  &0.81  &0.03  &1.26  &0.02\\ 
ind. ICR  &0.85  &0.85  &0.03  &0.85  &0.03  &0.86  &0.03  &0.85  &0.03  &1.14  &0.44\\ 
ind. ICR  &0.90  &0.89  &0.03  &0.89  &0.03  &0.90  &0.02  &0.89  &0.03  &1.61  &0.02\\ 
ind. ICR  &0.95  &0.95  &0.01  &0.95  &0.01  &0.95  &0.01  &0.95  &0.01  &1.97  &0.06\\ 
\midrule
subp. ICR  &0.75  &0.61  &0.04  &0.61  &0.04  &-  &-  &0.61  &0.04  &1.06  &0.03\\ 
subp. ICR  &0.85  &0.73  &0.03  &0.73  &0.03  &-  &-  &0.73  &0.03  &1.09  &0.34\\ 
subp. ICR  &0.90  &0.81  &0.04  &0.81  &0.04  &-  &-  &0.81  &0.04  &1.42  &0.05\\ 
subp. ICR  &0.95  &0.90  &0.03  &0.90  &0.03  &-  &-  &0.90  &0.03  &1.73  &0.06\\
\bottomrule
\end{tabular}
\label{table:7var-covid}
\end{table*}

In this section we report all experimental results in tabular form. More specifically, the results for \textit{3var-causal} are reported in Table \ref{table:3var-causal}, for \textit{3var-noncausal} in Table \ref{table:3var-noncausal}, for \textit{5var-skill} in Table \ref{table:5var-skill} and for \textit{7var-covid} in Table \ref{table:7var-covid}. For each experiment we report the specified confidence $\gamma$ (or $\eta$ for CR), as well as the observed improvement rate $\gamma_{obs}$, the observed acceptance rate $\eta_{obs}$, the observed acceptance rate by the individualized post-recourse predictor $\eta_{obs}^{\text{indiv.}}$, the observed acceptance rate on refits $\eta_{obs}^{\text{refit}}$ and the average recourse cost for individuals who were rejected and whom were provided with a recourse recommendation. A visual summary of the results is provided in Section \ref{sec:simulation}.

In order to enable a more direct comparison of the CR and ICR targets, we equalize the optimization thresholds for ICR and CR. More specifically, for CR we require the (individualized or subpopulation-based) acceptance probability to be $\geq \eta$, and for ICR we require the (individualized or subpopulation-based) improvement probability to be $\geq \overline{\gamma}$, where $\overline{\gamma} = \overline{\eta}$.\footnote{A short comment on the choice of a non-adaptive threshold can be found in \ref{appendix:details:expectation-objective}.} Furthermore, in order to be able to estimate the effects of recourse actions, CR assumes causal sufficiency, meaning that there are no two endogeneous variables that share an unobserved cause. If the target variable $Y$ is exogeneous then any causal model with more than one endogeneous direct effect of $Y$ violates the assumptions. In order to enable an application of CR on datasets with more than one effect variable we assume knowledge of the SCM including $Y$ for CR as well and draw ground-truth interventional samples from the SCM instead of identifying the interventional distribution from observational data.

For \textit{3var-causal} and \textit{3var-noncausal} we configured NSGA-II to optimize over $600$ generations with a population size of $300$, for \textit{5var-skill} and \textit{7var-covid} $1000$ generations with $500$ individuals were used. For all experiments the crossover probability was $0.3$ and the mutation probability $0.05$. For all settings continuous variables were rounded to $1$ decimal point. For the 3 variable settings a standard \texttt{sklearn} \texttt{LogisticRegression} was used, for the refits without penality. For the nonlinear dataset a \texttt{RandomForestClassifier} with max depth $30$, $50$ estimators and balanced subsampling was applied. The experimental results were computed on a Quad core Intel Core i7-7700 Kaby Lake processor. For each setting, the experiments took between 24 to 48 hours.

\newpage
\section{Proofs}
\label{appendix:proofs}

As follows we provide the full proofs for Propositions \ref{prop:ind-post-recourse:link-gamma} - \ref{prop:individualized-post-recourse-prediction}.
\subsection{Linking individualized prediction with \texorpdfstring{$\gamma^{ind}$}{}, Proof of Proposition \ref{prop:ind-post-recourse:link-gamma}}

\begin{repproposition}{prop:ind-post-recourse:link-gamma}
The expected individualized post-recourse score is equal to the individualized improvement probability $\gamma^{ind}(x^{pre},a) := P(Y^{post}=1|x^{pre}, do(a))$, i.e.
\begin{equation*}
E[\hat{h}^{*, ind}(x^{post})|x^{pre}, do(a)] = \gamma^{ind}(a).
\end{equation*}
\end{repproposition}

\textit{Proof:} It holds that
\begin{align*}
    &E[h^{*, ind}(x^{post})|x^{pre}, do(a)] \\
    &= E[E[Y|x^{pre}, x^{post}]| x^{pre}, do(a)]\\
    \qquad &\myeq{total exp.} \qquad E[Y|x^{pre}, do(a)]\\
    &= \gamma^{ind}(a).
\end{align*}

\subsection{Intervention stability w.r.t. ICR actions, Proposition \ref{prop:causes-only}}

\begin{repproposition}{prop:causes-only}
 %Given nonzero cost for all interventions, then ICR exclusively suggests actions on causes of $Y$.
 Given nonzero cost for all interventions, ICR exclusively suggests actions on causes of $Y$. Assuming causal sufficiency, for any optimal predictor the conditional distribution of $Y$ given the variables that the model uses $X_S$ (i.e. $P(Y|X_S)$) is stable w.r.t interventions on causes. Therefore, optimal predictors are intervention stable w.r.t. ICR actions.
\end{repproposition}

\textit{Proof:} We prove the statement in six steps.

\textit{ICR only intervenes on causes:} The goal of meaningful recourse is to improve $Y$ with minimal cost. Only interventions on causes alter $Y$. Consequently, actions on non-causes of $Y$ would not be suggested by meaningful recourse.

\textit{Given causal sufficiency, a graph $\mathcal{G}$ and an endogenous $Y$, the set of endogeneous direct parents, direct effects and direct parents of effects are the minimal $d$-separating set $S_{\mathcal{G}}$:} Standard result, see e.g. \citet{Peters2017book}, Proposition 6.27. %The set of direct parents, direct effects and direct parents of effects $S_\mathcal{G}$ is the minimal set that $d$-separates all endogenous variables from $Y$: All members of $S_\mathcal{G}$ are trivially $d$-separted from $Y$. Every path $p$ from a variable $X_l \not \in S_\mathcal{G}$ to $Y$ must enter $Y$ either via one of its causes or via one of its effects. Since all direct causes are observed, all paths via causes $\dots \rightarrow X_i \rightarrow Y$ or $\dots \leftarrow X_i \rightarrow Y$ are blocked. All paths via direct effects may either be of the form $Y \rightarrow X_i \rightarrow \dots$, which are blocked since $X_i$ are observed, or of the form $Y \rightarrow X_i \leftarrow X_j \leftarrow \dots $ or $Y \rightarrow X_i \leftarrow X_j \rightarrow \dots $ which are blocked since any direct parent of $X_i$ is in $S$. Therefore, it also $d$-separates any subset of the endogenous variables from $Y$. If any direct parent or direct effect would not be in the set, then the respective variable would not be $d$-separated by $Y$. Since all direct effects are in the set, any parent of a direct effect has to be in the set, since otherwise the parent would not be $d$-separated from $Y$.\\

\textit{The set $S_{\mathcal{G}^*}$ in the augmented graph $\mathcal{G}^*$ coincides with $S_{\mathcal{G}}$:} The minimal $d$-separating set contains direct causes, direct effects and direct parents of direct effects. $I_l$ is never a direct cause of $X_l$. Also, since $I_l$ has no endogenous causes, it cannot be a direct effect. Furthermore, since we restrict interventions to be performed on causes, $I_l$ cannot be a direct parent of a direct effect.

\begin{figure}[t]
    \centering
    % tikz code for the intervention-types figure

\begin{tikzpicture}[thick, scale=0.6, every node/.style={scale=.6, line width=0.25mm, black, fill=white}]
    \usetikzlibrary{shapes}
    
        % pre recourse nodes
        \node[circle, draw,scale=0.9] (x1) at (-2, 1) {$X_1$};
        \node[circle, double, draw,scale=0.9] (x2) at (-1, 1) {$X_2$};
        \node[circle, draw,scale=0.9] (x3) at (-3, 0) {$X_3$};
        \node[circle, double, draw,scale=0.9] (x4) at (-2, 0) {$X_4$};
        \node[circle, double, draw,scale=0.9] (x5) at (-1, -1) {$X_5$};
        \node[circle, double, draw,scale=0.9] (x6) at (1, 1) {$X_6$};
        \node[circle, double, draw,scale=0.9] (x7) at (2, 1) {$X_7$};
        \node[circle, double, draw,scale=0.9] (x8) at (1, -1) {$X_8$};
		\node[draw,scale=1.3] (y) at (0, 0) {$Y$};
		
		% pre recourse relationships
		\draw[->] (x2) -- (x1);
		\draw[->] (x2) -- (y);
		\draw[->] (x3) -- (x4);
		\draw[->] (x4) -- (y);
		\draw[->] (x5) -- (y);
		\draw[->] (y) -- (x6);
		\draw[->] (x7) -- (x6);
		\draw[->] (x8) -- (x6);
		\draw[->] (x8) -- (x5);
		
		% intervention auxilary variables

		\node[draw=orange, ellipse, scale=0.9] (i1) at (-2, 2) {$\mathcal{I}_1$};
		\node[draw=green, ellipse, scale=0.9] (i2) at (-1, 2) {$\mathcal{I}_2$};
		\node[draw=green, ellipse, scale=0.9] (i3) at (-3, -2) {$\mathcal{I}_3$};
		\node[draw=green, ellipse, scale=0.9] (i4) at (-2, -2) {$\mathcal{I}_4$};
		\node[draw=green, ellipse, scale=0.9] (i5) at (-1, -2) {$\mathcal{I}_5$};
		\node[draw=red, ellipse, scale=0.9] (i6) at (1, 2) {$\mathcal{I}_6$};
		\node[draw=orange, ellipse, scale=0.9] (i7) at (2, 2) {$\mathcal{I}_7$};
		\node[draw=green, ellipse, scale=0.9] (i8) at (1, -2) {$\mathcal{I}_8$};

		\draw[->, orange] (i1) -- (x1);
		\draw[->, green] (i2) -- (x2);
		\draw[->, green] (i3) -- (x3);
		\draw[->, green] (i4) -- (x4);
		\draw[->, green] (i5) -- (x5);
		\draw[->, red] (i6) -- (x6);
		\draw[->, orange] (i7) -- (x7);
		\draw[->, green] (i8) -- (x8);
		
		\draw[-, dotted] (3, -2.5) -- (3, 2.5);
		
		% all variables observed
		
		\node[circle, double, draw,scale=0.9] (x1r) at (5, 1) {$X_1$};
        \node[circle, dotted, draw,scale=0.9] (x2r) at (6, 1) {$X_2$};
        \node[circle, draw,scale=0.9] (x3r) at (4, 0) {$X_3$};
        \node[circle, double, draw,scale=0.9] (x4r) at (5, 0) {$X_4$};
        \node[circle, double, draw,scale=0.9] (x5r) at (6, -1) {$X_5$};
        \node[circle, double, draw,scale=0.9] (x6r) at (8, 1) {$X_6$};
        \node[circle, double, draw,scale=0.9] (x7r) at (9, 1) {$X_7$};
        \node[circle, dotted, draw,scale=0.9] (x8r) at (8, -1) {$X_8$};
		\node[draw,scale=1.3] (yr) at (7, 0) {$Y$};

		% pre recourse relationships
		\draw[->] (x2r) -- (x1r);
		\draw[->] (x2r) -- (yr);
		\draw[->] (x3r) -- (x4r);
		\draw[->] (x4r) -- (yr);
		\draw[->] (x5r) -- (yr);
		\draw[->] (yr) -- (x6r);
		\draw[->] (x7r) -- (x6r);
		\draw[->] (x8r) -- (x6r);
		\draw[->] (x8r) -- (x5r);
		
		% intervention auxilary variables

		\node[draw=red, ellipse, scale=0.9] (i1r) at (5, 2) {$\mathcal{I}_1$};
		\node[draw=red, ellipse, scale=0.9] (i2r) at (6, 2) {$\mathcal{I}_2$};
		\node[draw=green, ellipse, scale=0.9] (i3r) at (4, -2) {$\mathcal{I}_3$};
		\node[draw=green, ellipse, scale=0.9] (i4r) at (5, -2) {$\mathcal{I}_4$};
		\node[draw=red, ellipse, scale=0.9] (i5r) at (6, -2) {$\mathcal{I}_5$};
		\node[draw=red, ellipse, scale=0.9] (i6r) at (8, 2) {$\mathcal{I}_6$};
		\node[draw=orange, ellipse, scale=0.9] (i7r) at (9, 2) {$\mathcal{I}_7$};
		\node[draw=red, ellipse, scale=0.9] (i8r) at (8, -2) {$\mathcal{I}_8$};

		\draw[->, red] (i1r) -- (x1r);
		\draw[->, red] (i2r) -- (x2r);
		\draw[->, green] (i3r) -- (x3r);
		\draw[->, green] (i4r) -- (x4r);
		\draw[->, red] (i5r) -- (x5r);
		\draw[->, red] (i6r) -- (x6r);
		\draw[->, orange] (i7r) -- (x7r);
		\draw[->, red] (i8r) -- (x8r);
		
    \end{tikzpicture}
    \caption{A schematic drawing illustrating under which interventions $I_1, \dots, I_8$ the Markov blanket (double circle) is intervention stable. In this setting, we consider the intervention variables to be independent treatment variables: We would like to know how the different actions influence the conditional distribution, irrespective of how likely they are to be applied. Therefore, they are modeled as parent-less variables. Green indicates intervention stability, red indicates no intervention stability. Orange indicates intervention stability of non-causal variables. Dotted variables are not observed. \textit{Left:} Since all endogenous variables are observed, $MB_O(Y)$ is stable w.r.t. interventions on every endogenous cause of $Y$ (Proposition \ref{proposition:all-observed}). \textit{Right:} Unobserved variables ($X_2, X_8$) open paths between interventions on causes and $Y$.
    }
    \label{fig:intervention-types}
\end{figure}
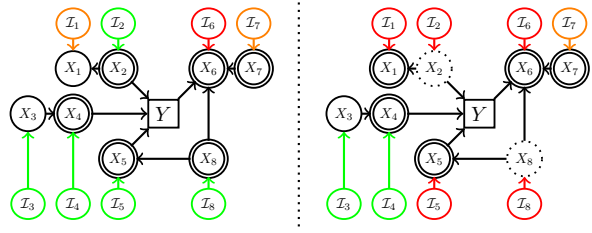

\textit{$S_\mathcal{G}$ is intervention stable:} As follows, all intervention variables are $d$-separated from $Y$ in $\mathcal{G^*}$ by $S_\mathcal{G}$. Therefore $S_{\mathcal{G}}$ is intervention stable. An example is given in Figure \ref{fig:intervention-types}.

\textit{Then also the markov blanket is intervention stable:} %Without loss of generality we can assume that $MB(Y) = S$.
Since $d$-separation implies independence $MB(Y) \subseteq S_\mathcal{G}$. Therefore, if $X_T \idp Y | X_{MB(Y)}$ then also $X_T \idp Y | S_{\mathcal{G}}$. If any element $s \in S_\mathcal{G}$ it holds that $s \not \in MB(Y)$, then it must hold that $X_s \idp Y | X_{MB(Y)}$. Therefore, if $X_T \idp Y | X_{MB(Y)}, X_s$ then also $X_T \idp Y | X_{MB(Y)}$ and therefore any independence entailed by $S_\mathcal{G}$ also holds for $MB(Y)$. Since \citet{pfister_stabilizing_2019} only require the independence that is implied by $d$-separation in their invariant conditional proof, the same implication holds for the $MB(Y)$. As follows, $P(Y|X_{MB(Y)})$ is invariant with respect to interventions on any set of endogenous causes.

\textit{Then any superset of the markov blanket is intervention stable:} We prove the statement by contradiction. The markov blanket $d$-separates the target variable $Y$ from any other set of variables. If adding a set of variables $S_1$ to the markov blanket would open a path to any other set of variables $S_2$, then it would hold that $S := S_1 \cup S_2$ is not $d$-separated from $Y$ ($P(Y|MB(Y)) = P(Y|MB(Y), S_1, S_2) \neq P(Y|MB(Y), S_1) = P(Y|MB(Y))$)

\subsection{Linking observational prediction and \texorpdfstring{$\gamma^{sub}$}{}, Proposition \ref{proposition:all-observed}}
\label{proof:proposition:all-observed}

\begin{repproposition}{proposition:all-observed}
%Given causal sufficiency, any set $S$ that allows for an optimal prediction (i.e., $MB(Y) \subseteq S$) is stable with respect to actions on causes of $Y$.
Given causal sufficiency and positivity\footnote{Positivity ensures that the post-recourse observation lies within the observational support , where the model was trained (i.e., $p^{pre}(x^{post}) > 0)$, \citep{neal2020introduction}).}, %any set $S$ that allows for an optimal prediction (i.e., $MB(Y) \subseteq S$) is stable with respect to ICR actions.
%As a consequence, 
for interventions on causes
the expected subgroup-wide optimal score $h^*$ is equal to the subgroup-wide improvement probability $\gamma^{sub}(a) := P(Y^{post} = 1|do(a), x^{pre}_{G_{a}})$, i.e.
\[E[\hat{h}^{*}(x^{post})|x^{pre}_{G_{a}}, do(a)] = \gamma^{sub}(a).\]
\end{repproposition}

\textit{Proof:} The proposition follows from Proposition \ref{prop:causes-only}. More specifically
\begin{align}
    E[h^{*}(x^{post,a})|x^{pre}_{G}, a] 
    = E[E[Y|x^{post, a}]| x^{pre}_{G}, a] 
    \qquad \myeq{total exp.} \qquad E[Y|x^{pre}_{G}, a]
    \qquad \myeq{def. $\gamma^{sub}$} \qquad \gamma^{sub}(a).
\end{align}

\subsection{Acceptance Bound, Proof of Proposition \ref{prop:acceptance-bound}}
\label{appendix:proof:acceptance-bound}

\begin{repproposition}{prop:acceptance-bound}
%Let $x_S^{pre}$ be the subset of the pre-recourse observed variables that were taken into account to compute ICR, i.e. $S=D$ for individualized recourse and $S=G_{a'}$ for subpopulation-based recourse.
%For subpopulation-based ICR, we furthermore assume that $p^{pre}(x^{post}) > 0$.
%For pre-recourse state $x^{pre}$, a $\gamma$-confident action $a$, the (individualized) optimal predictor $h^{*}$ and a global decision threshold $t$ the post-recourse acceptance probability $\eta$ is
%
%$$\eta(t; x^{pre}_S, a) \geq \frac{\gamma(x^{pre}_S,a) - t}{1 - t}.$$
%
%where $\gamma(x^{pre}_S, a) = p(Y^{post,a}=1|x^{pre}_S, a)$. 
Let $g$ be a predictor with $E[g(x^{post}) | x_S^{pre}, do(a)] = \gamma(x_S^{pre}, a)$.
Then for a decision threshold $t$ the post-recourse acceptance probability $\eta(t; x^{pre}_S, a) := P(g(x^{post}) > t|x^{pre}_S, do(a))$ is lower bounded:
\[ \eta(t; x^{pre}_S, a) \geq \frac{\gamma(x^{pre}_S,a) - t}{1 - t}.\]
\end{repproposition}
\textit{Proof:}
Positivity ($p^{pre}(x^{post}) > 0$) is necessary for subpopulation-based ICR since only then we can assume that the model is actually optimal for any input that it receives. The problem is discussed in more detail in \citet{hernan_ma_causal_2020,neal2020introduction}.

As follows we denote $\hat{h}^{*}$ as the random variable indicating the predictions of the post-recourse predictors described in Section \ref{sec:mcr}.\\
From Propositions \ref{prop:ind-post-recourse:link-gamma} and \ref{proposition:all-observed}, for both individualized and subpopulation-based post-recourse predictors we know that

$$E[\hat{h}(x^{post,a})^{*}|x^{pre}_{S}, do(a)] = \gamma(x^{pre}_{S},a).$$

We decompose the expected prediction

\begin{align}
    \gamma(x^{pre}_S, a)
    &= E[\hat{h}^{*}|x^{pre}_{S}, a] \\
    \begin{split}
    &\left. = %\begin{tabular}{l}
    E[\hat{h}^{*}|\hat{h}^{*} > t]P(\hat{h}^{*} > t)%\\
    + E[\hat{h}^{*}|\hat{h}^{*} \leq t] P(\hat{h}^{*} \leq t)% \\
    %\end{tabular}
    \right|_{x^{pre}_{S}, a}
    \end{split}\\
    \begin{split}
    &\left. = %\begin{tabular}{l}
      E[\hat{h}^{*}|\hat{h}^{*} > t]P(\hat{h}^{*} > t)
      + E[\hat{h}^{*}|\hat{h}^{*} \leq t] (1 - P(\hat{h}^{*} > t))
    %\end{tabular}
    \right |_{x^{pre}_{S}, a}\\
    \end{split}\\
    \begin{split}
    &= \left. 
    %\begin{tabular}{l}
        E[\hat{h}^{*}|\hat{h}^{*} > t]P(\hat{h}^{*} > t)
        + E[\hat{h}^{*}|\hat{h}^{*} \leq t] - P(\hat{h}^{*} > t) E[\hat{h}^{*}|\hat{h}^{*} \leq t]
    %\end{tabular}
     \right|_{x^{pre}_{S}, a}
    \end{split}\\
    \begin{split}
    &= \left.%\begin{tabular}{l}
        E[\hat{h}^{*}|\hat{h}^{*} \leq t]
        + P(\hat{h}^{*} > t) \Big(E[\hat{h}^{*}|\hat{h}^{*} > t] - E[\hat{h}^{*}|\hat{h}^{*} \leq t]\Big)
    %\end{tabular}
      \right|_{x^{pre}_{S}, a}
    \end{split}
\end{align}

which can be reformulated to yield the acceptance rate $\eta$:
\begin{align}
    \left. \frac{\gamma-E[\hat{h}^{*}|\hat{h}^{*} \leq t]}{E[\hat{h}^{*}|\hat{h}^{*} > t] - E[\hat{h}^{*}|\hat{h}^{*} \leq t]}  \right |_{x^{pre}_S, a} %\\
    = P(\hat{h}^{*} > t|x^{pre}_S, a) = \eta(x^{pre}_S, a).
\end{align}
It holds that $E[\hat{h}^{*,ind}|\hat{h}^{*} \leq t] = FNR(t)$ and $E[\hat{h}^{*}|\hat{h}^{*} > t] = TPR(t)$.\\
\\
We can show that $E[\hat{h}^{*}|\hat{h}^{*} \leq t] \leq t$:
\begin{align}
    &0 \leq FNR(t|x^{pre}_S, a) \\
    &= P(Y^{a, post} = 1 | h^{*} \leq t, x^{pre}_S, a)\\
    &= E[Y^{a, post}|h^{*} \leq t, x^{pre}_S, a]\\
    &= E[E[Y^{a, post}|x^{post, a}]|h^{*} \leq t, x^{pre}_S, a]\\
    &= E[h^{*}|h^{*}\leq t, x^{pre}_S, a]\\
    &\leq t
\end{align}
and analog that $1 \geq TPR(t) \geq t$. Therefore
\begin{align}
\eta(t, x^{pre}_S, a) 
= \left. \frac{\gamma - FNR(t)}{TPR(t) - FNR(t)} \right |_{x^{pre}_S, a} 
\geq \frac{\gamma(x^{pre}_{S}, a) - FNR(t)}{1 - FNR(t)} \geq \frac{\gamma(x^{pre}_S, a) - t}{1 - t}.
\end{align}
\subsection{Individualized post-recourse prediction, proof of Proposition \ref{prop:individualized-post-recourse-prediction}}
\label{appendix:proofs:prop:individualized-post-recourse-prediction}
\begin{repproposition}{prop:individualized-post-recourse-prediction}
In general, the individualized post-recourse predictor can be estimated as
\begin{align}
&p(y^{post}|x^{pre}, x^{post}, do(a))\\
&= \frac{\int_{\mathcal{U}} p(y^{post}, x^{post}|u, do(a)) p(u|x^{pre})du}{\sum_{y' \in \{0, 1\}} \left( \int_{\mathcal{U}} p(y', x^{post}|u, do(a)) p(u|x^{pre}) du\right)}
\label{eq:individualized-post-recourse-prediction-appendix}
\end{align}
Given binary decision problems with invertible structural equations, the individualized post-recourse prediction function reduces to
\begin{align}
    &p(y^{post}|x^{post}, x^{pre}, do(a)) \\
    &= \frac{p(U_{-I} = f_{do(a)}^{-1}(y^{post}, x^{post})|x^{pre}, do(a))}{\sum_{y' \in \{0, 1\}} p(U_{-I} = f_{do(a)}^{-1}(y', x^{post})|x^{pre}, do(a))}.
    \label{eq:individualized-post-recourse-prediction-invertible-appendix}
\end{align}
\end{repproposition}

\textit{Proof:} It holds that 
\begin{align}
    p(y^{post}|x^{pre}, x^{post}, do(a)) %\\
    \qquad \myeq{def. cond.} \qquad \frac{p(y^{post}, x^{post}|x^{pre}, do(a))}{p(x^{post}|x^{pre}, do(a))}\\
\end{align}
We can reformulate the conditional distribution $p(y^{post}, x^{post}|x^{pre}, do(a))$ as two parts, one that describes the probability of a state of the context given $x^{pre}$, and one that describes the probability of a post-recourse state $x^{post}, y^{post}$ given a certain noise state $u$ and $do(a)$.
\begin{align}
    &p(y^{post}, x^{post}|x^{pre}, do(a))\\
    \quad \quad &\myeq{marginal.} \quad \int_{\mathcal{U}} p(y^{post}, x^{post}, u |x^{pre}, do(a)) du \\
    &\myeq{chain rule} \quad \int_{\mathcal{U}} p(y^{post}, x^{post}|u, x^{pre}, do(a))p(u|x^{pre}) du\\
    &\myeq{$(y, x)^{post} \perp x^{pre} | u$} \quad \quad \quad \int_{\mathcal{U}} p(y^{post}, x^{post}|u, do(a))p(u|x^{pre}) du.
\end{align}
In combination we yield
\begin{align}
&p(y^{post}|x^{pre}, x^{post}, do(a))\\
&= \frac{\int_{\mathcal{U}} p(y^{post}, x^{post}|u, do(a)) p(u|x^{pre})du}{\int_{\mathcal{Y}} \left( \int_{\mathcal{U}} p(y', x^{post}|u, do(a)) p(u|x^{pre}) du\right) dy'} \\
&= \frac{\int_{\mathcal{U}} p(y^{post}, x^{post}|u, do(a)) p(u|x^{pre}) du}{\sum_{y' \in {0, 1}} \left( \int_{\mathcal{U}} p(y', x^{post}|u, do(a)) p(u|x^{pre}) du\right)}
\end{align}
For a setting with invertible structural equations this reduces to
\begin{align}
    &p(y^{post}|x^{post}, x^{pre}, do(a))\\
    &= \frac{p(y^{post}, x^{post}|x^{pre}, do(a))}{p(x^{post}|x^{pre}, do(a))}\\
    &= \frac{p(U_{-I} = f^{-1}(y^{post}, x^{post})|x^{pre}, do(a))}{\sum_{y' \in \{0, 1\}} p(U_{-I} = f^{-1}(y^{post}, x^{post})|x^{pre}, do(a)) }.
\end{align}
where $-I$ is the index set for variables that have not been intervened on (since the noise terms for the intervened upon variables are isolated variables in the interventional graph).%

\newpage
\section{Misc}
\label{appendix:details}

\subsection{Negative Result: Algorithmic recourse is neither meaningful nor robust}
%\label{sec:negative-result}
\label{appendix:details:negative-result}

In the introduction we claimed that CR recommendations \citep{karimi_algorithmic_2020,karimi_algorithmic_2021} may not lead to improvement.
Now, we formally demonstrate the case on the Covid hospital admission example (Figure \ref{fig:example-intro}) which we extend with the full structural causal model (Example \ref{example:covid-admission}). Furthermore, we show that CR is not robust to refits of the model on mixed pre- and post-recourse data. All code is publicly available via \url{https://anonymous.4open.science/r/icr-aaai/README.md}.
\begin{example} Let $V$ indicate whether someone is fully vaccinated, $Y$ indicate whether someone is free of Covid and $S$ whether someone is asymptomatic. The data is generated by the following structural causal model (SCM) entailing the causal graph depicted in Figure \ref{fig:example-intro}:
\begin{align}
    &V := U_V, &U_V \sim Bern(0.5)\\
    &Y := V + U_Y \texttt{ mod } 2, &U_Y \sim Bern(0.09) \\
    &S := Y + U_S \texttt{ mod } 2, &U_S \sim Bern(0.05)
\end{align}
For prediction, a \texttt{sklearn} logistic regression model is fit on $2000$ samples, yielding $\hat{h}$ with $\beta_v \approx 3.7$, $\beta_s \approx 5.1$, $\beta_0 \approx -4.3$. Visitors are allowed to enter the hospital if $\hat{h} < 0.5$. Intervening on (flipping) $V$ and $S$ costs $0.5$ and $0.1$ respectively.
\label{example:covid-admission}
\end{example}
\textit{Lack of improvement:} Given a decision threshold of $0.5$, the model admits everyone without symptoms ($S=1$), irrespective of their vaccination status $V$. Therefore, in order to revert rejections ($S = 0$), both individualized and subpopulation-based CR suggest removing the symptoms $S$ ($do(S=1)$, for instance by taking cough drops).
%All recourse-seeking individuals can revert the model's decision.
However, since they only treat the symptoms $S$, the actual Covid risk $Y$ is unaffected: none of the recourse-implementing individuals actually improve. We say the predictor is \textit{gamed}.

%\begin{robustness}
\textit{Lack of robustness:} For individuals who implement recourse the association between symptom state $S$ and Covid risk $Y$ is broken. Thus, the predictive power of the model for recourse-seeking individual drops from $\approx 95$ percent pre-recourse to $\approx 5$ percent post-recourse.\footnote{The previously wrongly-rejected individuals are correctly classified after implementing recourse.}
A refit of the model on a mix pre- and post-recourse data ($2000$ samples each) yields $\hat{h}$ with $\beta_V\approx 4.1, \beta_S \approx 3.3, \beta_0 \approx -4.8$. Since the association between symptom state and disease status is broken post-recourse, the new model rejects individuals if they are not vaccinated, irrespective of their symptom state. For that reason, recourse recommendations that were designed for the original model only lead to acceptance by the refitted model for those individuals who happened to be vaccinated anyway.\\
The example demonstrates that CR recommendations are prone to gaming the predictor and therefore may neither lead to improvement nor be robust to model refits.
%\end{robustness}
%

\subsection{Interpretability of improvement confidence \texorpdfstring{$\gamma$}{}}
\label{appendix:details:expectation-objective}

Counterfactuals are concerned with changing the inputs to the model such that the model prediction changes in the desired way. Since the prediction function is deterministic and accessible, the post-recourse prediction can be determined exactly.\\
In contrast CR and ICR deal with the effects of real-world interventions on real-world variables. As such, the effects of recourse actions on the covariates (and the underlying prediction target) cannot be determined exactly. Therefore both CR and ICR have to deal with uncertainty.\\
CR deals with this uncertainty by phrasing the optimization objective for CR in terms of an expectation over the prediction distribution and by using an action-adaptive confidence threshold. This threshold $\texttt{thresh}$ bounds the expected prediction away from the model's decision threshold (e.g. $t=0.5$). Using the conservativeness parameters, the user can roughly steer how far the expected prediction shall be away from the decision boundary.\\
In contrast, ICR deals with the uncertainty by letting the user specify the confidence $\gamma$, which can be intuitively interpreted as improvement probability (whereas the expected prediction cannot be interpreted as acceptance probability). A lower-bound on the acceptance probability for a combination of $\gamma$ and $t$ is given in Proposition \ref{prop:acceptance-bound}. Furthermore, we can estimate the individualized and subpopulation-based acceptance rates for a specific situation $(a, x^{pre})$ as detailed in \ref{appendix:estimation:individualized} and \ref{appendix:estimation:sampling-sub}. The human-interpretable improvement and acceptance confidences are vital for the explainee to make an informed decision.\\
In order to allow a direct comparison of the methods, we rephrase the CR objective to optimize the acceptance probability $\eta$ in our experiments.

\subsection{Imbalance between standard predictors and individualized ICR recommendations}
\label{appendix:details:imbalance}

In Section \ref{sec:accurate-post-recourse} we argued that there is an imbalance in predictive capability between (optimal) observational predictors and the pre-recourse SCM (which used to predict $\gamma^{ind}$). We illustrate the problem on a simple example.
\begin{example}
Let there be a three variable chain $X_1 \rightarrow Y \rightarrow X_2$ where at every step the value is incremented by one with $50\%$ chance and the maximum value is set to 2 ($X_1 := U_1$, $Y := X_1 + U_Y$, $X_2 := min(2, Y + U_2)$ where $U_1, U_2, U_Y \sim Bern(0.5)$). Let us assume a factual observation $x^{pre} = (0, 2)$ and action $a = {do(X_1 = 1)}$ yielding $x^{post} = (1, 2)$. For the observation $x^{pre} = (0, 2)$ we can infer that $U_Y$ must have been $1$, since two increments are needed to get from $0$ to $2$. However, from the post-intervention observation $x^{post}=(1,2)$ we cannot infer where the increment happened ($U_Y$ or $U_2$). As a consequence, an optimal predictive model that only has access to $x^{post}$ would predict that $y^{post}$ for $x^{post} = (1, 2)$ could be $1$ or $2$ with equal likelihood. In contrast, with access to $x^{pre}$ and the $SCM$ we can infer that $y^{post}=2$ since $U_Y=1$.
\end{example}
In the above example, given knowledge of the SCM, the pre-intervention observation $x^{pre}$ and the performed action $a$ we can already abduct $U_Y$ perfectly and therefore correctly determine the post-intervention state of $Y$ (even without access to the post-intervention observation $x^{post}$).
In contrast, with the post-recourse observation alone it is impossible to reconstruct $U_Y$ and therefore impossible to determine the post-intervention state of $Y$.\footnote{The optimal pre-recourse predictor $\hat{h}^*(x^{post})$ predicts $0.5$ for both $y=1$ and $y=2$.}
In the context of ICR this means that the observational predictor's post-recourse predictions are not directly linked with $\gamma$: they may not honor the implementation of actions with $\gamma^{ind}=1$. As a consequence, we suggested to use the SCM for post-recourse prediction in Section \ref{sec:accurate-post-recourse}.
 %%% Uncomment this line and comment out the ``thebibliography'' section below to use the external .bib file (using bibtex) .





% \end{thebibliography}

\end{document}